\documentclass[twoside]{article}
\usepackage{ecj,palatino,epsfig,latexsym,natbib}

\usepackage[latin1]{inputenc}
\usepackage{stfloats}
\usepackage{graphicx}
\usepackage{subfigure}
\usepackage{fancyhdr,fancybox}
\usepackage{amssymb,amsfonts,amstext,amscd,amsbsy,amsmath,marvosym}
\usepackage[algoruled,linesnumbered,vlined]{algorithm2e}
\usepackage{psfrag}
\usepackage{url}
\usepackage{multirow}
\usepackage{longtable}
\usepackage{rotating}
\usepackage[table]{xcolor}
\usepackage{array}

\newcommand{\eps}{\epsilon}

\usepackage{fancyhdr}
\begin{document}

\title{\bf The Cone $\eps$-Dominance: An Approach for Evolutionary Multiobjective Optimization}  

\author{\name{\bf Lucas de Souza Batista} \hfill \addr{lusoba@ufmg.br}\\ 
        \addr{Universidade Federal de Minas Gerais, Departamento de Engenharia Elétrica, Av. Antônio Carlos 6627, 31270-010, Belo Horizonte, MG, Brasil}
\AND
       \name{\bf Felipe Campelo} \hfill \addr{fcampelo@ufmg.br}\\
        \addr{Universidade Federal de Minas Gerais, Departamento de Engenharia Elétrica, Av. Antônio Carlos 6627, 31270-010, Belo Horizonte, MG, Brasil}
\AND
       \name{\bf Frederico Gadelha Guimar\~aes} \hfill \addr{fredericoguimaraes@ufmg.br}\\
        \addr{Universidade Federal de Minas Gerais, Departamento de Engenharia Elétrica, Av. Antônio Carlos 6627, 31270-010, Belo Horizonte, MG, Brasil}
\AND
       \name{\bf Jaime Arturo Ram\'irez} \hfill \addr{jramirez@ufmg.br}\\
        \addr{Universidade Federal de Minas Gerais, Departamento de Engenharia Elétrica, Av. Antônio Carlos 6627, 31270-010, Belo Horizonte, MG, Brasil}              
}

\maketitle

\begin{abstract}
We propose the cone$\eps$-dominance approach to improve convergence and diversity in multiobjective evolutionary algorithms (MOEAs). A cone$\eps$-MOEA is presented and compared with MOEAs based on the standard Pareto relation (NSGA-II, NSGA-II*, SPEA2, and a clustered NSGA-II) and on the $\eps$-dominance ($\eps$-MOEA). The comparison is performed both in terms of computational complexity and on four performance indicators selected to quantify the quality of the final results obtained by each algorithm: the convergence, diversity, hypervolume, and coverage of many sets metrics. Sixteen well-known benchmark problems are considered in the experimental section, including the ZDT and the DTLZ families. To evaluate the possible differences amongst the algorithms, a carefully designed experiment is performed for the four performance metrics. The results obtained suggest that the cone$\eps$-MOEA is capable of presenting an efficient and balanced performance over all the performance metrics considered. These results strongly support the conclusion that the cone$\eps$-MOEA is a competitive approach for obtaining an efficient balance between convergence and diversity to the Pareto front, and as such represents a useful tool for the solution of multiobjective optimization problems.
\end{abstract}


\section{Introduction}
\label{sec:1}

Applied design problems are usually expressed as optimization problems that involve multiple and often conflicting objectives. The goal of solving these multiobjective optimization problems (MOPs) is to find a set of trade-off solutions, known as Pareto-optimal or nondominated solutions, in which each solution represents different compromises for the objectives. In this context, multiobjective evolutionary algorithms (MOEAs) have been widely recognized as suitable approaches to solve MOPs, mainly due to their robustness and generality. In general, researchers have found that MOEAs are more practical compared to the preference-based classical approaches, since the user gets an opportunity to analyze a range of other trade-off solutions before choosing the preferable one. Besides, the MOEA search procedure is also algorithmically efficient as the population evolves simultaneously towards many different regions of the Pareto-optimal front \citep{journal.Zitzler2000,book.Coello2002,journal.Deb2005a}.

In the design of a MOEA, the algorithm is often required to find a well-distributed set of solutions near to the global Pareto front, thus enabling the decision-maker to choose the most suitable design by considering some higher-level problem information. Essentially, a high-quality approximation set should approach the true Pareto front as close as possible, and be well-spread along its extension \citep{rep.Zitzler2001}. This fact implies that the evaluation of the quality of Pareto approximated sets achieved by evolutionary multiobjective algorithms represents itself a multi-criteria problem, in which convergence and diversity have to be considered simultaneously. To consider this compromise, most MOEAs incorporate an external archive in which the nondominated solutions estimated during the search process are stored. The external or memory archives impose an elitist mechanism to MOEAs, and only the solutions that are globally nondominated with respect to all the solutions generated so far by the MOEA are maintained. As the external archive size is usually finite, some truncation techniques have been developed in order to enforce a good distribution of solutions into the archive, in order to obtain a good characterization of the global Pareto front. In general, a Pareto-based fitness assignment method is usually designed in order to guide the search toward the global Pareto-optimal front, whereas the population diversity is commonly promoted by employing density estimation methods such as clustering \citep{journal.Zitzler1999}, crowding \citep{journal.Deb2002}, adaptive grids \citep{journal.Knowles2000}, entropy \citep{inproc.Farhang-Mehr2002}, or relaxed forms of Pareto dominance \citep{inproc.Ikeda2001,journal.Laumanns2002,journal.HernandezDiaz2007,inproc.Sato2007,inproc.Batista2011}. To help contextualize the scope of this work, some of these techniques are reviewed throughout the text.

In the current paper, we propose a relaxed form of dominance called cone$\eps$-dominance that is incorporated into a steady-state MOEA. The cone$\eps$-MOEA is contrasted to five well known approaches, namely the NSGA-II (two versions of this method are considered) \citep{journal.Deb2002}, the $\eps$-MOEA \citep{proc.Deb2003,journal.Deb2005a}, the SPEA2 \citep{rep.Zitzler2001} and the clustered-NSGA-II \citep{proc.Deb2003}, both in terms of computational complexity and on a number of performance metrics selected to quantify the convergence and diversity preservation abilities of the algorithms. After analyzing the results obtained, both in terms of statistical significance and magnitude of effects, it becomes clear that the proposed cone$\eps$-MOEA is capable of achieving an efficient balance between convergence and diversity, and as such represents an interesting and competitive approach to the solution of MOPs.

The paper is organized as follows: Section \ref{sec:2} reviews the basic definitions of the multiobjective optimization problem and the usual dominance criterion.  Section \ref{sec:2b} discusses some of the most popular relaxed dominance criteria, and Section \ref{sec:3} reports some practical diversity preservation methods employed in MOEAs. Section \ref{sec:4} contains a detailed description and mathematical definition of the cone$\eps$-dominance approach. Section \ref{sec:5} describes the MOEAs, the test problems, the performance metrics, and the statistical design employed in the evaluation of the proposed strategy. Section \ref{sec:6} presents a comprehensive analysis of the performance of the proposed approach and the other algorithms compared. Finally, the conclusions are outlined in Section \ref{sec:7}. 

\section{Mathematical Background}
\label{sec:2}

\subsection{Multiobjective Optimization Problem}

Multiobjective optimization problems can be stated, without loss of generality, as:
\begin{equation}
\begin{split}
\min\limits_{\pmb{x}}&\ \ \pmb{f}\left(\pmb{x}\right) = \left[f_1(\pmb{x}),\ldots,f_{m}(\pmb{x})\right]^{T}\\
\makebox{subject to: } &\begin{cases}
&g_{i}\left(\pmb{x}\right) \leq 0,\ i = 1,\ldots,n_g\\
&h_{j}\left(\pmb{x}\right) = 0,\ j = 1,\ldots,n_h\\
&\pmb{x} \in \mathbb{X}
\end{cases}
\end{split}
\label{eq:prob-ot}
\end{equation}

\noindent where $\pmb{x} \in \mathbb{X}$ is the vector of optimization variables; $\mathbb{X}\subset\mathbb{R}^n$ is the optimization domain, defined by the Cartesian product of the domains of each optimization variable;  $\pmb{f}\left(\cdot\right): \mathbb{X}\mapsto\mathbb{R}^{m}$ are the objective functions of the problem; $\pmb{g}\left(\cdot\right): \mathbb{X}\mapsto\mathbb{R}^{n_g}$ and $\pmb{h}\left(\cdot\right): \mathbb{X}\mapsto\mathbb{R}^{n_h}$ represent the inequality and equality constraints of the problem, respectively; and the set of feasible solutions is represented by $\Omega\subseteq\mathbb{X}$. In this context, the goal of multiobjective evolutionary algorithms is to obtain a diverse set of estimates of the Pareto optimal set, which contains the nondominated solutions of the multiobjective problem.

\subsection{The Standard Dominance Relation}

Pareto dominance \citep{book.Deb2001} has been the most commonly adopted criterion used to discriminate among solutions in the multiobjective context, and therefore it has been the basis to develop most of the MOEAs proposed so far, e.g., SPEA \citep{journal.Zitzler1999} and NSGA-II \citep{journal.Deb2002}.

By definition, a feasible solution $\pmb{x} \in \Omega$ Pareto dominates another point $\pmb{x}' \in \Omega$ if the following relation holds:
\begin{equation}
\pmb{f}(\pmb{x})\leq \pmb{f}(\pmb{x}')\mbox{ and } \pmb{f}(\pmb{x})\neq \pmb{f}(\pmb{x}')
\end{equation}

\noindent in which the relation operators $\leq$ and $\neq$ are defined as:
\begin{align}
\pmb{f}(\pmb{a}) &\leq \pmb{f}(\pmb{b})\Leftrightarrow f_i(\pmb{a}) \leq f_i(\pmb{b}),~~~ \forall~ i \in \left\{1,\ldots,m\right\}\\
\pmb{f}(\pmb{a}) &\neq \pmb{f}(\pmb{b})\Leftrightarrow \exists~ i \in \left\{1,\ldots,m\right\}: f_i(\pmb{a})\neq f_i(\pmb{b})
\end{align}

\noindent in which $\pmb{a}$ and $\pmb{b}$ represent two different decision vectors. In short, this dominance is usually expressed as $\pmb{f}(\pmb{x}) \prec \pmb{f}(\pmb{x}')$.

All solutions that are not dominated by any other decision vector of a given set are called {\it nondominated} regarding this set. In this way, the Pareto optimal set $\mathcal{P}$ is defined as the set of nondominated solutions:
\begin{equation}
\mathcal{P} = \left\{\pmb{x}^*\in\Omega \mid \nexists~ \pmb{x}\in\Omega: \pmb{f}(\pmb{x}) \prec \pmb{f}(\pmb{x}^*)\right\}~.
\end{equation}

The image of this set in the objective space is called the {\it Pareto front} $\mathcal{F} = \pmb{f}(\mathcal{P})$, i.e.:
\begin{equation}
\mathcal{F} = \left\{\pmb{y}=\pmb{f}(\pmb{x}) : \pmb{x} \in \mathcal{P}\right\}~.
\end{equation}

\section{Relaxed Dominance Criteria}
\label{sec:2b}
A few years ago, \cite{inproc.Ikeda2001} proposed a relaxed dominance criterion to deal with dominance resistant solutions, i.e., solutions that are extremely inferior to others in at least one objective, but hardly-dominated. The idea behind this approach, called $\alpha$-dominance, is to set lower and upper bounds of trade-off rates between two objectives, such that solutions characterized by a small improvement in some objectives (compared to the amount of detriment in one objective), which would be nondominated according to the standard Pareto dominance, would be rejected. Since there is no explicit formula for the calculation of $\alpha$, its choice is usually left to the designer, representing a difficult and problem-dependent task. Furthermore, even though the selective pressure induced by the $\alpha$-approach improves the convergence of a MOEA, there is no guarantee of finding a representative well-spread estimation of the real Pareto front.

Following Ikeda's work, \cite{journal.Laumanns2002} proposed two relaxed dominance methods: the additive and the multiplicative $\eps$-dominance schemes. These mechanisms act as an archiving strategy to ensure both properties of convergence towards the Pareto-optimal front and diversity among the solutions found. These archiving approaches essentially promote the convergence to a representative well distributed approximation of the global Pareto front, always preserving the best solutions found at intermediate iterations of the algorithm. These techniques guarantee that no two archived solutions can share a given $\eps_{i}$ neighborhood on the $i^{th}$ objective (with $i = 1,\ldots,m$, $m$ being the number of objectives), with the $\eps$ values usually provided by the designer to control the size (resolution) of the solution set. However, useful $\eps$ values are normally not known before executing a MOEA, since the equations used to estimate the values of $\eps$ are only valid for the case of linear Pareto fronts. This leads to difficulties in computing appropriate values of $\eps$ to provide the desired number of nondominated points, since the geometrical features of the Pareto-optimal front are commonly unknown by the designer, and the $\eps$-dominance strategy can lose a high number of viable solutions when the $\eps$ values are badly estimated. Moreover, this approach tends to neglect viable solutions since it does not allow two points with a difference of $\eps_{i}$ in the $i^{th}$ objective to be mutually nondominated. Because of this property, it is usually not possible to obtain solutions at the corners of the estimated Pareto front, contributing negatively to the spread of solutions along its extension \citep{journal.HernandezDiaz2007}. The $\eps$-dominance concept has been used and adapted in a number of studies, e.g. to improve the convergence characteristics of MOEAs \citep{journal.Zhao2010,journal.Li2011,journal.Hernandz-Diaz2011}, to address specific problems \citep{inproc.Aguirre2009}, or to tackle many-objective optimization problems \citep{inproc.Pasia2011}.

An alternative $\eps$-dominance strategy to overcome some of the limitations of $\eps$-dominance was later proposed by \cite{journal.HernandezDiaz2007}. This technique, called Pareto adaptive $\eps$-dominance (pa$\eps$-dominance for short), considers not only a different $\eps$ value for each objective, but a vector $\pmb{\eps} = \left(\pmb{\eps}_1,\ldots,\pmb{\eps}_m\right)$ associated to the objectives $\pmb{f} = \left(f_1,\ldots,f_m\right)$ depending on the geometrical characteristics of the Pareto-optimal front. In this way, different intensities of dominance are considered for each objective according to the position of each solution along the Pareto front. The size of the boxes, defined by the $\eps$ values, is adapted depending on the corresponding position in the objective space, so that smaller boxes are created where needed (e.g., at the extremes of the Pareto front), and larger ones in other less problematic parts of the front. Unfortunately, this approach also has some drawbacks. There are problems in which pa$\eps$-dominance is not able to maintain a good distribution of solutions at the extreme parts of the Pareto front. Moreover, the adaptive vector $\pmb{\eps}$ depends on the quality of the front used for its estimation \citep{journal.HernandezDiaz2007}. In fact, this affects the spread of solutions along the front and also has a negative effect on the performance of the method.

\cite{inproc.Sato2007} proposed to control the dominance area of solutions in order to induce appropriate selective pressure and ranking (in the objective domain) in MOEAs. This strategy controls the degree of contraction or expansion of the dominance area of solutions by modifying the fitness value of each objective function, which is attained by changing a user-defined parameter $S_i$, for all $i = 1, \dots, m$. As shown in \citep{inproc.Sato2007}, when $S_i < 0.5$, the $i^{th}$ fitness value $f_i(\pmb{x})$ is increased to $f_i'(\pmb{x}) > f_i(\pmb{x})$. On the other hand, when $S_i > 0.5$, $f_i(\pmb{x})$ is decreased to $f_i'(\pmb{x}) < f_i(\pmb{x})$. At last, when $S_i = 0.5$, $f_i'(\pmb{x}) = f_i(\pmb{x})$, which is equivalent to conventional dominance. However, since different rankings can be produced, the optimum parameter $\pmb{S}^*$ that yields maximum search performance depends strongly on the problem. In addition, either convergence or diversity can be emphasized by contracting or expanding the dominance area, but not always both simultaneously \citep{inproc.Sato2007}.

To address some of the limitations discussed above, a relaxation of the strict dominance concept based on an extension of the $\eps$-dominance criterion and on the use of cones to control the dominance region of solutions was suggested in \citep{inproc.Batista2011}. The main idea of this relaxation was to maintain the good convergence properties of $\eps$-dominance while simultaneously improving the control over the diversity and resolution of the estimated Pareto front, providing a dominance relation that is less sensitive to the geometrical features of the front. As seen before, some relaxed forms of dominance are sensitive to the loss of efficient solutions, mainly those located in certain portions of the front, e.g., segments with degenerated tradeoffs (small gains for one objective at the expense of large losses in another) and the extremes of the front. 

This loss of potentially interesting solutions is an undesired characteristic in many cases, since it usually leads to degradation in the  diversity of the fronts obtained. In fact, the design of effective mechanisms to maintain diversity remains as a key issue in a number of cases, for instance, when extending particle swarm optimizers to solve multiobjective optimization problems \citep{inproc.Villalobos2005}. Additionally, better diversity in the sampling of the Pareto front might be useful for (i) some interactive decision-making tools based on reconstructing the Pareto front \citep{inproc.Chen2002,book.Chankong2008}; (ii) the search for design rules and principles, such as in the \textit{innovization} concept advanced by \cite{inproc.Deb2006}; or (iii) \textit{a posteriori} analysis of the Pareto front to identify the variables that are relevant for the tradeoffs \citep{report.Deb2006}. As shown in \citep{inproc.Batista2011}, since the influence on the ordering of points performed by the cone of dominance can be limited to a local neighborhood in the objective space, the cone$\eps$-criterion enables the approximation of nondominated points in some adjacent boxes that would otherwise be $\eps$-dominated, including the extreme parts of the front, which contributes positively for the performance of the method. In this work we expand and improve the concepts initially introduced in that preliminary work, providing greater insight into its characteristics and addressing some of the questions regarding the diversity promoting properties of the cone$\eps$-dominance.

\section{Diversity Preservation into MOEA's Memory Archive}
\label{sec:3}

As indicated before, some techniques have been suggested over the years in order to approach and maintain a good distribution of solutions in the space of objectives. The ones that are used in this work are discussed in this section.

\subsection{Clustering}

In optimization problems defined in continuous search domains, the Pareto-optimal set can be extremely large or even infinite. Nevertheless, since approximating a large set of nondominated solutions is useless from the decision maker's point of view, pruning the external archive while maintaining some desirable characteristics is necessary or even mandatory. For instance, the SPEA proposed by \cite{journal.Zitzler1999} uses a clustering technique named \emph{average linkage method}. This approach is employed to prune the contents of the bounded memory archive of SPEA. In a general way, (i) each external nondominated solution compose initially a distinct cluster; (ii) the two closest ones amalgamate into a larger one; (iii) the second step is repeated until the desired number is achieved; finally, (iv) the reduced nondominated set is obtained by selecting a representative solution per cluster.

Another example concerns to the clustering technique used in SPEA2 \citep{rep.Zitzler2001}. This mechanism presents similarities with the truncation method of SPEA, however it does not loose boundary points. In short, when the nondominated front exceeds the archive limit, the point which has the smallest distance to the $k$ nearest ones in the archive is deleted, at each stage, thus reducing the memory archive to its desired size. Also, if there are several solutions with minimum distance, the tie is broken by considering the second smallest distance and so forth. 

Some different clustering techniques have also been adopted in MOEAs for maintaining diversity in their memory archives and even in decision variable space, e.g., \citep{inproc.Pulido2004,inproc.Janson2005,journal.Padhye2009}.

\subsection{Crowding Distance}

The crowding distance mechanism was suggested by \cite{journal.Deb2002} to perform diversity preservation in the NSGA-II. This algorithm uses the Pareto dominance to rank the solutions into different fronts (rank based on dominance depth). In each front, the crowding distance of a solution is estimated by calculating the sum of the Euclidean distances among the two neighboring solutions from either side of the solution along the objectives. As illustrated in Fig. \ref{fig:crowding}, the crowding distance of a solution $i$ is given by the sum $f_{1}^{(i)} + f_{2}^{(i)}$.

Variations of the crowding comparison operator have been used by several researchers to maintain diversity in a memory archive, e.g., \citep{inproc.Raquel2005,journal.Wang2010}.

\subsection{Pareto $\eps$-Dominance}

Since the performance of the multiplicative $\eps$-approach proposed by \cite{journal.Laumanns2002} tends to be more sensitive to the geometrical characteristics of a frontier \citep{journal.HernandezDiaz2007}, only the additive scheme will be discussed hereinafter.

Formally, supposing that all objectives $f_{i}$, $i \in \left\{1,\ldots,m\right\}$, are to be minimized, and also that $1 \leq f_{i} \leq K$, for all $i$, then, given a vector $\pmb{y} \in \mathbb{R}^{m}$ and $\eps > 0$, $\pmb{y}$ is said to $\eps$-dominate $\pmb{y}' \in \mathbb{R}^{m}$, denoted as $\pmb{y} \stackrel{\eps}{\prec} \pmb{y}'$, if and only if,
\begin{equation}
y_{i} - \eps \leq y_{i}', \mbox{ for all } i \in \left\{1,\ldots,m\right\}~.
\label{eq:eps_dominance}
\end{equation}

As introduced by \cite{journal.Laumanns2002}, a two-level selection mechanism is implemented in the $\eps$-dominance approach. First, it creates a hypergrid in the objective space where each box uniquely contains one solution vector. Basically, a box-level dominance relation is used, so that the algorithm always maintains a set of nondominated boxes, thus guaranteeing the diversity property. Second, if two vectors share the same box, the usual Pareto dominance relation is applied, so that the best one is selected and convergence is guaranteed. However, if none of these two vectors dominates the other, it is usual to keep the point closest to the origin of the box, i.e., to the corner where all objectives would have the lowest values within that box \citep{proc.Deb2003,journal.Deb2005a}.

The $\eps$-dominance mechanism generates a hypergrid in the objective space with $\left(\left(K-1\right)/\eps\right)^{m}$ boxes which accommodate a maximum of $\left(\left(K-1\right)/\eps\right)^{m-1}$ non $\eps$-dominated points. Then, supposing that the designer wants a maximum of $T$ non $\eps$-dominated points in the archive, the $\eps$ value can be easily calculated as:
\begin{equation}
\eps =  \frac{K-1}{T^{1/(m-1)}}~\cdot
\label{eq:eps_value}
\end{equation}

\begin{figure}[!ht]
\centering
\subfigure[\small Crowding distance assignment.]{
\label{fig:crowding}
\includegraphics[width=5.5cm]{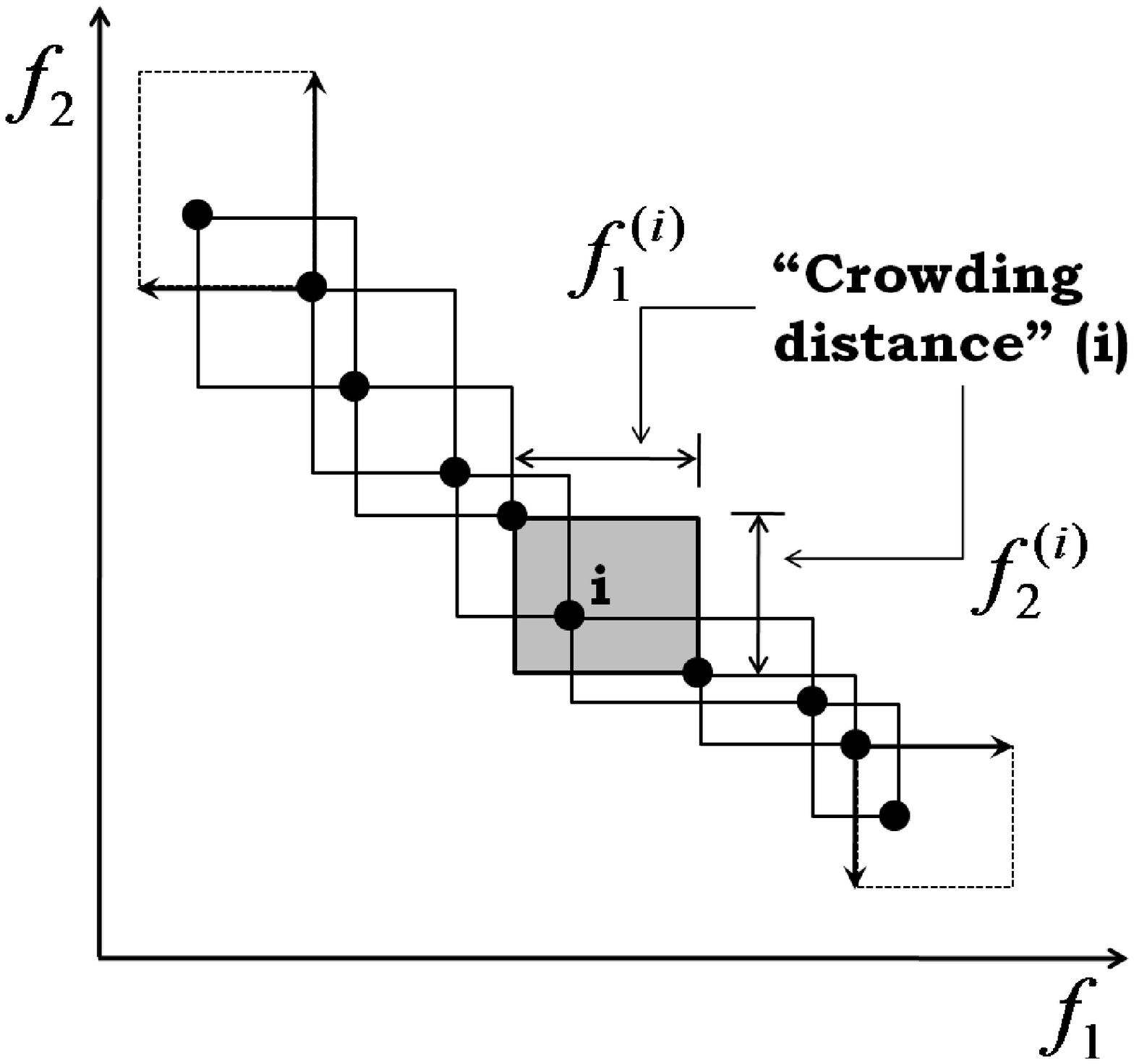}}
\hspace{0.6cm}
\subfigure[\small Diversity access by $\eps$-dominance.]{
\label{fig:eps_grid}
\includegraphics[width=5.5cm]{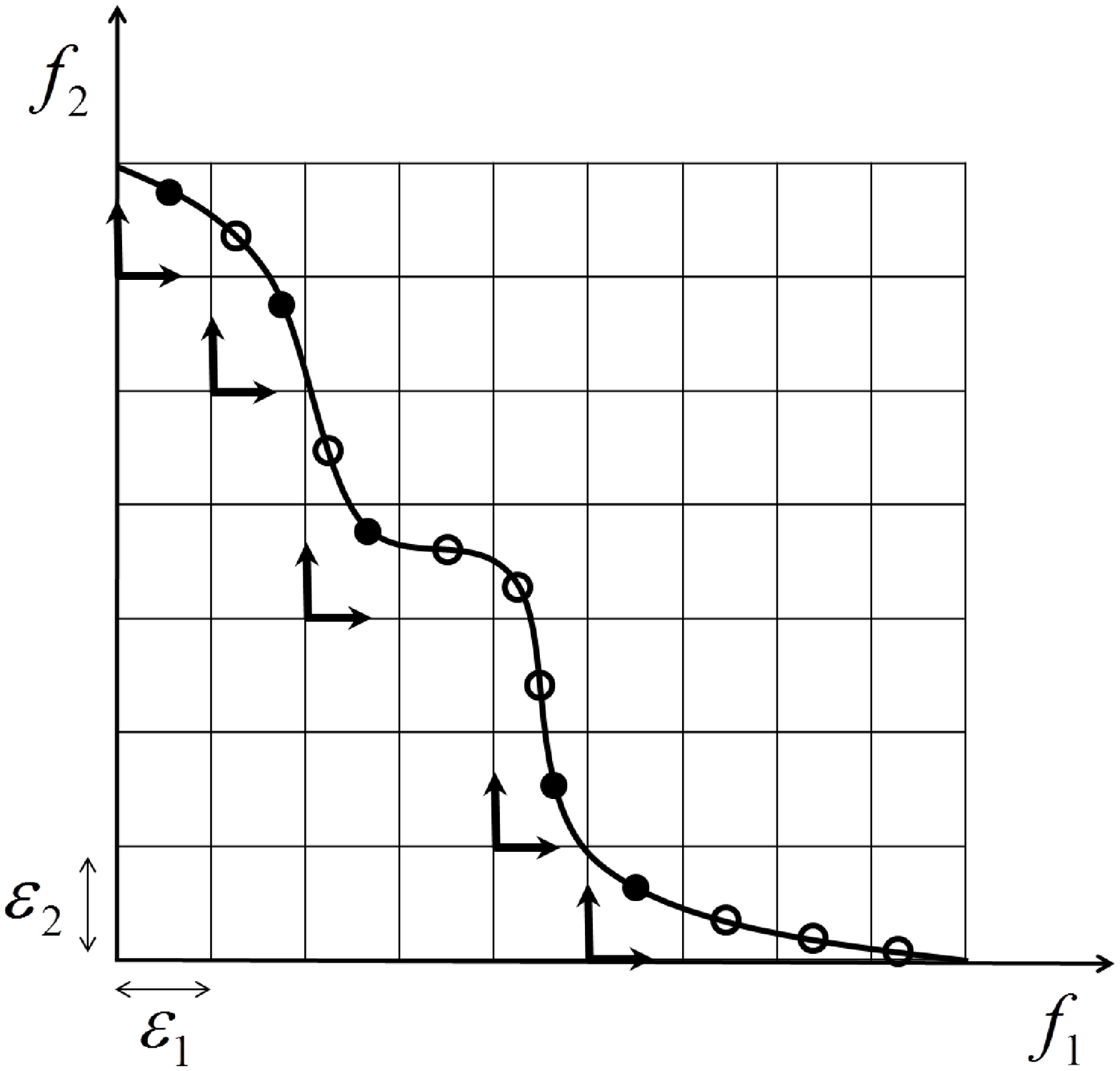}}
\caption{(a) Crowding distance assignment used by the NSGA-II. For example, the crowding measure for the solution $i$ is given by $f_{1}^{(i)} + f_{2}^{(i)}$. (b) Distribution achieved by $\eps$-dominance. It is normal to lose the extreme points of the Pareto front, as well as points located in segments of the front that are almost parallel to the objective axes. As shown, the points (\Circsteel) are $\eps$-Pareto, and the points (\Circpipe), although Pareto-optimal, are $\eps$-dominated. This illustrates the fact that a high number of nondominated solutions can be lost if the decison maker does not take into account, or does not know beforehand, the geometric features of the true Pareto front of the problem to be solved.}
\label{fig:diversity_preservation}
\end{figure}

It should be noted, however, that the $\eps$-dominance strategy is only able to obtain this number $T$ in cases where the Pareto-front is linear. For other cases (see Fig. \ref{fig:eps_grid}) this value is merely an upper limit, with the actual number of nondominated points found being much smaller \citep{journal.HernandezDiaz2007}. Also, it is interesting to note that the given definitions can be generalized by considering a different $\eps$ value for each objective.

\section{The Pareto Cone $\eps$-Dominance Strategy}
\label{sec:4}

\begin{figure}[!ht]
\centering
\subfigure[\small $\eps$-dominance.]{
\label{fig:eps_dominance}
\includegraphics[width=5cm]{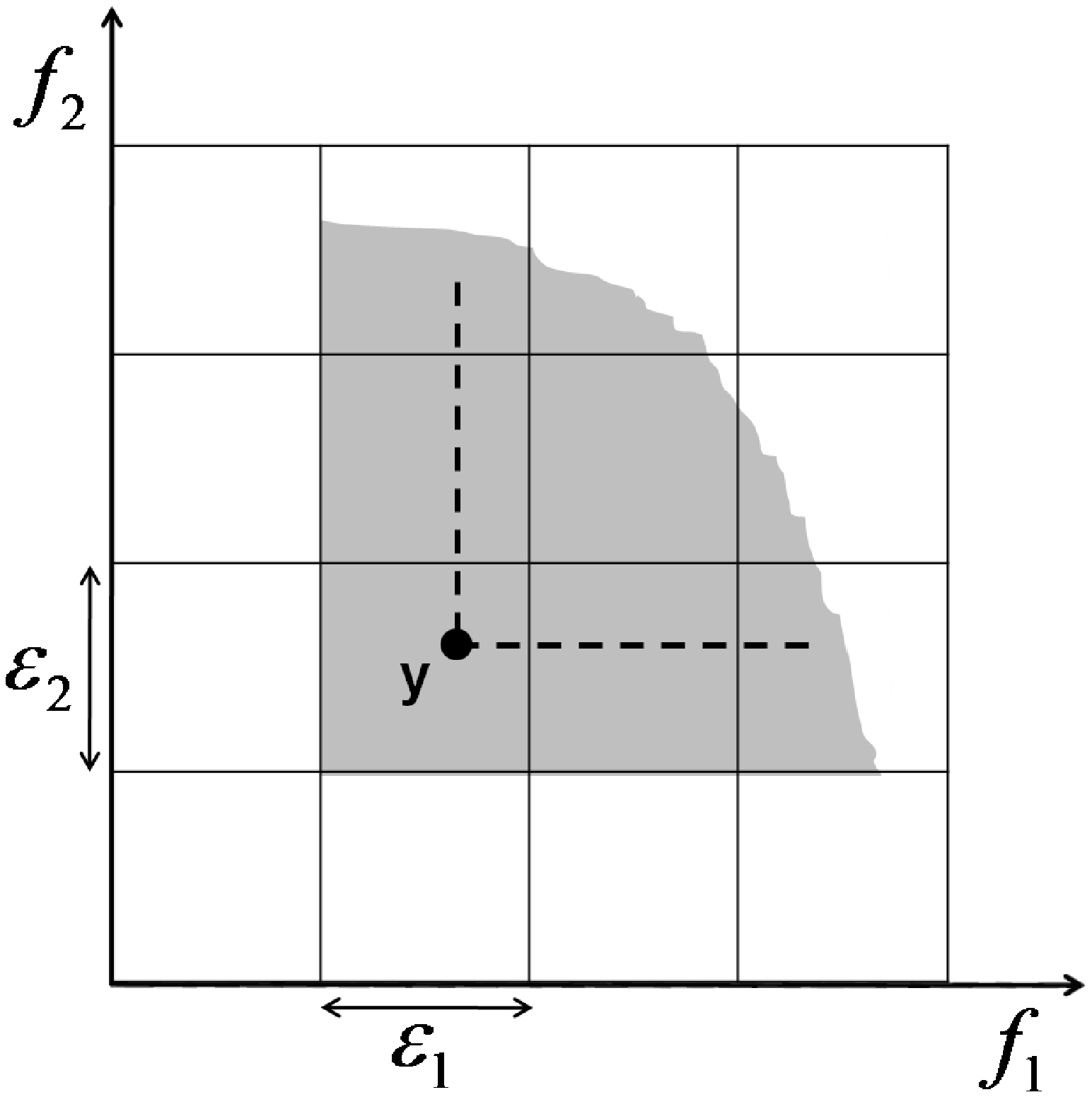}}
\hspace{0.4cm}
\subfigure[\small cone$\eps$-dominance.]{
\label{fig:cone_dominance}
\includegraphics[width=5cm]{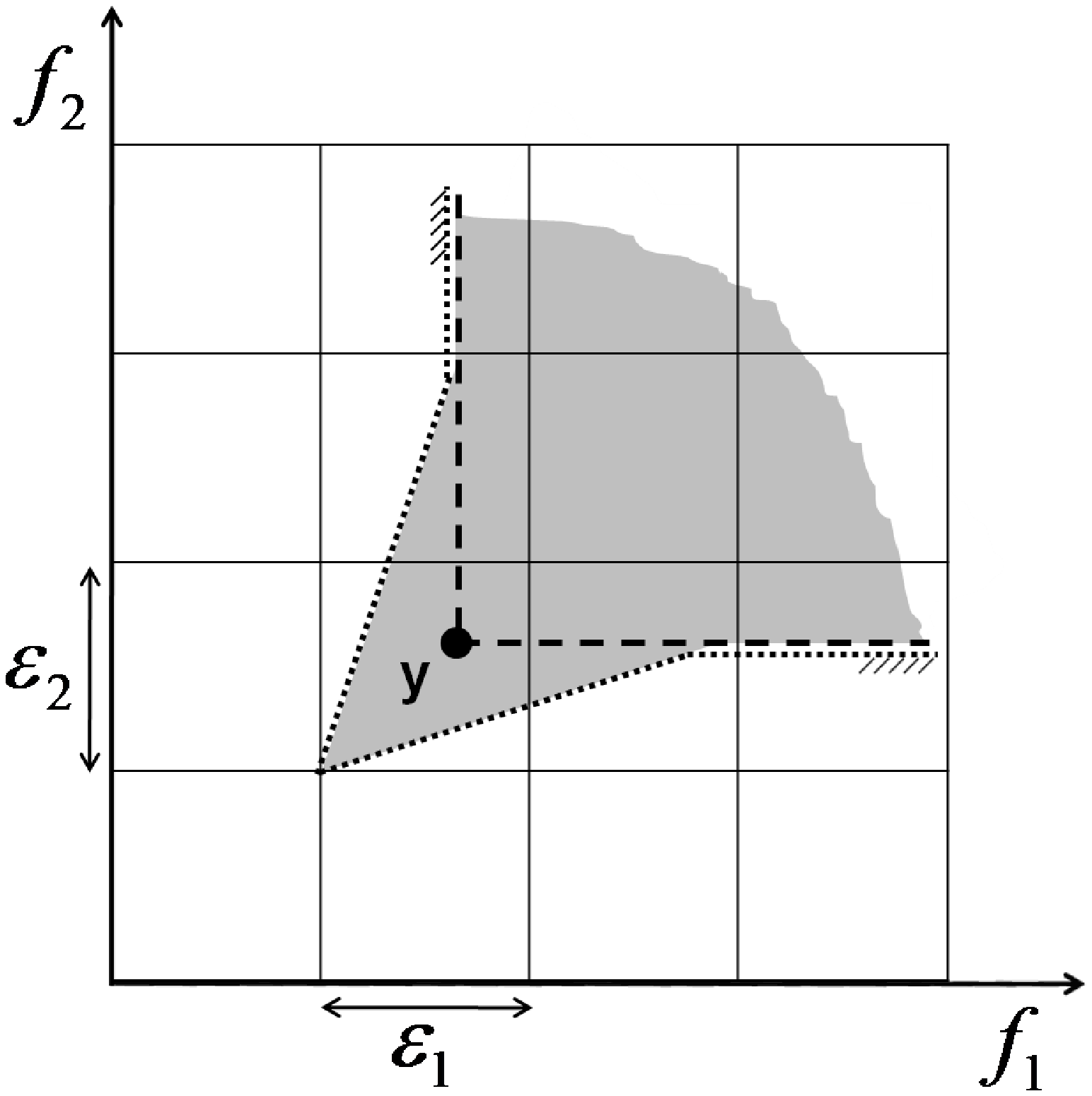}}
\caption{Illustration of the region dominated by a solution $\pmb{y}$ regarding the use of the $\eps$ and cone$\eps$ concepts for a two-objective minimization problem.}
\label{fig:relaxed_dominances}
\end{figure}

Before we present the formal definition of the cone$\eps$-approach, a conceptual interpretation is given \citep{inproc.Batista2011}. Figure \ref{fig:relaxed_dominances} contrasts both the $\eps$-dominance and the cone$\eps$-dominance strategies, and emphasizes the grid generated in the objective space and the different areas dominated by a solution $\pmb{y}$. In order to get a nondominated solution set, the cone$\eps$-dominance mechanism entails both the area dominated by the cone and by the standard Pareto dominance, i.e., the shaded region in Fig. \ref{fig:cone_dominance}. Indeed, the hypervolume dominated by $\pmb{y}$ using the cone$\eps$-dominance approach represents a relaxation of that dominated by $\pmb{y}$ when using the usual dominance. As illustrated, the proposed relaxation enables the approximation of nondominated points in some adjacent boxes that would be $\eps$-dominated. Essentially, since the influence on the ordering of points performed by the cone$\eps$-dominance is limited to a local neighborhood in the objective space, its effects are less dramatic than those of the $\eps$-dominance, which makes it possible to obtain a better distribution of solutions. 

\begin{figure}[!ht]
\centering
\subfigure[\small Diversity by cone$\eps$-dominance.]{
\label{fig:desemp_cone}
\includegraphics[width=5.5cm]{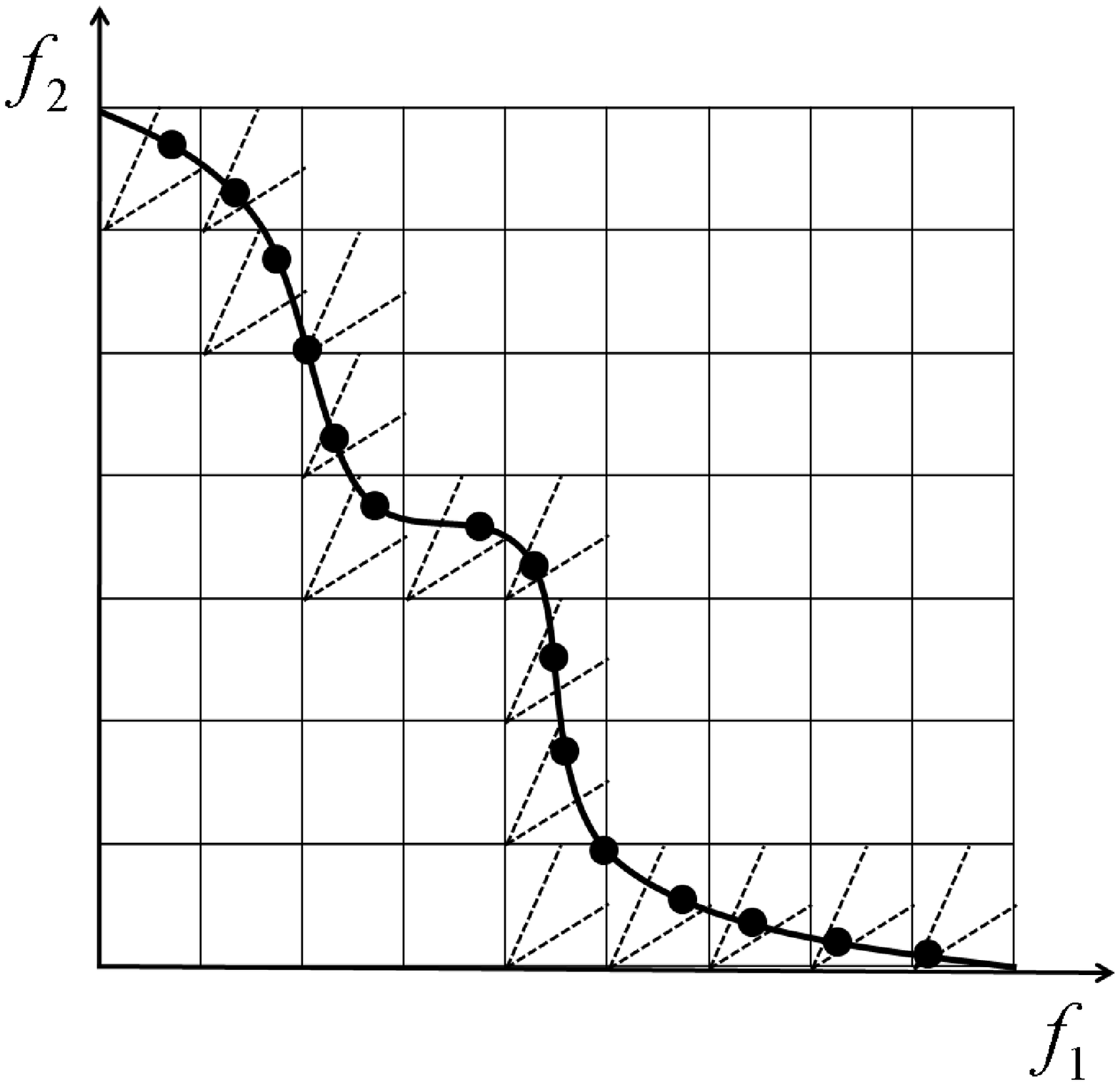}}
\hspace{0.6cm}
\subfigure[\small Diversity by $\eps$-dominance.]{
\label{fig:desemp_epsi}
\includegraphics[width=5.5cm]{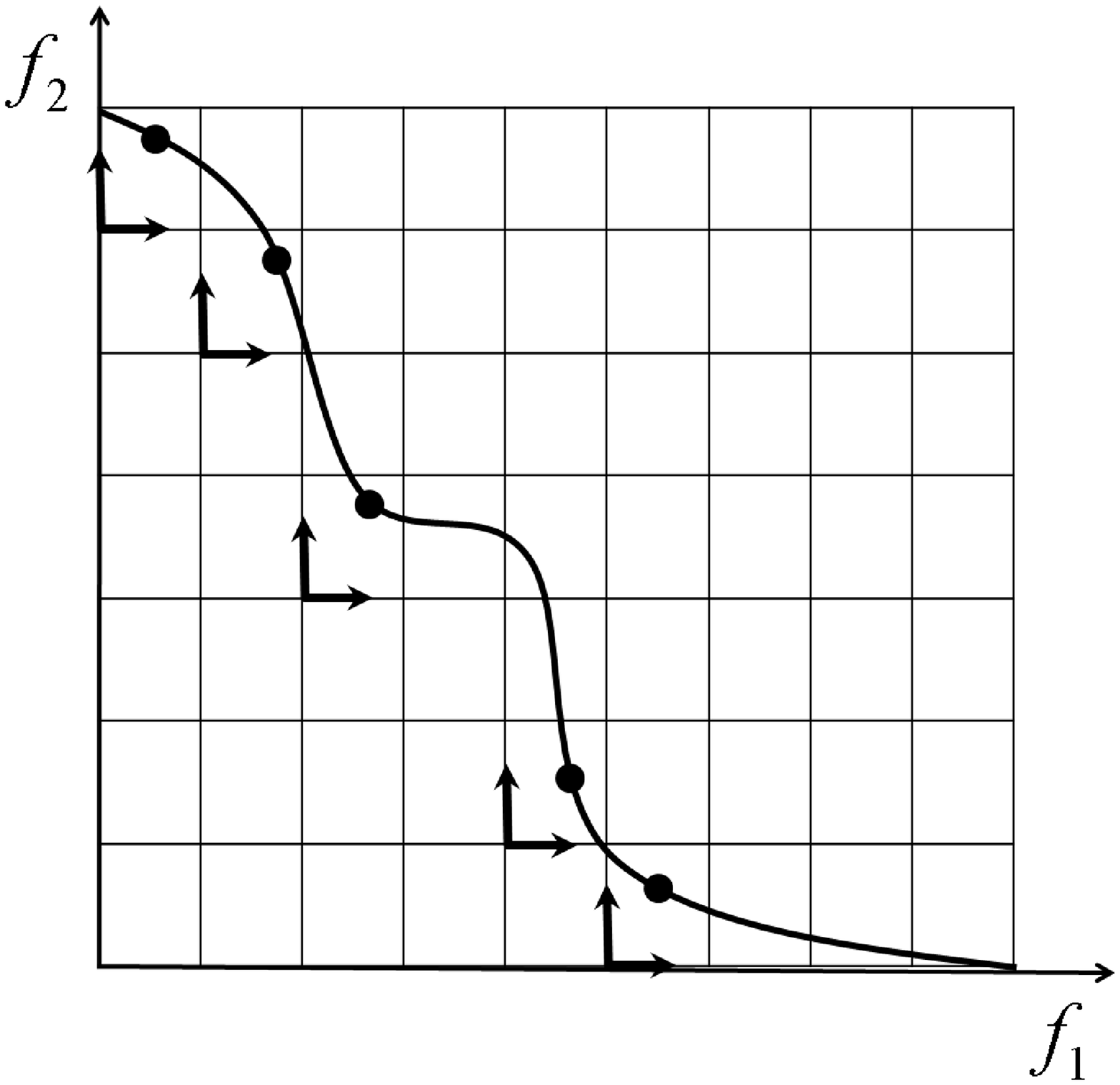}}
\caption{(a) Distribution performed by the cone$\eps$-approach in a connected Pareto front. (b) The $\eps$-relation is very sensitive to the geometry of the frontier and, therefore, inefficient to estimate representative solutions of the Pareto set.}
\label{fig:cone-local-effect}
\end{figure}

Note that the cone$\eps$-approach differs in some aspects from all the mechanisms reviewed in this paper: (i) the selection pressure is ensured by using both the Pareto criterion and the cone of dominance; (ii) the convergence and diversity preservation of the solutions found are simultaneously guaranteed by means of the hypergrid adopted; and (iii) a well-spread nondominated archive is expected due to the local effect of the cone in the objective domain. This last feature is illustrated in the Fig. \ref{fig:cone-local-effect}. 

The cone$\eps$-dominance can also be seen as a hybrid between $\eps$-dominance and the proper efficiency with respect to cones discussed in \citep{book.Miettinen1998}. The use of cones into the MOEA's structure can also be found in \citep{journal.Branke2001,inproc.Shukla2010,inproc.Hirsch2011}.

\subsection{Basic Definitions}

In this section we present the basic definitions needed for the formal presentation of the cone$\eps$-approach \citep{inproc.Batista2011}.\medskip

\noindent\textbf{Definition 1.} \textbf{(Cone)} A set $\mathcal{C}$ is a cone if $\lambda\pmb{y} \in C$ for any $\pmb{y}\in C$ and $\forall~ \lambda \geq0$.$\hfill$ $\Box$\medskip

\noindent\textbf{Definition 2.} \textbf{(Generated cone)} For two vectors $\pmb{w}_{1}$ and $\pmb{w}_{2}$, the cone generated by these vectors  is the set $\mathcal{C} = \left\{\pmb{z}: \pmb{z} = \lambda_{1}\pmb{w}_{1} + \lambda_{2}\pmb{w}_{2}, ~ \forall~ \lambda_{1},\lambda_{2} \geq0\right\}$.$\hfill$ $\Box$\medskip

\noindent Extending the generated cone concept to $m$ dimensions, we have that the hy\-per\-co\-ne generated by the vectors $\pmb{w}_{i}$, for all $i \in \left\{1,\ldots,m\right\}$, is the set $\mathcal{C} = \left\{\pmb{z}: \pmb{z} = \lambda_{1}\pmb{w}_{1} + \ldots + \lambda_{i}\pmb{w}_{i} + \ldots + \lambda_{m}\pmb{w}_{m}, ~ \forall~ \lambda_{i} \geq0\right\}$.

\begin{figure}[!hb]
\centering
\includegraphics[width=6cm]{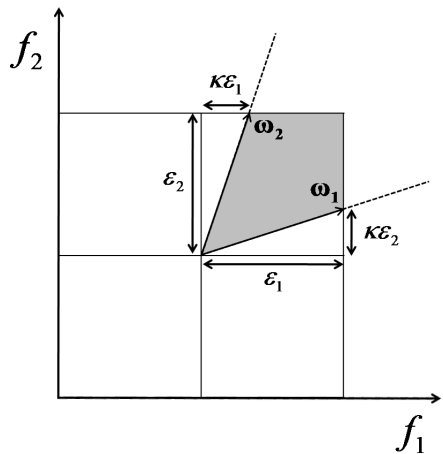}
\caption{Mechanism used in the cone$\eps$-dominance approach to control the hypervolume dominated by a specific cone.}
\label{fig:hbar}
\end{figure}

The definitions above form the basis for a mechanism to control the hypervolume dominated by a specific cone $\mathcal{C}$. For the 2D case (see Fig. \ref{fig:hbar}), it is easy to see that, with respect to the origin of the box, $\pmb{w}_{1} = \left[\eps_{1} ~~ \kappa\eps_{2}\right]^{T}$ and $\pmb{w}_{2} = \left[\kappa\eps_{1} ~~ \eps_{2}\right]^{T}$. The cone $\mathcal{C}$ can, therefore, be rewritten as:
\begin{equation}
\mathcal{C} = \Bigg\{\pmb{z}: \overbrace{\left[\begin{array}{c} z_{1} \\ z_{2} \end{array}\right]}^{\pmb{z}} = \overbrace{\left[\begin{array}{cc} \eps_{1} & \kappa\eps_{1} \\ \kappa\eps_{2} & \eps_{2} \end{array}\right]}^{\Psi} \overbrace{\left[\begin{array}{c} \lambda_{1} \\ \lambda_{2} \end{array}\right]}^{\pmb{\lambda}},~ \forall~ \lambda_{1}, \lambda_{2}\geq0 \Bigg\}
\end{equation}

\noindent in which the parameter $\kappa \in [0,1)$ controls the opening of the cone $\mathcal{C}$, and $\Psi$ is the cone-dominance matrix, which in fact controls the hypervolume dominated by $\mathcal{C}$. Notice that the cone$\eps$-dominance strategy tends toward the $\eps$-dominance strategy when $\kappa \rightarrow 0$, and degenerates into the usual Pareto dominance for $\kappa = 1$.

For the general case with $m$-objectives we have: 
\begin{equation}
\Psi(\eps_{i},\kappa) = \left[\begin{array}{ccccc} 
\eps_{1} & \kappa\eps_{1} & \ldots & \kappa\eps_{1} \\ 
\kappa\eps_{2} & \eps_{2} & \ldots & \kappa\eps_{2} \\ 
\vdots & \vdots & \ddots & \vdots \\
\kappa\eps_{m} & \kappa\eps_{m} & \ldots & \eps_{m}
\end{array}\right]~\cdot
\end{equation}

\noindent which can be used to define the cone$\eps$-dominance strategy.\medskip

\noindent\textbf{Definition 3.} \textbf{(Cone $\eps$-dominance)} Given two vectors $\pmb{y}$, $\pmb{y}' \in \mathbb{R}^{m}$, $\pmb{y}$ is said to cone$\eps$-dominate $\pmb{y}'$ if and only if, $\pmb{y}$ Pareto-dominates $\pmb{y}'$ or the solution of the linear system $\Psi\pmb{\lambda} = \pmb{z}$, with $\pmb{z} = \pmb{y}' - [\pmb{y} - \pmb{\eps}]$, and $\eps_{i} > 0$, gives $\lambda_{i} \geq0 ~\forall~ i \in \left\{1,\ldots,m\right\}$. Equivalently, we say $\pmb{y} \stackrel{cone~\eps}{\prec} \pmb{y}'$ if and only if,
\begin{equation}
\left(\pmb{y} \prec \pmb{y}'\right) ~\vee~ \left(\Psi\pmb{\lambda} = \pmb{z} \mid \lambda_{i} \geq0 \mbox{ for all } i = \left\{1,\ldots,m\right\}\right)~.
\label{eq:cone_dominance}
\end{equation}
$\hfill$ $\Box$

Observe that this in fact represents a relaxed form of Pareto dominance, since the Pareto-dominated region of any given point is a subset of the cone$\eps$-dominated region. Moreover, it is reasonably easy to verify that both the properties of convergence and diversity are satisfied by this criterion. The convergence property is ensured by the Pareto-dominance characteristics of the cone$\eps$ criterion and by the storing of nondominated solutions in the archive population of a MOEA. As for the diversity, since each box accommodates only a single vector, this property is also guaranteed.

\subsection{Maintaining a Cone $\eps$-Pareto Front}

Similarly to the $\eps$-dominance, the archive update function for the cone$\eps$-dominance strategy also employs a two level concept. At the first, a discretization of the objective space into boxes is used, with a single solution within each box. Any algorithm based on the cone$\eps$-dominance relation always maintains a set of nondominated solutions according to this criterion, which ensures the diversity property. To this end, every solution in the archive is assigned a box index $\pmb{b} \in \mathbb{R}^{m}$:
\begin{equation}
\pmb{b}_{i}(\pmb{y}) = \left\{\begin{array}{ll} \eps_{i}\left\lfloor y_{i}/\eps_{i}\right\rfloor & , ~~i \in \left\{1,\ldots,m\right\} , ~~\mbox{for minimizing } f_{i} \smallskip \\
\eps_{i}\left\lceil y_{i}/\eps_{i}\right\rceil & , ~~i \in \left\{1,\ldots,m\right\} , ~~\mbox{for maximizing } f_{i} \end{array}\right.
\label{eq:box}
\end{equation}

\noindent where $\left\lfloor\cdot\right\rfloor$ and $\left\lceil\cdot\right\rceil$ return the closest lower and upper integer to their argument, respectively. 

On the second level, if two vectors share the same box, the former solution is only replaced if it is Pareto dominated within the box, or else by a point closest to the origin of the box, thus guaranteeing convergence. Algorithm \ref{alg:conj_cone_pareto} summarizes these concepts.

\begin{algorithm*}[!htb]
\caption{Archive update function performed by cone$\eps$-dominance.}
\label{alg:conj_cone_pareto}
\KwIn{$\mathcal{H}$, $\Psi$, $\pmb{y}$}
\Begin{
\uIf{$\pmb{y}$ is cone$\eps$-dominated by any $\pmb{y}' \in \mathcal{H}$}{
Reject $\pmb{y}$\;
}
\uElseIf{$\pmb{y}$ shares the same box with an archive member $\pmb{y}'$}{
\uIf{$\pmb{y}$ dominates $\pmb{y}'$ or $\pmb{y}$ is closer to the origin of the box than $\pmb{y}'$}{
Delete all of the cone$\eps$-dominated archive members\;
Replace $\pmb{y}'$ by $\pmb{y}$\;
}
\Else{ 
Reject $\pmb{y}$\;
}}
\uElseIf{$\pmb{y}$ cone$\eps$-dominates any $\pmb{y}' \in \mathcal{H}$}{
Delete all of the cone$\eps$-dominated archive members\;
Insert $\pmb{y}$ into the archive\;
}
\Else{ 
Insert $\pmb{y}$ into the archive\;
}}
\KwOut{$\mathcal{H}'$}
\end{algorithm*}

\subsection{Evaluating the Archive Size}

Assume that all objectives $f_{i}$, $i \in \left\{1,\ldots,m\right\}$, are to be minimized, and also that $1 \leq f_{i} \leq K$, for all $i$. As discussed previously, the cone$\eps$-dominance approach divides the objective space into $\left(\left(K-1\right)/\eps\right)^{m}$ boxes, each allowed to contain a single solution in the archive population $\mathcal{H}$. Since the usual dominance relation ensures a monotonic front between the extreme boxes of the hypergrid, the maximum number of boxes that can be ``touched'' by any front is limited. However, the estimation of feasible solutions inside these touched boxes depends on the connectivity of the Pareto front and on the value of $\kappa$.

\begin{figure*}[!b]
\centering
\subfigure[\small Connected front.]{
\label{fig:sizeA_a}
\includegraphics[width=4.1cm]{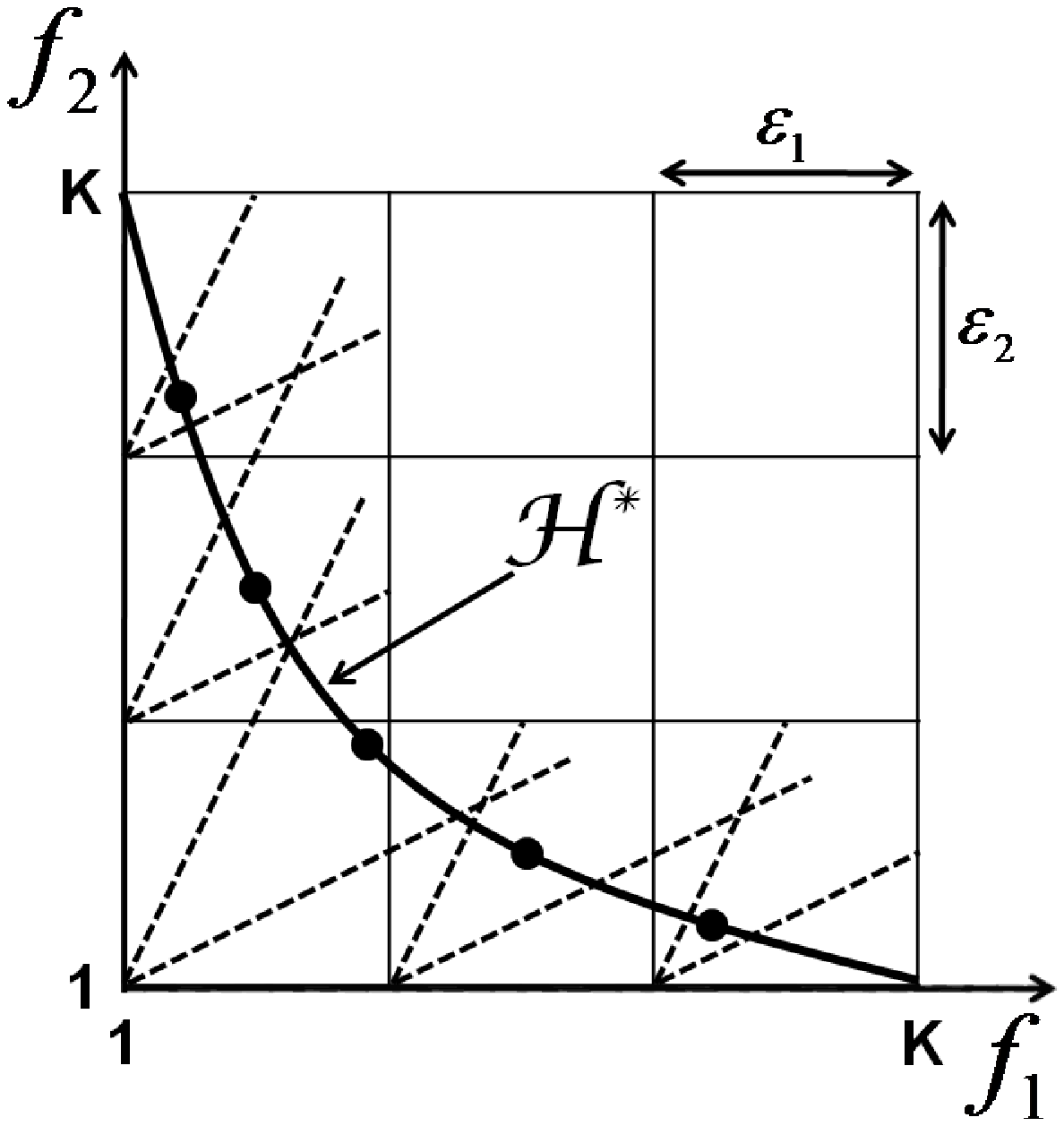}}
\subfigure[\small Concave surface.]{
\label{fig:sizeA3D}
\includegraphics[width=4.6cm]{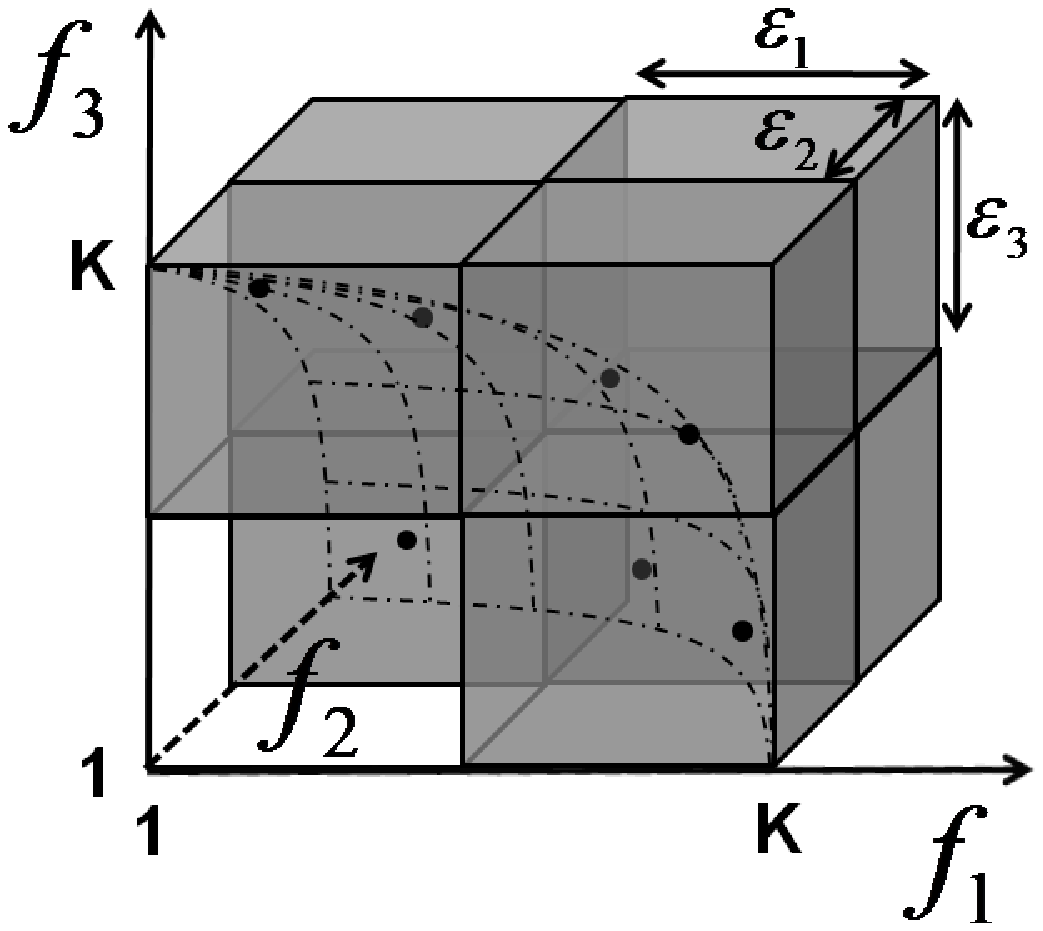}}
\subfigure[\small Disconnected front.]{
\label{fig:sizeA_b}
\includegraphics[width=4.1cm]{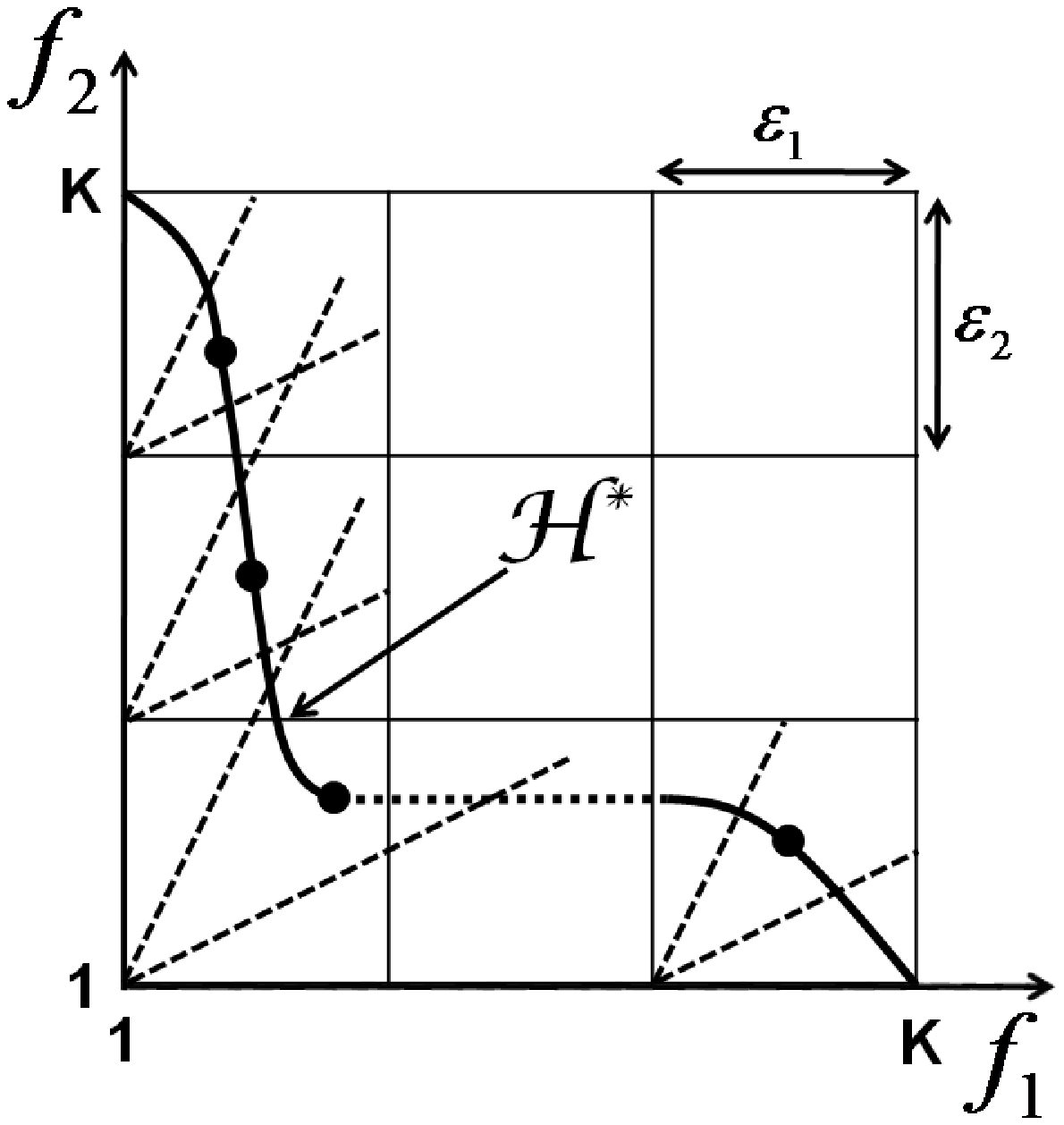}}
\caption{Illustration of the relation between $\eps$ and the size of the archive $\mathcal{H}$ \citep{inproc.Batista2011}.}
\label{fig:sizeA}
\end{figure*}

In general, the number of ``touched'' boxes is maximum if a connected monotonic front exists between the extreme boxes of the hypergrid. Figures \ref{fig:sizeA_a} and \ref{fig:sizeA3D} illustrate two possible situations in which the number of estimated cone$\eps$-Pareto solutions is maximum, i.e., equal to the number of boxes that are touched by the front which is equal to five and seven, respectively. These values are calculated by a simple relation governed by the number of objectives, the range of the objective values, and the values of $\eps_i$:
\begin{equation}
\left|\mathcal{H}\right| \leq m\left[\left(\prod\limits_{i=1}^{m-1} \frac{K-1}{\eps_i}\right) - \left(\prod\limits_{i=1}^{m-2} \frac{K-1}{\eps_i}\right)\right] + 1~.
\label{eq:sizeA_former}
\end{equation}

It must be emphasized, however, that the maximum size of $\mathcal{H}$ cannot be reached for a problem where the Pareto front is disconnected. Nevertheless, the cone$\eps$-dominance relation is still likely to estimate one solution for each box touched by the front.

In the particular case where $\eps_i = \eps$, $\forall i \in \left\{1,\ldots,m\right\}$, the previous expression can be rewritten as:
\begin{equation}
\left|\mathcal{H}\right| \leq m\left[\left(\frac{K-1}{\eps}\right)^{m-1} - \left(\frac{K-1}{\eps}\right)^{m-2}\right] + 1~.
\label{eq:sizeA}
\end{equation}

Figure \ref{fig:sizeA_b} illustrates a problem where the Pareto front is disconnected, which means that the maximum size of $\mathcal{H}$ cannot be reached. However, the cone$\eps$-dominance relation is still likely to estimate one solution for each box touched by the front. 

The $\eps$ value that provides a maximum of $T$ nondominated points in the archive, according to the cone$\eps$-criterion, can be easily calculated from \eqref{eq:sizeA} by solving \eqref{eq:coneps}:
\begin{equation}
\left(\frac{T-1}{m}\right)\eps^{(m-1)} + \left(K-1\right)^{m-2}\eps - \left(K-1\right)^{m-1} = 0~.
\label{eq:coneps}
\end{equation}

Finally, observe that specific bounds on the objective values are not used in the Alg. \ref{alg:conj_cone_pareto} and are not required to ensure the convergence. They are only employed to demonstrate the relation between $\eps$ and the size of the archive $\mathcal{H}$.

\section{Experimental Design and Validation of the Proposed Approach}
\label{sec:5}

To validate the proposed cone$\eps$-dominance approach, six algorithms are considered for the experimental study: the first two are the well-known NSGA-II \citep{journal.Deb2002} and SPEA2 \citep{rep.Zitzler2001} methods, both based on the standard Pareto dominance relation; the third is a modified NSGA-II, in which the crowding distance is re-calculated after each point removal in order to improve diversity; the fourth is a clustering version of the NSGA-II, in which the crowding distance procedure is replaced with a clustering approach; the fifth is a steady-state MOEA based on the $\eps$-dominance strategy; and the sixth is implemented by modifying the last one, replacing the $\eps$-dominance mechanism with the cone$\eps$-dominance approach. The comparison of the two last methods will enable us to show the performance of the same algorithm with and without cone$\eps$-dominance. Furthermore, this study will contrast the ability of the cone$\eps$-criterion in maintaining diversity against operators which handle diversity efficiently, e.g., the clustering operator \citep{journal.Zitzler1999} used in the C-NSGA-II and the truncation method of SPEA2 \citep{rep.Zitzler2001}. 

The multiobjective evolutionary algorithms are briefly described next:
\begin{enumerate}
	\item \textbf{NSGA-II:} This algorithm was proposed by \cite{journal.Deb2002}. In general terms, the parent and offspring populations are combined and evaluated using the fast nondominated sorting approach, an elitist approach, and an efficient crowding mechanism to implement density estimation. When the memory archive set is greater than its capability, only those solutions that are maximally apart from their neighbors, according to the crowding measure, are maintained. In the original NSGA-II the crowding measure is calculated only once, and this information is used to remove a specific number of solutions with the smallest crowding value. Considering a population with $N$ points, the computational complexity is $\mathcal{O}\left(mN^2\right)$ for the nondominated sort, $\mathcal{O}\left(mN\log N\right)$ for the crowding distance assignment, which makes the algorithm computationally fast, and $\mathcal{O}\left(N\log N\right)$ for the crowded comparison operator. In this way, the overall complexity of the NSGA-II is dominated by the nondominated sorting procedure, i.e., $\mathcal{O}\left(mN^2\right)$. Though the NSGA-II has been shown to perform quite well for two or three-objective problems, its crowding operator is not adequate in maintaining a good distribution of solutions in a problem having many objective functions \citep{journal.Deb2005a}. 	
	\item \textbf{SPEA2:} The SPEA2 \citep{rep.Zitzler2001} presents some differences with respect to its predecessor, SPEA \citep{journal.Zitzler1999}: (i) it employs a fine-grained fitness assignment strategy; (ii) it uses a nearest neighbor density estimation technique which guides the search more efficiently; (iii) the archive size is fixed; (iv) it has an enhanced archive truncation method that guarantees the preservation of boundary solutions; and (v) only members of the archive participate in the mating selection process. The computational complexity of the fitness assignment procedure is controlled by the density estimator, $\mathcal{O}\left(mN^2\log N\right)$, but the worst run-time complexity of the SPEA2 is dominated by the truncation operator, that is $\mathcal{O}\left(mN^3\right)$. However, on average the complexity will be $\mathcal{O}\left(mN^2\log N\right)$ since the sorting of the distances governs the overall complexity \citep{rep.Zitzler2001}.
	\item \textbf{NSGA-II*:} Instead of using the standard NSGA-II's crowding distance mechanism, the NSGA-II* performs an improved diversity preservation approach by re-calculating the crowding distance after each point removal. In this situation, the worst-case complexity of the crowding distance assignment is $\mathcal{O}\left(m(2N)\log (2N)\right)$. This means that the overall complexity of the NSGA-II* is $\mathcal{O}\left(mN^2\right)$, which is due to the nondominated sorting procedure, as observed for the NSGA-II. However, a better distributed set of solutions is supposed to be achieved with the NSGA-II*.
	\item \textbf{C-NSGA-II:} This method was discussed in \citep{proc.Deb2003}, and is a straightforward replacement of the NSGA-II's crowding mechanism with the clustering approach used in SPEA \citep{journal.Zitzler1999}, which has a computational complexity of $\mathcal{O}\left(mN^3\right)$, involving Euclidean distance calculations. In spite of the larger computational time required by this method, the clustered NSGA-II is expected to find a better distributed set of nondominated solutions than the original NSGA-II. 
	\item \textbf{$\eps$-MOEA:} This approach was proposed by \cite{proc.Deb2003,journal.Deb2005a}, and consists of a steady-state MOEA based on the $\eps$-dominance concept introduced in \cite{journal.Laumanns2002}. In this method, two populations, evolutionary population and archive population, are evolved simultaneously, and two offspring solutions are created by using one solution from each population. Each offspring is then used to update both parent and archive populations in a steady-state manner, thereby providing better chances of creating good offspring solutions. Note, however, that the archive population is updated based on the $\eps$-dominance concept, whereas an usual domination concept is used to update the parent population. Since a fitness assignment method is not implemented in the $\eps$-MOEA, its complexity includes only the operations of maintaining and truncating the memory population, being these techniques handled together in the archive update process. The computational complexity of the $\eps$-MOEA, for the entire population, requires comparisons of $\mathcal{O}\left(mN^2\right)$ in the worst case. However, it is important to note that a computational time of $\mathcal{O}\left(mN^2\right)$ is required in addition due to the simultaneous evolution of two independent populations \citep{journal.Li2011}. A careful observation reveals that the $\eps$-MOEA procedure emphasizes nondominated solutions, maintains the diversity in the archive by allowing only one solution inside each hyperbox on the Pareto-optimal front, and represents an elitist approach \citep{proc.Deb2003}.
	\item \textbf{cone$\eps$-MOEA:} This algorithm is a modification of the $\eps$-MOEA, in which we include cone$\eps$-dominance instead of the regular $\eps$-dominance concept. Due to the small differences regarding the archive update strategies of these methods, a similar discussion can be made for the time complexity of the cone$\eps$-MOEA. In this way, considering one updating operation of a single offspring (see Algorithm \ref{alg:conj_cone_pareto}), the complexity order of total comparisons required is $\mathcal{O}\left(mN\right)$ to check the Pareto dominance and $\mathcal{O}\left(m^2N\right)$ to check the cone of dominance, in which the $m^2$ time cost refers to a linear system solution by factorization. However, since no further comparisons are required if the candidate solution is dominated at the beginning of the process, the average time complexity gets lower, and the best case run-time required is $\mathcal{O}\left(m\right)$. In a straightforward manner, the computational complexity of the cone$\eps$-MOEA, for the entire population, requires a run-time of $\mathcal{O}\left((mN)^2\right)$ in the worst case.	
\end{enumerate}

Based on the previous discussion, and without considering the effect of special data structures, it is possible to sort the algorithms with respect to computational complexity. In this way, the sequence of run-time efficiency from the lowest to highest is: C-NSGA-II, SPEA2, cone$\eps$-MOEA, NSGA-II*, NSGA-II, and $\eps$-MOEA. These methods have been implemented in Matlab, and some of them are based on the source codes available in \citep{KanGAL}\footnote{The implementation of the algorithms, the samples of the true Pareto fronts used, and the raw and processed results of the experiments can be retrieved from \citep{Code}.}. Further information about test problems, performance metrics, parameter settings, and statistical design are presented in the following sections.

\subsection{Benchmark Test Problems}

In a previous work \citep{inproc.Batista2011}, we performed a limited simulation study in which the choice of problems was directed by the geometrical characteristics of the Pareto fronts rather than the difficulty of solving each test problem. Now, in order to carry out a more appropriate analysis, we consider more complex test problems, each providing a different kind of challenge for multiobjective evolutionary algorithms. First, we have chosen two continuous and simple tests, Deb52 \citep{journal.Deb1999} and Poloni's problem \citep{proc.Poloni1995}, both employed in the previous work \citep{inproc.Batista2011}. The five two-objective ZDT problems were also selected \citep{journal.Zitzler2000}. Since this paper deals only with continuous-parameter problems, the ZDT5 problem was not considered as it is defined for binary strings. At last, we consider nine three-objective DTLZ test problems \citep{proc.Deb2005}. In this study, the constraints were handled using a simple penalty method ($f_{i} + 1000 \sum_{j=1}^{n_{g}} \max\left(0,g_{j}\right)$) for all the algorithms. Further details about these benchmark problems are presented in Table \ref{tab:problems} and throughout the paper.

\begin{table*}
\scriptsize
\renewcommand{\arraystretch}{1.3}
\centering
\caption{Analytical test problems adopted in the experimental study.}
\label{tab:problems}
    \begin{math}
    \begin{array}{l|c|c|l|l}
    \hline
    \mbox{Problem} & $n$ & \mbox{Bounds} & \mbox{Objective functions} & \mbox{Pareto front} \\
    \hline
    \mbox{P1: Deb52} & 2 & [0,1] & f_{1}(\pmb{x})=1-\exp(-4x_{1})\sin^{4}(10\pi x_{1}) & \mbox{Concave} \\
         &    &      & f_{2}(\pmb{x})=g(x_{2})h(x_{1}) \mbox{ ; } g(x_{2})=1+x_{2}^2	 & 								\\         
         &    &      & h(x_{1})= \left\{\begin{array}{ll} 
         1-\left(\frac{f_{1}(\pmb{x})}{g(x_{2})}\right)^{10} & \mbox{if } f_{1}(\pmb{x})\leq g(x_{2}) 	\\ 
         0 & \mbox{otherwise.} \\ 
         \end{array} \right. & \\
    \hline
    \mbox{P2: Pol} & 2 & [-\pi,\pi] & f_{1}(\pmb{x})=1+(A_{1}-B_{1})^2+(A_{2}-B_{2})^2 & \mbox{Nonconvex and} \\
	       &    &    & f_{2}(\pmb{x})=(x_{1}+3)^2+(x_{2}+1)^2                        		 & \mbox{disconnected}  \\
 		     &    &    & A_{1}=0.5\sin1-2\cos1+\sin2-1.5\cos2                          		 &                      \\
 	  	   &    &    & A_{2}=1.5\sin1-\cos1+2\sin2-0.5\cos2                          		 &                      \\
	       &    &    & B_{1}=0.5\sin x_{1}-2\cos x_{1}+\sin x_{2}-1.5\cos x_{2}      		 &                      \\
	       &    &    & B_{2}=1.5\sin x_{1}-\cos x_{1}+2\sin x_{2}-0.5\cos x_{2}      		 &                      \\   	    
    \hline
    \mbox{P3: ZDT1}  & 30 & [0,1] & f_{1}(\pmb{x})=x_{1} \mbox{ ; } f_{2}(\pmb{x})=1-\sqrt{x_{1}/g(\pmb{x})}  & \mbox{Convex}   \\         
         &    &      & g(\pmb{x})=1+9\left(\sum_{i=2}^{n}x_{i}\right)/(n-1)       			 											&     						\\
    \hline
    \mbox{P4: ZDT2}  & 30 & [0,1] & f_{1}(\pmb{x})=x_{1} \mbox{ ; } f_{2}(\pmb{x})=1-\left(x_{1}/g(\pmb{x})\right)^2  & \mbox{Nonconvex}  \\         
         &    &      & g(\pmb{x})=1+9\left(\sum_{i=2}^{n}x_{i}\right)/(n-1)       			 															&      							\\
    \hline
    \mbox{P5: ZDT3}  & 30 & [0,1] & f_{1}(\pmb{x})=x_{1} \mbox{ ; } f_{2}(\pmb{x})=1-\sqrt{h}-h\sin\left(10\pi x_{1}\right) & \mbox{Disconnected}  \\         
         &    &      & g(\pmb{x})=1+9\left(\sum_{i=2}^{n}x_{i}\right)/(n-1) \mbox{ and } h=x_{1}/g(\pmb{x})									& \mbox{convex parts}  \\
    \hline
    \mbox{P6: ZDT4}  & 10       & x_{1}\in[0,1] & f_{1}(\pmb{x})=x_{1}                                 														 & \mbox{Nonconvex}   \\
         &    & x_{i}\in[-5,5], & f_{2}(\pmb{x})=1-\sqrt{x_{1}/g(\pmb{x})}                          															 & \mbox{(multimodal} \\
         &    & i=2,...,n       & g(\pmb{x})=1+10\left(n-1\right)+\sum_{i=2}^{n}\left(x_{i}^2-10\cos\left(4\pi x_{i}\right)\right) & \mbox{problem)}    \\ 
    \hline
    \mbox{P7: ZDT6}  & 10 & [0,1] & f_{1}(\pmb{x})=1-\exp(-4x_{1})\sin^{6}(6\pi x_{1})                       	& \mbox{Nonconvex}     \\
         &    &      & f_{2}(\pmb{x})=1-(f_{1}/g(\pmb{x}))^2                                            			& \mbox{(nonuniform}   \\
         &    &      & g(\pmb{x})=1+9\left[\left(\sum_{i=2}^{n} x_{i}\right)/(n-1)\right]^{0.25} 							& \mbox{search space)} \\
    \hline
    \mbox{P8: DTLZ1}  & 7 & [0,1] & f_{1}(\pmb{x})=0.5x_{1}x_{2}(1+g(\pmb{x}_{m}))                     							& \mbox{Linear} 			\\
         &    &       & f_{2}(\pmb{x})=0.5x_{1}(1-x_{2})(1+g(\pmb{x}_{m}))            															& \mbox{hyperplane}	\\
         &    &       & f_{3}(\pmb{x})=0.5(1-x_{1})(1+g(\pmb{x}_{m}))                            										& \\
         &    &       & g(\pmb{x}_{m})=100\left(\left|\pmb{x}_{m}\right| + \sum_{x_{i}\in \pmb{x}_{m}} h_{i}\right) & \\
         &    &       & h_{i}=\left(x_{i}-0.5\right)^{2}-\cos(20\pi(x_i-0.5))																			  & \\    
    \hline
    \mbox{P9: DTLZ2}  & n & [0,1] & f_{1}(\pmb{x})=(1+g(\pmb{x}_{m}))\prod_{i=1}^{m-1}\cos(x_{i}^\alpha\pi/2)    					 							& \mbox{Concave surface} \\
         &    &       & f_{j}(\pmb{x})=(1+g(\pmb{x}_{m}))\left(\prod_{i=1}^{m-j}\cos(x_{i}^\alpha\pi/2)\right)\sin(x_{M}^\alpha\pi/2) 	& \\
         &    &       & f_{m}(\pmb{x})=(1+g(\pmb{x}_{m}))\sin(x_{1}^\alpha\pi/2) \mbox{ ; } M=m-j+1 																		& \\
         &    &       & g(\pmb{x}_{m})=\sum_{x_{i}\in \pmb{x}_{m}} \left(x_{i}-0.5\right)^{2} \mbox{ ; } \alpha=1       								& \\
         &    &       & n=m+k-1 \mbox{ , } k=10 \mbox{ and } j=2,\ldots,m-1																															& \\     
    \hline
    \mbox{P10: DTLZ3} & 12 & [0,1] & f_{1} \mbox{, } f_{2} \mbox{ and } f_{3} \mbox{ as in DTLZ2}    									& \mbox{Concave surface} \\
         &    &       & g(\pmb{x}_{m})=100\left(\left|\pmb{x}_{m}\right| + \sum_{x_{i}\in \pmb{x}_{m}} h_{i}\right) 	& \\
         &    &       & h_{i}=\left(x_{i}-0.5\right)^{2}-\cos(20\pi(x_i-0.5))																			  	& \\
    \hline
    \mbox{P11: DTLZ4} & 12 & [0,1] & f_{1} \mbox{, } f_{2} \mbox{ and } f_{3} \mbox{ as in DTLZ2 with } \alpha=100 		& \mbox{Concave surface} \\    
    \hline
    \mbox{P12: DTLZ5} & 12 & [0,1] & f_{1}(\pmb{x})=(1+g(\pmb{x}_{m}))\cos(\theta_{1})\cos(\theta_{2}) 						& \mbox{Concave curve} \\
         &    &       & f_{2}(\pmb{x})=(1+g(\pmb{x}_{m}))\cos(\theta_{1})\sin(\theta_{2}) 												& \\
         &    &       & f_{3}(\pmb{x})=(1+g(\pmb{x}_{m}))\sin(\theta_{1})                      										& \\
         &    &       & \theta_{1}=x_1\pi/2 \mbox{ and } \theta_{2}=\frac{\pi}{4(1+g(r))}(1+2g(r)x_{2})						& \\  
         &    &       & g(\pmb{x}_{m})=\sum_{x_{i}\in \pmb{x}_{m}} \left(x_{i}-0.5\right)^2												& \\ 
    \hline
    \mbox{P13: DTLZ6} & 12 & [0,1] & f_{1} \mbox{, } f_{2} \mbox{ and } f_{3} \mbox{ as in DTLZ5} 								& \mbox{Concave curve} \\         
         &    &       & g(\pmb{x}_{m})=\sum_{x_{i}\in \pmb{x}_{m}} x_{i}^{0.1}    																& \\ 
    \hline
    \mbox{P14: DTLZ7} & 22 & [0,1] & f_{1}(\pmb{x})=x_{1} \mbox{ ; } f_{2}(\pmb{x})=x_{2} \mbox{ ; } f_{3}(\pmb{x})=(1+g(\pmb{x}_{m}))h		& \mbox{Disconnected}    \\
         &    &       & h=3 - \sum_{i=1}^{2}\left[\frac{f_{i}}{1+g}\left(1+\sin(3\pi f_{i})\right)\right] 																& \mbox{(noncontiguous}  \\
         &    &       & g(\pmb{x}_{m})=1+\frac{9}{20}\sum_{x_{i}\in \pmb{x}_{m}} x_{i}																										& \mbox{convex regions)} \\ 
    \hline
    \mbox{P15: DTLZ8} & 30 & [0,1] & f_{i}(\pmb{x})=0.1\sum_{j=10(i-1)+1}^{10j}x_{i} \mbox{ ; } i=1,2,3			      & \mbox{Combination of}   \\
         &    &       & g_{i}(\pmb{x})=f_{3}(\pmb{x})+4f_{i}(\pmb{x})-1\geq0 \mbox{ ; } i=1,2											& \mbox{a straight line}  \\
         &    &       & g_{3}(\pmb{x})=2f_{3}(\pmb{x})+f_{1}(\pmb{x})+f_{2}(\pmb{x})-1\geq0    										& \mbox{and a hyperplane} \\                 
    \hline
    \mbox{P16: DTLZ9} & 30 & [0,1] & f_{i}(\pmb{x})=\sum_{j=10(i-1)+1}^{10j}x_{i}^{0.1} \mbox{ ; } i=1,2,3			      & \mbox{Concave curve}  \\
         &    &       & g_{i}(\pmb{x})=f_{3}^{2}(\pmb{x})+f_{i}^{2}(\pmb{x})-1\geq0 \mbox{ ; } i=1,2									&   										\\
	  \hline    
    \end{array}
    \end{math}
\end{table*}

\subsection{Performance Metrics}

Evolutionary multiobjective optimization techniques are required to consider two different goals, i.e., beyond acquiring convergence to the Pareto-optimal front, an equally important task is to find and maintain a diverse set of solutions. In addition, 
achieving these goals in a small computational time is also an important issue in MOEAs. To consider this multi-criterion nature in the evaluation of multi-objective algorithms, regarding the convergence and diversity of the solutions found, we have used four different metrics.

\begin{figure}[!thb]
\centering
\subfigure[\small Convergence metric ($\gamma$).]{\includegraphics[width=5cm]{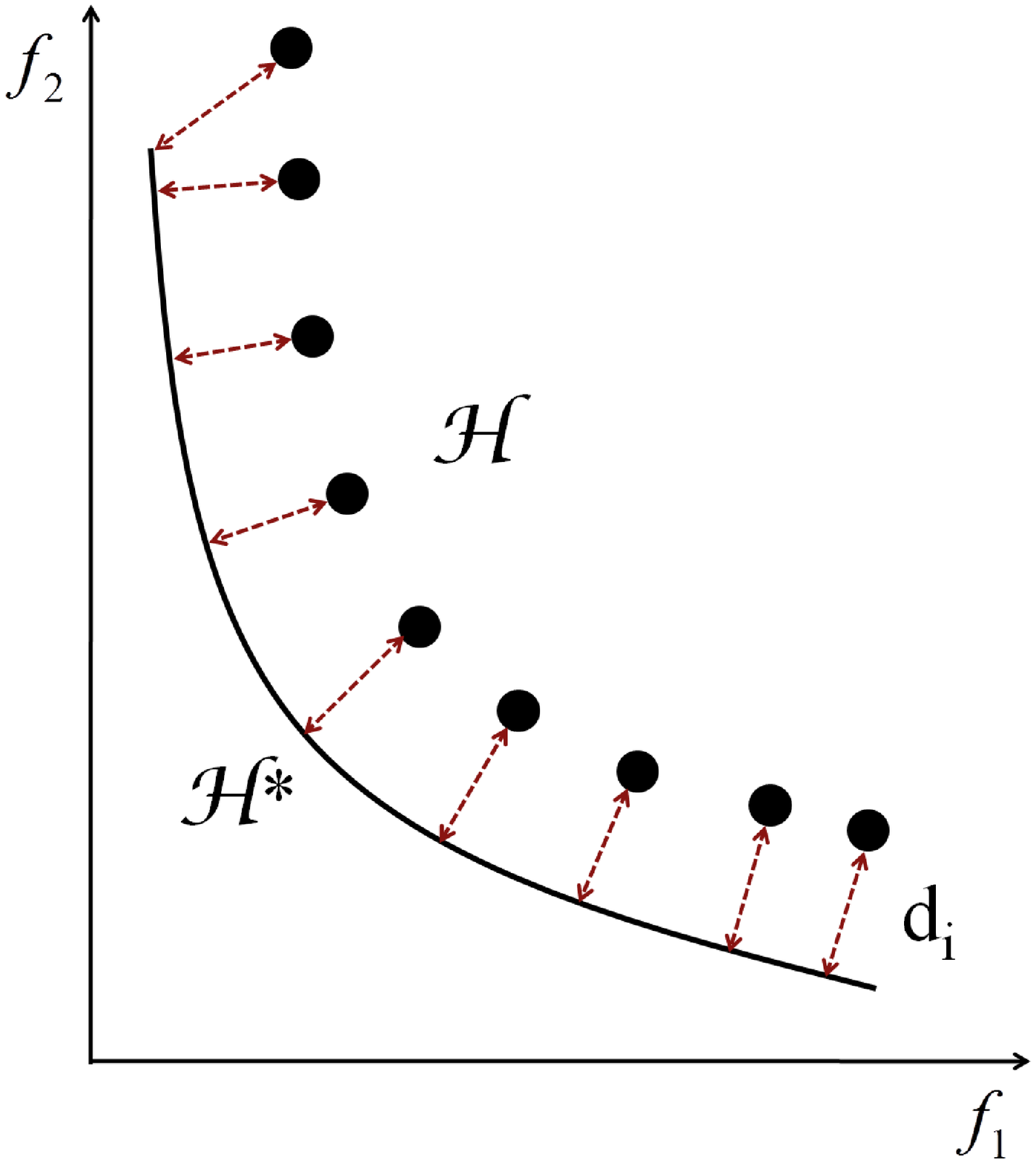}\label{fig:gamma}}\hspace{0.6cm}
\subfigure[\small Diversity metric ($\Delta$).]{\includegraphics[width=5cm]{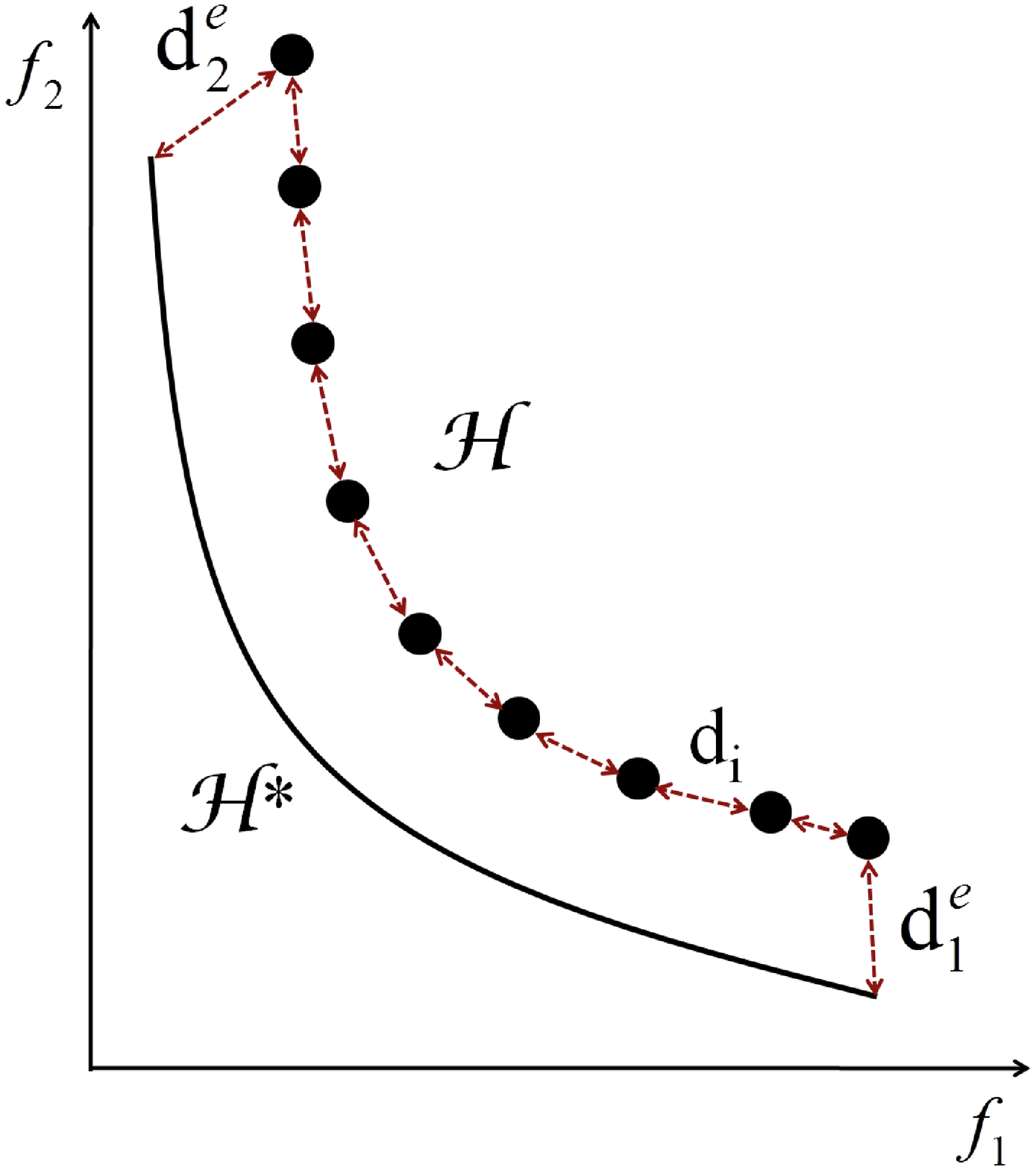}\label{fig:delta}}
\subfigure[\small S-metric (HV).]{\includegraphics[width=5cm]{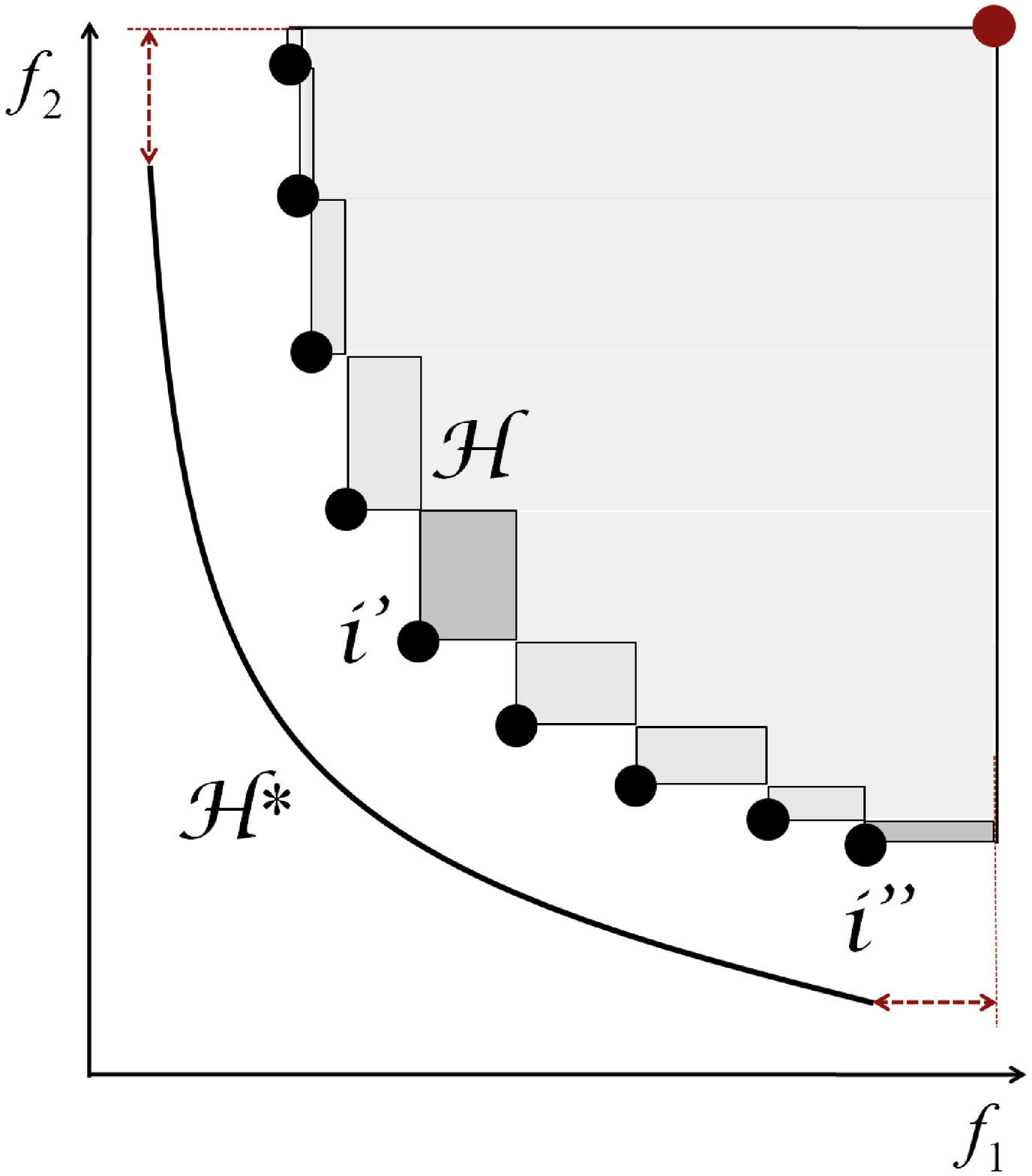}\label{fig:smetric}}\hspace{0.6cm}
\subfigure[\small Coverage of many sets ($CS$).]{\includegraphics[width=5cm]{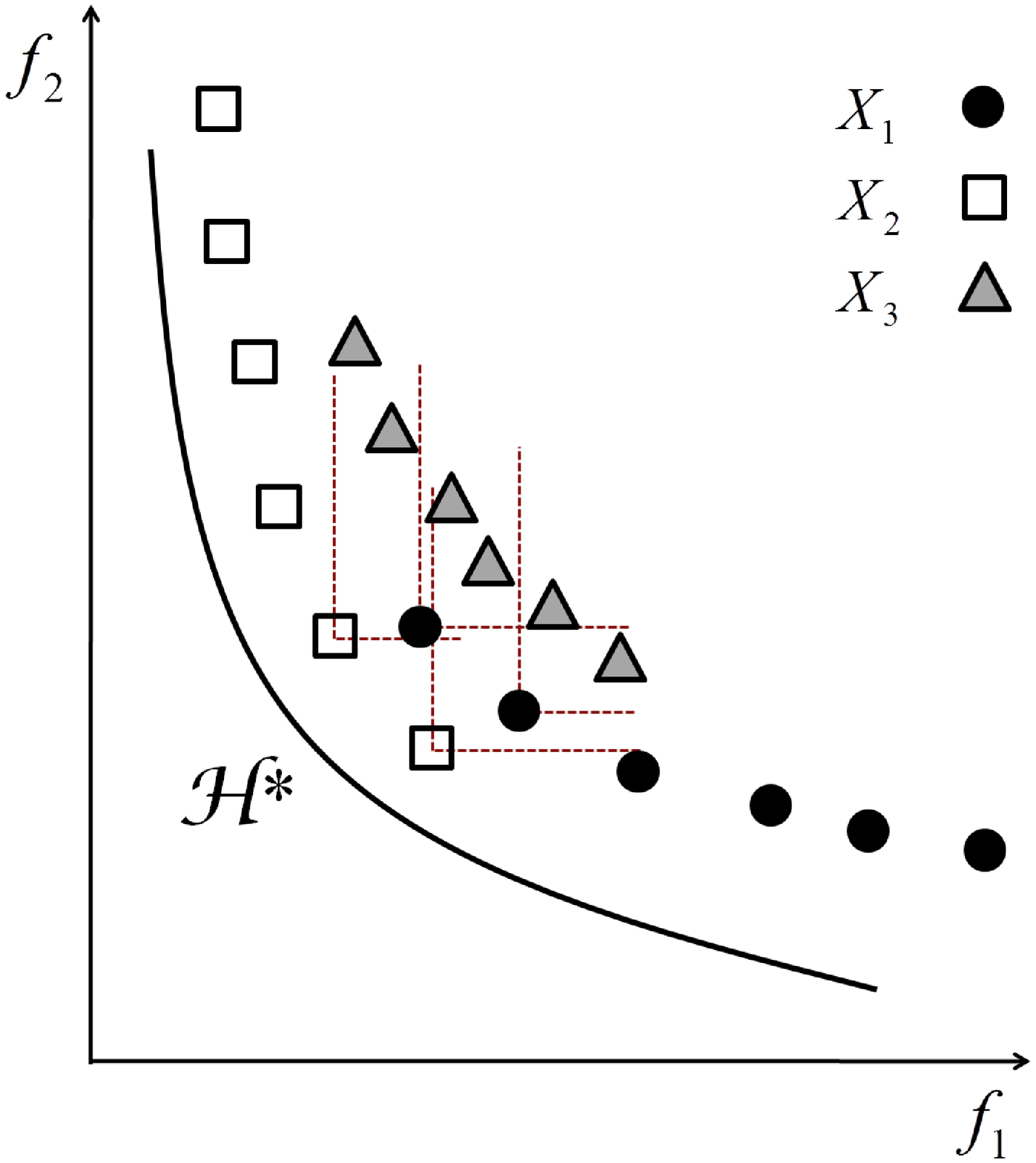}\label{fig:csmetric}}
\caption{Quality indicators adopted in the experimental study. In these illustrations, the sets $\mathcal{H}^*$ and $\mathcal{H}$ represent the global Pareto front and the estimated one, respectively. According to letter (c), the contribution of the solution $i'$ for the $HV$ measure is greater than that of the solution $i''$. In (d), the coverage of many sets measures are $CS\left(X_{1},U_{1}\right)=4/12$, $CS\left(X_{2},U_{2}\right)=8/12$, and $CS\left(X_{3},U_{3}\right)=0$, indicating the superior quality of the front $X_2$, followed by $X_1$ and $X_3$.}
\label{fig:metrics}
\end{figure}

In order to assess how near the solutions found are from the Pareto-optimal front, the convergence metric $(\gamma)$, proposed by \cite{journal.Deb2002}, was considered. This quality indicator, see Fig. \ref{fig:gamma}, measures the distance between the obtained nondominated front $\mathcal{H}$ and a detailed sampling of the true Pareto-optimal front $\mathcal{H}^*$:
\begin{equation}
\gamma = \frac{\sum_{i=1}^{\left|\mathcal{H}\right|} d_{i}}{\left|\mathcal{H}\right|}
\label{eq:gamma}
\end{equation}

\noindent where $d_{i}$ is the Euclidean distance, in the objective space, between the solution $i\in \mathcal{H}$ and the nearest member of $\mathcal{H}^*$, and the operator $\left|~\cdot~\right|$ returns the cardinality of the set in its argument. So, the lower the $\gamma$ value the better the convergence of the solutions in $\mathcal{H}$. As illustrated in the Fig. \ref{fig:gamma}, a result with $\gamma = 0$ means $\mathcal{H} \subseteq \mathcal{H}^*$, in which all the estimated solutions are Pareto-optimal.  

Because MOEAs are required to achieve an interesting trade-off between the convergence and distribution of the solutions approximated, the diversity metric $(\Delta)$ \citep{journal.Deb2002} was also used. This quality indicator measures the extent of spread achieved among the obtained nondominated solutions in $\mathcal{H}$. Considering that it is desirable to obtain a set of solutions that spans the entire Pareto-optimal region, the $\Delta$ value is defined as:
\begin{equation}
\Delta = \frac{\sum_{i=1}^{m} d_{i}^{e} + \sum_{i=1}^{\left|\mathcal{H}\right|} \left|d_{i}-\bar{d}\right|}{\sum_{i=1}^{m} d_{i}^{e} + \left|\mathcal{H}\right|\bar{d}}
\label{eq:delta}
\end{equation}

\noindent where $d_{i}^{e}$ denotes the Euclidean distance between the extreme points in $\mathcal{H}$ and $\mathcal{H}^*$ along the $i^{th}$ coordinate, and $d_{i}$ measures the Euclidean distance of each point in $\mathcal{H}$ to its closest neighbor. Therefore, a lower $\Delta$ value indicates a better distribution of solutions, as can be seen from Fig. \ref{fig:delta}. Notice that a result with $\Delta = 0$ means the extreme points of $\mathcal{H}^*$ have been found and $d_{i}$ equals to $\bar{d}$ for all $i$.

The third metric, known as S-metric or Hypervolume $(HV)$ metric \citep{journal.Zitzler1999}, calculates the hypervolume enclosed by the estimated front $\mathcal{H}$ and a reference point $\pmb{y}_{ref}$ dominated by all solutions in this front. Regarding a minimization MOP, the larger the dominated hypervolume, the better the front is (see the Fig. \ref{fig:smetric}). For all test problems, the reference point was defined as $10\%$ greater than the upper boundaries of the real Pareto-optimal front. Formally, this metric is described as the Lebesgue measure $\Lambda$ of the union of hypercubes $h_i$ defined by a nondominated point $\pmb{y}_i$ and $\pmb{y}_{ref}$:
\begin{equation}
HV(\mathcal{H}) = \Lambda\left(\left\{\bigcup\limits_i h_i \mid \pmb{y}_i \in \mathcal{H}\right\}\right) = \Lambda\left(\bigcup\limits_{\pmb{y} \in \mathcal{H}} \left\{\pmb{y}' \mid \pmb{y} \prec \pmb{y}' \prec \pmb{y}_{ref}\right\}\right)
\label{eq:hv}
\end{equation}

\noindent Even though this metric estimates both convergence and diversity of the solutions in $\mathcal{H}$, it is more sensitive to the convergence of the points towards the Pareto-optimal front, and it prefers convex regions to non-convex ones \citep{inproc.Zitzler2007}. For instance, as shown in Fig. \ref{fig:smetric}, the contribution of the solution $i'$ to the $HV$ measure is greater than that of the solution $i''$.

In order to evaluate the convergence of each algorithm in contrast to all others, a generalization of the Coverage of Two Sets metric \citep{journal.Zitzler1999} is proposed here. This metric, called here the Coverage of Many Sets $(CS)$, quantifies the domination of the final population of one algorithm over the union of the remaining ones. The proposed $CS$ function is stated as:
\begin{equation}
CS(X_i,U_i) = \frac{\left|a''\in U_i;~ \exists~ a'\in X_i:~ a'\preceq a''\right|}{\left|U_i\right|}
\label{eq:cs}
\end{equation}

\noindent where $X_i$ and $U_i$ are sets of objective vectors, and $a'\preceq a''$ means that $a'$ covers $a''$, that is, either $a'\prec a''$ or $a' = a''$. Function $CS$ maps the pair $\left(X_{i},U_{i}\right)$ to the interval $[0,1]$, in which $X_{i}$, for all $i = 1,\ldots,k$, denotes the final Pareto front resulting from algorithm $i$, and $U_{i}$, defined as:
\begin{equation}
U_{i} = \bigcup\limits_{\substack{j=1\\j\neq i}}^{k} X_j
\label{eq:cs2}
\end{equation}

\noindent represents the union of the final Pareto fronts of all the $k$ algorithms, except $i$. The value $CS\left(X_{i},U_{i}\right) = 1$ implies that all points in $U_{i}$ are dominated by or equal to points in $X_{i}$. The opposite, $CS\left(X_{i},U_{i}\right) = 0$, represents the situation when none of the points in $U_{i}$ are covered by the set $X_{i}$. Note that for $k=2$, the $CS$ metric tends to the Coverage of Two Sets proposed in \citep{journal.Zitzler1999}. Although the measures $CS\left(X_{i},U_{i}\right)$ and $CS\left(U_{i},X_{i}\right)$ are not complementary, by simplicity this study considers only the comparisons $CS\left(X_{i},U_{i}\right)$, for all $i = 1,\ldots,k$. Figure \ref{fig:csmetric} shows the $CS$ measures for a particular $k=3$ case. As illustrated, $CS\left(X_{1},U_{1}\right)=4/12$, $CS\left(X_{2},U_{2}\right)=8/12$, and $CS\left(X_{3},U_{3}\right)=0$, indicating the superior quality of the front $X_2$, followed by $X_1$ and $X_3$, respectively.

For the first two metrics, a detailed sampling of the true Pareto-optimal front of each problem must be known. Since we are dealing with test problems, the true Pareto-optimal front is not difficult to obtain. In this work, we have used uniformly spaced Pareto-optimal solutions as the approximation of the true Pareto-optimal front. The reference fronts used here can be retrieved online \citep{Code}.

\subsection{Parameter Settings}

With the objective of providing a comparison baseline for the performance of the proposed cone$\eps$-dominance strategy, some parameter settings were adopted for all the algorithms: population size $N = 100$, and probabilities of crossover and mutation $p_{xover} = 1$ and $p_{mut} = 1/n$, respectively. Since all test problems involve real-valued decision variables, we have used the simulated binary crossover (SBX) operator \citep{journal.Deb1995} and the polynomial mutation operator \citep{journal.Deb1996} to create the offspring solutions. The crossover ($\eta_{xover}$) and mutation ($\eta_{mut}$) distribution indices adopted for each problem are shown in Table \ref{tab:parameters}. Furthermore, for a fair comparison, all algorithms are iterated for a fixed number of solution evaluations: 100,000 evaluations for ZDT4 and DTLZ8; 50,000 for DTLZ9; 30,000 for DTLZ3 and DTLZ4; and 20,000 for the remaining ones.

\begin{table}
\small
\renewcommand{\arraystretch}{1.0}
\centering
\caption{Crossover and mutation distribution indices adopted for the algorithms.}
\label{tab:parameters} 
    \begin{tabular}{c|c|c|c}    
    \hline
         & \mbox{Deb52, Pol, ZDT1, ZDT3} & \mbox{ZDT4, DTLZ1,} 					& \mbox{ZDT2,}     \\
         & \mbox{ZDT6, DTLZ2, DTLZ4} & \mbox{DTLZ3} 					& \mbox{DTLZ5--DTLZ9}     \\
    \hline            
    $\eta_{xover}$ & 15 & 2  & 2  \\
    $\eta_{mut}$   & 20 & 20 & 5  \\             
    \hline      
    \end{tabular}
\end{table}

\begin{table*}
\renewcommand{\arraystretch}{1.05}
\centering
\caption{Calculated (first line) and estimated (second line) $\pmb{\eps}$ values for the algorithms $\underline{\pmb{\eps}}$-MOEA and $\underline{\pmb{c}}$one$\eps$-MOEA. These values have been obtained in order to get roughly 100 solutions in the archive at the end of the run. The sign ``--'' means that it was not necessary to estimate values for $\pmb{\eps}$. In these cases, the specific calculation method has provided adequate values.}
\label{tab:eps_values} 
		\resizebox{\textwidth}{!}{ %
    \begin{tabular}{c|c|c|c|c|c|c|c}
    \multicolumn{8}{c}{Two-Objective Test Problems} \\
    \hline
    \mbox{Alg.} 	& \mbox{Deb52} & \mbox{Pol} & \mbox{ZDT1} & \mbox{ZDT2} & \mbox{ZDT3} & \mbox{ZDT4} & \mbox{ZDT6} \\		    								  
    \hline            
    $\pmb{\eps}$ 	& [0.0083,0.010] & [0.16,0.25]    & $\eps_{i}$ = 0.010  & $\eps_{i}$ = 0.010  & [0.0085,0.018]      & $\eps_{i}$ = 0.010  & [0.0072,0.0093]  	  \\	
    				 			& [0.003,0.003]  & [0.038,0.038]  & $\eps_{i}$ = 0.0075 & $\eps_{i}$ = 0.0076 & $\eps_{i}$ = 0.0026 & $\eps_{i}$ = 0.0065 & $\eps_{i}$ = 0.0067 \\      \hline     
    $\pmb{c}$  & [0.0164,0.0198] & [0.3168,0.4950] & $\eps_{i}$ = 0.0198 & $\eps_{i}$ = 0.0198 & [0.0168,0.0356]     & $\eps_{i}$ = 0.0198 & [0.0143,0.0184] \\    
    				 	 & --              & [0.20,0.33]     & --                  & --                  & [0.012,0.025]       & --                  & --              \\	
    \hline      
    \end{tabular}}
    \resizebox{0.8\textwidth}{!}{ %
    \begin{tabular}{c|c|c|c|c|c} 
    \multicolumn{6}{c}{Three-Objective Test Problems} \\
    \hline
    \mbox{Alg.} & \mbox{DTLZ1} & \mbox{DTLZ2} & \mbox{DTLZ3} & \mbox{DTLZ4} & \mbox{DTLZ5} \\		    								  
    \hline            
    $\pmb{\eps}$	& $\eps_{i}$ = 0.05 	& $\eps_{i}$ = 0.10   & $\eps_{i}$ = 0.10   & $\eps_{i}$ = 0.10   & [0.007,0.007,0.01] 		\\   				  
    				 			& [0.02,0.02,0.05]    & [0.06,0.06,0.066]   & [0.06,0.06,0.066]   & $\eps_{i}$ = 0.062  & $\eps_{i}$ = 0.005   	\\          
    \hline     
    $\pmb{c}$			& $\eps_{i}$ = 0.0798 & $\eps_{i}$ = 0.1595 & $\eps_{i}$ = 0.1595 & $\eps_{i}$ = 0.1595 & [0.014,0.014,0.02]  	\\    
    				 			& [0.05,0.05,0.0833]  & --                  & --                  & --                  & $\eps_{i}$ = 0.025  	\\    
    \hline      
    \end{tabular}} 
    \resizebox{0.7\textwidth}{!}{ %
    \begin{tabular}{c|c|c|c|c} 
    \multicolumn{5}{c}{} \\
    \hline
    \mbox{Alg.} & \mbox{DTLZ6} & \mbox{DTLZ7} & \mbox{DTLZ8} & \mbox{DTLZ9} \\		    								  
    \hline            
    $\pmb{\eps}$	& [0.007,0.007,0.01] & [0.086,0.086,0.3386] 	& [0.075,0.075,0.10]	& $\eps_{i}$ = 0.010 	\\   				  
    				 			& $\eps_{i}$ = 0.005 & $\eps_{i}$ = 0.05 			& [0.02,0.02,0.04]   	& $\eps_{i}$ = 0.025 	\\          
    \hline     
    $\pmb{c}$			& [0.014,0.014,0.02] & [0.1372,0.1372,0.5404]	& [0.12,0.12,0.16]  	& $\eps_{i}$ = 0.0198 \\    
    				 			& $\eps_{i}$ = 0.017 & [0.12,0.12,0.30]				& $\eps_{i}$ = 0.03 	& [0.27,0.27,0.25] 		\\    
    \hline      
    \end{tabular}}   
\end{table*}

In order to obtain a final archive population size of $100$ solutions, we have used the $\pmb{\eps}$ values showed in Table \ref{tab:eps_values}. The calculated $\pmb{\eps}$ values have been obtained following the guidelines provided by \cite{journal.Laumanns2002} for the $\eps$-MOEA \eqref{eq:eps_value}, and by \cite{inproc.Batista2011} for the cone$\eps$-MOEA \eqref{eq:coneps}. Since both $\eps$-dominance and cone$\eps$-dominance may lose nondominated points, we have employed estimated $\pmb{\eps}$ values to get roughly $100$ solutions in the final archive. For the $\eps$-approach, the estimated values were obtained from \citep{proc.Deb2003,journal.Deb2005a}, whereas for the cone$\eps$-approach the estimation was performed by testing different $\pmb{\eps}$ values to get roughly $100$ solutions in the archive at the end of the run.

\begin{figure}[thb]
\centering
\psfrag{f1}[][]{$f_{1}$}
\psfrag{f2}[][]{$f_{2}$}
\psfrag{f3}[][]{$f_{3}$}
\psfrag{Fronteira Pareto Global}[][]{\scriptsize Pareto front}
\psfrag{cone-MOEA}[][]{\scriptsize ~~~~~~~~~~~~~~~~~~cone$\eps$-MOEA}
\psfrag{e-MOEA}[][]{\scriptsize ~~~~~~~~~~~~~~$\eps$-MOEA}
\subfigure[\small cone$\eps$-MOEA distribution on test Deb52: \newline $\pmb{\eps}=(0.0535,0.0645)$, $\left|\mathcal{H}\right|=31$, $T=30$.]{\includegraphics[width=6.6cm]{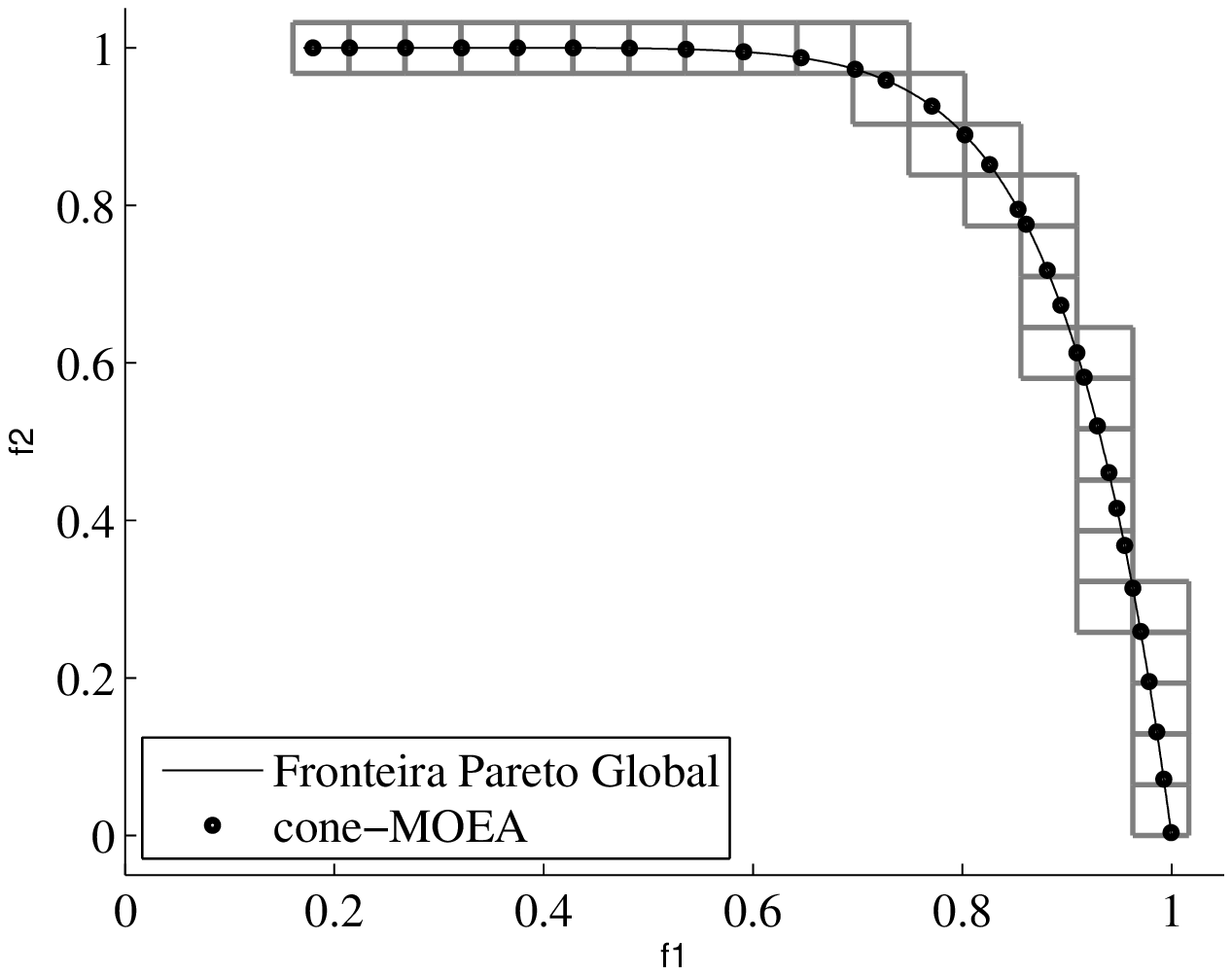}}
\subfigure[\small $\eps$-MOEA distribution on test Deb52: \newline $\pmb{\eps}=(0.0277,0.0333)$, $\left|\mathcal{H}\right|=12$, $T=30$.]{\includegraphics[width=6.6cm]{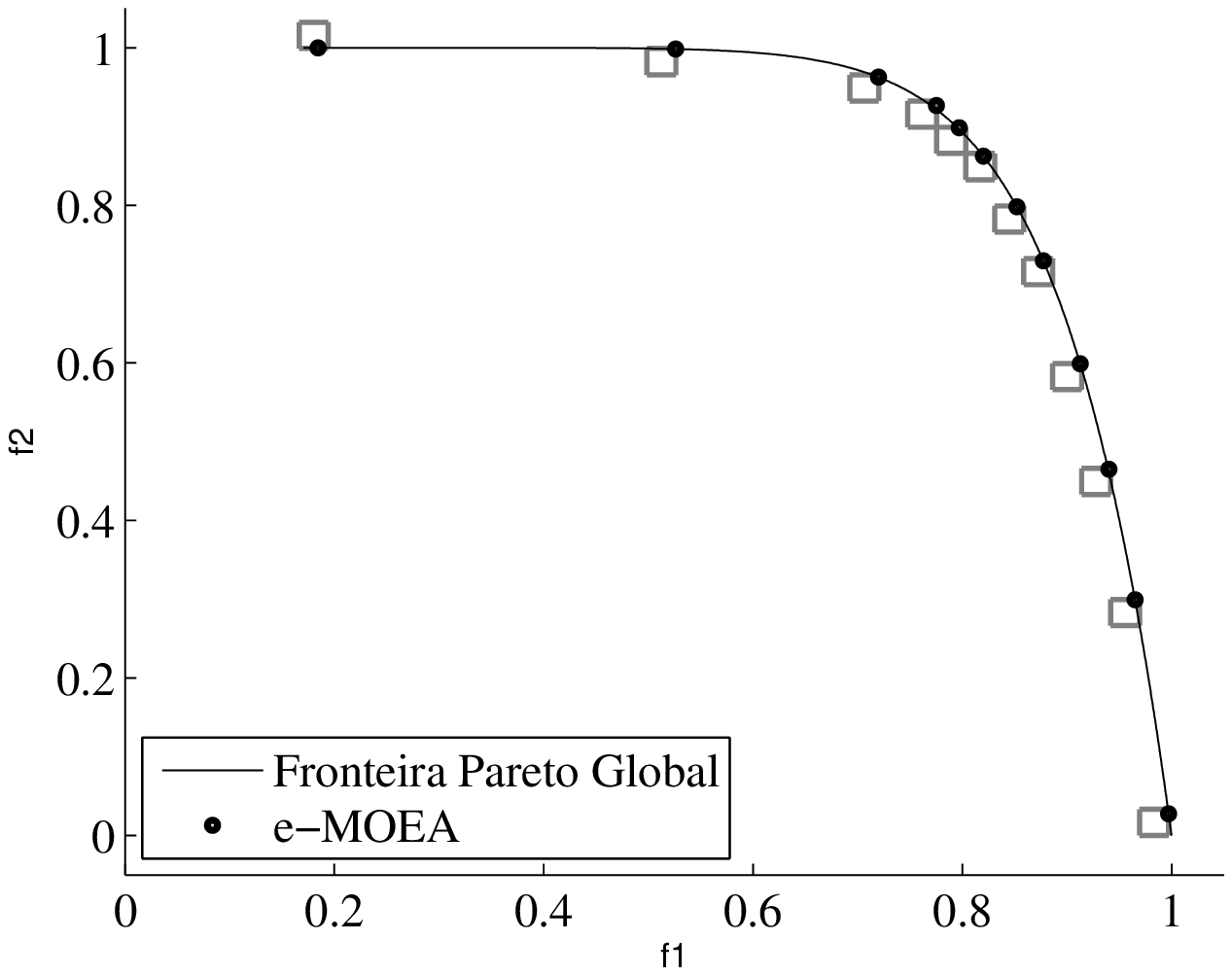}}
\subfigure[\small cone$\eps$-MOEA distribution on test Pol: \newline $\pmb{\eps}=(0.6180,0.9804)$, $\left|\mathcal{H}\right|=38$, $T=50$.]{\includegraphics[width=6.6cm]{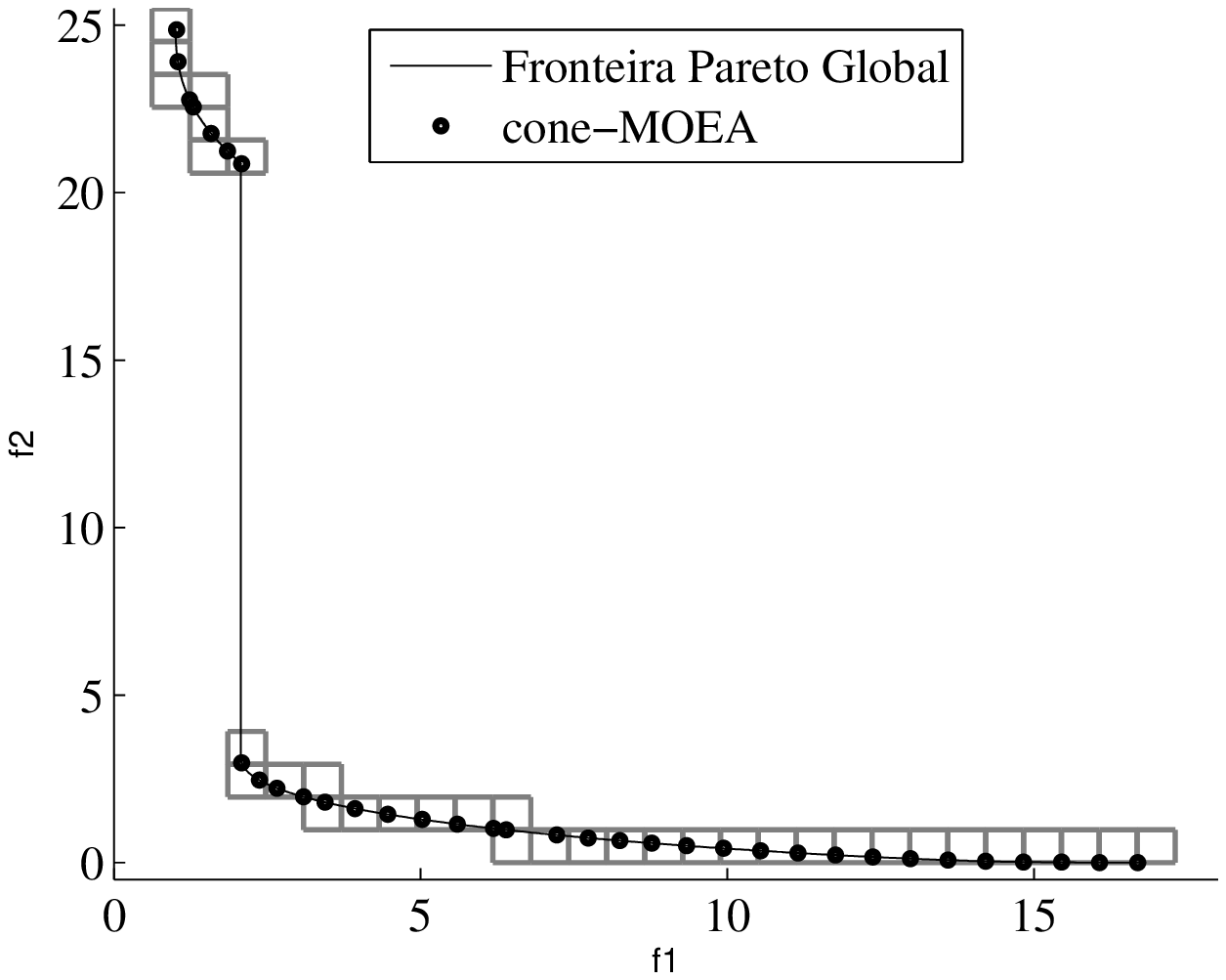}}
\subfigure[\small $\eps$-MOEA distribution on test Pol: \newline $\pmb{\eps}=(0.3151,0.5000)$, $\left|\mathcal{H}\right|=09$, $T=50$.]{\includegraphics[width=6.6cm]{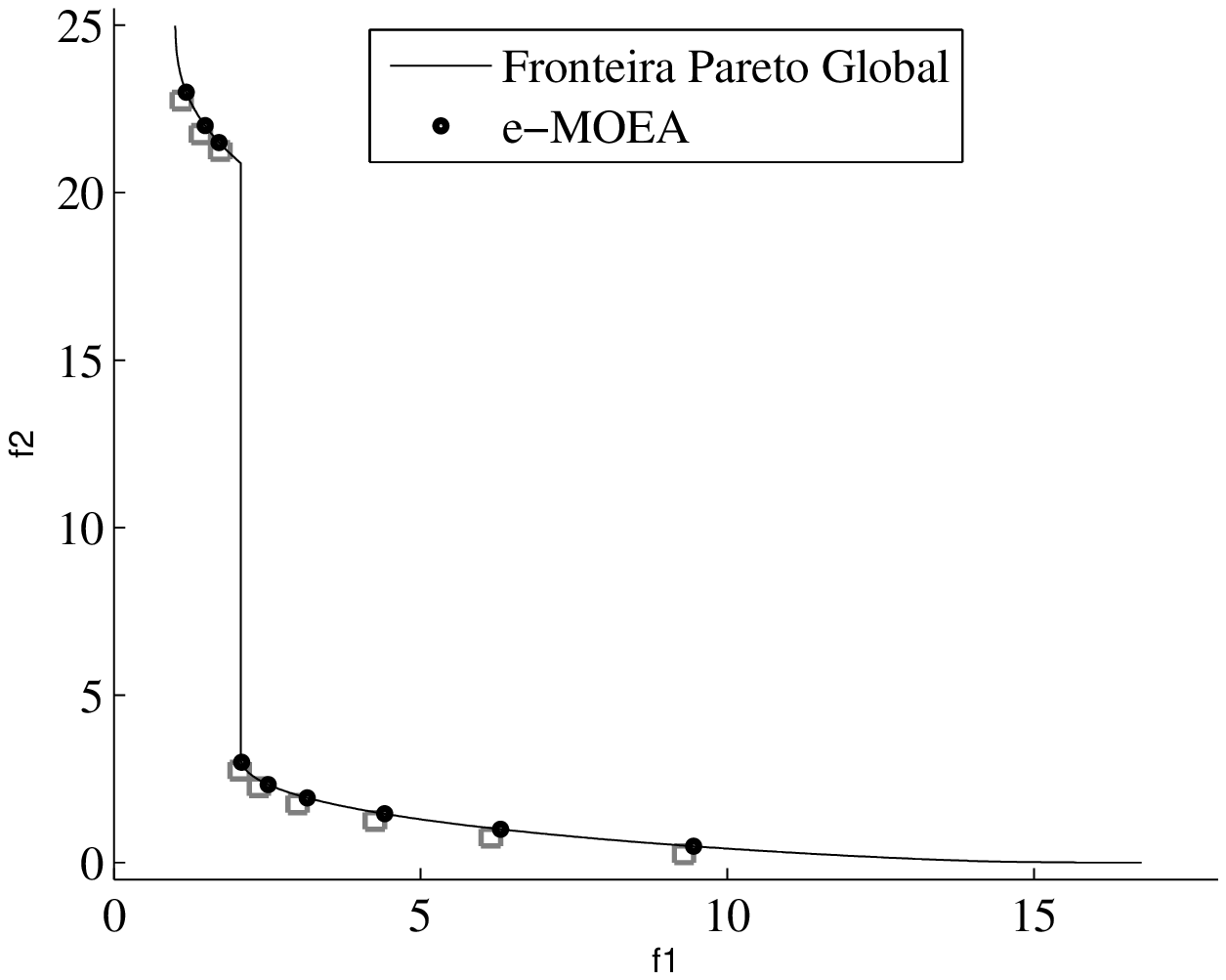}}
\caption{Efficient solutions generated by cone$\eps$-MOEA and $\eps$-MOEA considering calculated $\pmb{\eps}$ values to provide a maximum of $T$ solutions in the final archive. The number of solutions approximated by each approach is indicated by the cardinality of the final Pareto front $\left|\mathcal{H}\right|$. The presented fronts are the outcome of a typical run.}
\label{fig:viewbox}
\end{figure}

Note that, for the cone$\eps$-MOEA, the estimated $\pmb{\eps}$ values on Table \ref{tab:eps_values} differ from the calculated ones only in eight of the sixteen problems, among which three have a disconnected Pareto-optimal front, and a non adequate calculated $\pmb{\eps}$ value was expected. On the other hand, in the $\eps$-MOEA all $\pmb{\eps}$ values have changed. This phenomenon suggests that the cone$\eps$-dominance approach may be less susceptible to the loss of nondominated solutions than the $\eps$-dominance. This situation is clear from Fig. \ref{fig:viewbox}, which shows the distribution of solutions obtained on problems Deb52 and Pol in order to get $T=30$ and $T=50$ solutions, respectively. The number of solutions approximated by both approaches is indicated in the figure by considering the cardinality of the final Pareto front $\left|\mathcal{H}\right|$. In this simulation, the $\pmb{\eps}$ values were calculated using \eqref{eq:eps_value} for the $\eps$-MOEA and \eqref{eq:coneps} for the cone$\eps$-MOEA. The boxes within which a solution lies are also shown in the figure. It is interesting to notice that all solutions are nondominated according to the $\eps$-dominance or the cone$\eps$-dominance depending on the strategy adopted, and each box is occupied by only one solution. Furthermore, the cone$\eps$-approach has achieved a better spread of solutions in both tests, even in the Pol problem, which has a disconnected Pareto-optimal front. Because of the loss of efficient solutions by both strategies, only the ``estimated'' $\pmb{\eps}$ values are considered in the remainder of the experimental study.

As the cone$\eps$-dominance criterion is influenced by the $\kappa$ parameter, we have performed some preliminary testing to observe the effect of different $\kappa$ values on the performance of the cone$\eps$-MOEA. Table \ref{tab:hbar} shows the effect of this parameter on the values of the unary metrics $\gamma$, $\Delta$ and $HV$ for the benchmark problems Deb52, ZDT1, and DTLZ2 with three and four objective functions. In this simple test, intermediate values for $\kappa$ seem to yield reasonably good performance values for all metrics, from which a value of $\kappa = 0.5$ was chosen for all experiments conducted in this work. However, as the number of objectives increases, smaller values of $\kappa$ (i.e., $\kappa\rightarrow 0$) seem to be more adequate, at the cost of the loss of some nondominated solutions since the effect of the cone of dominance approximates that one of the $\eps$-approach. 

\begin{table*}
\renewcommand{\arraystretch}{1.02}
\centering
\caption{Influence of different $\kappa$ values on the performance of the cone$\eps$-MOEA on test problems Deb52, ZDT1, and DTLZ2 with three and four objective functions. Median (M) and standard deviation (SD) over 30 independent runs are shown. Intermediate values for $\kappa$ seem to yield reasonably good performance values for all metrics. A more appropriate study is required in order to formally characterize the effect of this parameter.}
\label{tab:hbar}
		\resizebox{\textwidth}{!}{ %
    \begin{math}    
    \begin{array}{clccccccccccc}
    \\\hline
    \multicolumn{2}{l}{\mbox{Metric}} & \multicolumn{11}{c}{\mbox{$\kappa$ ; Deb52}}  \\
   									\cline{3-13}
    			&&        \mbox{0.0} & \mbox{0.1} & \mbox{0.2} & \mbox{0.3} & \mbox{0.4} & \mbox{0.5} & \mbox{0.6} & \mbox{0.7} & \mbox{0.8} & \mbox{0.9} & \mbox{0.99} \\ 
    \hline    
    \mbox{$\gamma$} & \mbox{M}  & 0.0006   & 0.0006   & 0.0005   & 0.0006   & 0.0005   & 0.0006   & 0.0006    & 0.0006   & 0.0006   & 0.0006   & 0.0006   \\
                    & \mbox{SD} & <10^{-4} & <10^{-4} & <10^{-4} & <10^{-4} & <10^{-4} & <10^{-4} & <10^{-4}  & <10^{-4} & <10^{-4} & 0.0001   & 0.0001   \\
    \mbox{$\Delta$} & \mbox{M}  & 0.6766   & 0.6813   & 0.5244   & 0.2991   & 0.2552   & 0.2432   & 0.2648    & 0.2892   & 0.3147   & 0.3194   & 0.3199   \\
                    & \mbox{SD} & 0.0004   & 0.0021   & 0.0025   & 0.0027   & 0.0034   & 0.0039   & 0.0017    & 0.0019   & 0.0016   & 0.0042   & 0.0066   \\
    \mbox{HV}       & \mbox{M}  & 0.2735   & 0.2779   & 0.2794   & 0.2802   & 0.2806   & 0.2806   & 0.2806    & 0.2806   & 0.2806   & 0.2806   & 0.2806   \\
                    & \mbox{SD} & <10^{-4} & <10^{-4} & <10^{-4} & <10^{-4} & <10^{-4} & <10^{-4} & <10^{-4}  & <10^{-4} & <10^{-4} & <10^{-4} & <10^{-4} \\
    \mbox{$\left|\mathcal{H}\right|$} & \mbox{M}  &  19.00 &  51.00 &  74.00 &  93.00  & 101.00   & 101.00    & 101.00   & 101.00   & 101.00   & 101.00 & 101.00 \\
                                      & \mbox{SD} & <10^{-4} & <10^{-4} & 0.2537 & 0.3457 & 0.4842 & <10^{-4} & <10^{-4} & <10^{-4} & 0.1826   & 0.1826 & 0.1826 \\
    \hline    
    \\\hline
    \multicolumn{2}{l}{\mbox{Metric}} & \multicolumn{11}{c}{\mbox{$\kappa$ ; ZDT1}}  \\
   									\cline{3-13}
    			&&        \mbox{0.0} & \mbox{0.1} & \mbox{0.2} & \mbox{0.3} & \mbox{0.4} & \mbox{0.5} & \mbox{0.6} & \mbox{0.7} & \mbox{0.8} & \mbox{0.9} & \mbox{0.99} \\ 
    \hline    
    \mbox{$\gamma$} & \mbox{M}  & 0.0103 & 0.0069 & 0.0055 & 0.0059 & 0.0074 & 0.0040 & 0.0042 & 0.0051 & 0.0053 & 0.0050 & 0.0038 \\
                    & \mbox{SD} & 0.0072 & 0.0038 & 0.0057 & 0.0047 & 0.0049 & 0.0042 & 0.0060 & 0.0058 & 0.0040 & 0.0050 & 0.0034 \\
    \mbox{$\Delta$} & \mbox{M}  & 0.3046 & 0.5543 & 0.3678 & 0.2084 & 0.1818 & 0.1812 & 0.1898 & 0.1937 & 0.1934 & 0.1956 & 0.1891 \\
                    & \mbox{SD} & 0.0122 & 0.0607 & 0.0480 & 0.0408 & 0.0235 & 0.0220 & 0.0234 & 0.0251 & 0.0240 & 0.0232 & 0.0155 \\
    \mbox{HV}       & \mbox{M}  & 0.8435 & 0.8561 & 0.8602 & 0.8607 & 0.8598 & 0.8652 & 0.8650 & 0.8636 & 0.8633 & 0.8638 & 0.8657 \\
                    & \mbox{SD} & 0.0115 & 0.0066 & 0.0094 & 0.0079 & 0.0082 & 0.0069 & 0.0099 & 0.0096 & 0.0066 & 0.0083 & 0.0057 \\
    \mbox{$\left|\mathcal{H}\right|$} & \mbox{M}  &  37.00 &  63.00 &  84.50 &  98.00 & 100.00 & 101.00 & 101.00 & 101.00 & 101.00 & 101.00 & 101.00 \\
                                      & \mbox{SD} & 0.6397 & 5.7211 & 2.8730 & 5.0901 & 3.8201 & 0.5467 & 0.8584 & 0.9371 & 0.9377 & 1.3515 & 0.7112 \\      
    \hline
    \\\hline
    \multicolumn{2}{l}{\mbox{Metric}} & \multicolumn{11}{c}{\mbox{$\kappa$ ; DTLZ2 (m = 3)}}  \\
   									\cline{3-13}
    			&&        \mbox{0.0} & \mbox{0.1} & \mbox{0.2} & \mbox{0.3} & \mbox{0.4} & \mbox{0.5} & \mbox{0.6} & \mbox{0.7} & \mbox{0.8} & \mbox{0.9} & \mbox{0.99} \\ 
    \hline    
    \mbox{$\gamma$} & \mbox{M}  & 0.0062 & 0.0069 & 0.0072 & 0.0070 & 0.0074 & 0.0079 & 0.0074 & 0.0076 & 0.0074 & 0.0078 & 0.0072 \\
                    & \mbox{SD} & 0.0002 & 0.0013 & 0.0015 & 0.0013 & 0.0012 & 0.0014 & 0.0010 & 0.0019 & 0.0007 & 0.0014 & 0.0009 \\
    \mbox{$\Delta$} & \mbox{M}  & 0.0503 & 0.6066 & 0.3029 & 0.2411 & 0.2386 & 0.2308 & 0.2274 & 0.2175 & 0.2079 & 0.2173 & 0.1982 \\
                    & \mbox{SD} & 0.0041 & 0.0422 & 0.0357 & 0.0302 & 0.0264 & 0.0219 & 0.0316 & 0.0275 & 0.0306 & 0.0295 & 0.0239 \\
    \mbox{HV}       & \mbox{M}  & 0.6731 & 0.7149 & 0.7383 & 0.7435 & 0.7458 & 0.7469 & 0.7467 & 0.7469 & 0.7470 & 0.7470 & 0.7471 \\
                    & \mbox{SD} & 0.0066 & 0.0042 & 0.0023 & 0.0012 & 0.0007 & 0.0006 & 0.0005 & 0.0005 & 0.0005 & 0.0005 & 0.0003 \\
    \mbox{$\left|\mathcal{H}\right|$} & \mbox{M}  &  21.00 &  69.00 &  88.00 &  93.00 &  94.50 &  95.00 &  95.00 &  95.00 &  95.00 &  95.00 &  94.00 \\
                                      & \mbox{SD} & 1.3047 & 3.1639 & 2.8367 & 2.0424 & 1.7750 & 1.9464 & 2.2894 & 2.0197 & 1.5643 & 2.2614 & 1.7100 \\    
    \hline
    \\\hline
    \multicolumn{2}{l}{\mbox{Metric}} & \multicolumn{11}{c}{\mbox{$\kappa$ ; DTLZ2 (m = 4)}}  \\
   									\cline{3-13}
    			&&        \mbox{0.0} & \mbox{0.1} & \mbox{0.2} & \mbox{0.3} & \mbox{0.4} & \mbox{0.5} & \mbox{0.6} & \mbox{0.7} & \mbox{0.8} & \mbox{0.9} & \mbox{0.99} \\ 
    \hline    
    \mbox{$\gamma$} & \mbox{M}  & 0.0001 & 0.0311 & 0.0385 & 0.0312 & 0.0449 & 0.0404 & 0.0445 & 0.0488 & 0.0590 & 0.0489 & 0.0534 \\
                    & \mbox{SD} & 0.0001 & 0.0198 & 0.0218 & 0.0239 & 0.0283 & 0.0240 & 0.0369 & 0.0281 & 0.0284 & 0.0241 & 0.0304 \\
    \mbox{$\Delta$} & \mbox{M}  & 0.1390 & 0.4700 & 0.3602 & 0.3296 & 0.3299 & 0.3429 & 0.3377 & 0.3258 & 0.3253 & 0.3304 & 0.3319 \\
                    & \mbox{SD} & 0.1173 & 0.0304 & 0.0307 & 0.0226 & 0.0255 & 0.0263 & 0.0187 & 0.0262 & 0.0210 & 0.0254 & 0.0259 \\
    \mbox{$\left|\mathcal{H}\right|$} & \mbox{M}  &  14.00 &  79.50 &  90.00 &  92.00 &  95.00 &  96.00 &  95.50 &  97.00 &  98.00 &  95.50 &  97.00 \\
                                      & \mbox{SD} & 1.9815 & 4.9642 & 4.8476 & 4.2372 & 4.6307 & 4.2129 & 5.8530 & 5.0496 & 4.5945 & 4.6233 & 4.3423 \\    
    \hline
    \end{array}           
    \end{math}}
\end{table*}

\subsection{Statistical Design}

To evaluate the possible differences in the performance of the methods tested, we have employed tests designed to detect statistically significant differences and to estimate their sizes. The tests were performed independently for each of the quality metrics described previously, and are outlined in the following paragraphs. 

The raw dataset used for this analysis was composed of the final Pareto-front obtained on $50$ independent runs of each algorithm on each problem, from which the quality criteria described earlier were calculated. The statistical tests were then performed using these calculated values.

In order to detect whether any of the six algorithms presented a significant overall difference (i.e., when integrated over the whole test set), we employed a randomized complete block design \citep{book.Montgomery2008}, with the algorithms as levels of the experimental factor \textit{Algorithm}, and the problems as blocking levels. The statistical model considered by this experimental design can be defined, e.g., for the $\Delta$ metric, as:
\begin{equation}
y^{\Delta}_{ijk} = \mu^{\Delta} + \tau_i^{\Delta} + \beta_j^{\Delta} + \epsilon_{ijk}^{\Delta}
\end{equation}

\noindent in which $y^{\Delta}_{ijk}$ represents the value of this metric on the $k^{th}$ run  of the $i^{th}$ algorithm on the $j^{th}$ problem; $\mu^{\Delta}$ is the mean of $\Delta$ over all observations; $\tau_i^{\Delta}$ is the contribution due to the $i^{th}$ algorithm; $\beta_j^{\Delta}$ is the component due to the $j^{th}$ problem (block); and $\epsilon_{ijk}^{\Delta}$ is the residual.

For each metric considered, the ANOVA was applied\footnote{In all tests reported, the assumptions of the parametric ANOVA were verified by means of residual analysis. While small deviations from the normality assumption were detected, these were deemed unimportant, since the test is robust to moderate departures from normality in balanced designs with large sample sizes \citep{book.Montgomery2008}.} to test the null hypothesis of absence of difference among the six algorithms evaluated over all problems against it two-sided alternative, i.e. (again, using the $\Delta$ metric as an example):
\begin{equation}
\begin{split}
\label{eq:h0h1c}
\begin{cases} 
H_0^{\Delta}: \tau_i^{\Delta} = 0,\ \forall i
\medskip \\
H_1^{\Delta}: \tau_i^{\Delta}\neq 0\ \mbox{for any } i
\end{cases}
\end{split}
\end{equation}

After the significance tests, estimators of the block effects were obtained \citep{book.Montgomery2008} and subtracted from the samples, thus allowing for problem-independent pairwise comparisons of the algorithms. Bootstrap estimations of the distribution of the means \citep{book.Davison1997} for each algorithm on each metric were then obtained, which enabled the use of simple two-sample t-tests and parametric estimators of the differences of means \citep{book.Montgomery2006} for the post-hoc pairwise comparisons. These comparisons allow not only the determination of statistical significance, but also the estimation of the magnitude of the differences, which is important in assessing the \textit{practical significance} of results \citep{Ellis2010}.

Besides the overall analysis performed using the design above, statistical tests were also performed for each problem independently, in order to assess the relative performance of the algorithms on each of the test problems. For each problem and performance metric, the null hypotheses used were those of equality of mean values against two-sided alternative hypotheses. An one-way ANOVA design was used in these tests, and effect size estimators were also calculated using bootstrap, allowing post-hoc pairwise comparisons between each pair of algorithms. In all cases, the significance level used was $99\%$.

\section{Results and Discussion}
\label{sec:6}

The results obtained for the four performance metrics considered, using the estimated values of $\pmb{\eps}$, are summarized in this section. The analysis of the results is divided in four parts. In the first we present an overall analysis of the results, based on the observed general performance of the algorithms tested on the four performance criteria defined in the previous section. The second part consists of a detailed analysis of the behavior of each algorithm for the individual problems, in terms of their known properties and geometrical characteristics of the Pareto fronts. The third part consists of a cardinality analysis regarding the number of solutions estimated by both $\eps$ and cone$\eps$-approaches, and their effects on the performance of the methods. Finally, possible extensions are dicussed in the fourth part.

\subsection{Overall Analysis}

\begin{figure*}[!ht]
\centering
\psfrag{spea2xxxxxxxxx}[][]{\tiny SPEA2}
\psfrag{cnsgaiixxxxxxx}[][]{\tiny C-NSGA-II\ \ \ \ }
\psfrag{coneepsmoeaxxx}[][]{\tiny cone$\eps$-MOEA\ \ \ \ \ }
\psfrag{epsmoeaxxxxxxx}[][]{\tiny $\eps$-MOEA\ \ }
\psfrag{nsgaiixxxxxxxx}[][]{\tiny NSGA-II\ \ \ \ }
\psfrag{nsgaii*xxxxxxx}[][]{\tiny NSGA-II*\ \ \ \ }
\psfrag{Gamma}[][]{\footnotesize $\gamma$}
\psfrag{Delta}[][]{\footnotesize $\Delta$}
\psfrag{CS}[][]{\footnotesize CS}
\psfrag{HV}[][]{\footnotesize HV}
\psfrag{0.3}[][]{\footnotesize $0.3$}
\psfrag{0.4}[][]{\footnotesize $0.4$}
\psfrag{0.5}[][]{\footnotesize $0.5$}
\psfrag{0.6}[][]{\footnotesize $0.6$}
\psfrag{0.7}[][]{\footnotesize $0.7$}
\psfrag{0.30}[][]{\footnotesize $0.30$}
\psfrag{0.35}[][]{\footnotesize $0.35$}
\psfrag{0.40}[][]{\footnotesize $0.40$}
\psfrag{0.45}[][]{\footnotesize $0.45$}
\psfrag{0.50}[][]{\footnotesize $0.50$}
\psfrag{0.92}[][]{\footnotesize $0.92$}
\psfrag{0.93}[][]{\footnotesize $0.93$}
\psfrag{0.94}[][]{\footnotesize $0.94$}
\psfrag{0.95}[][]{\footnotesize $0.95$}
\psfrag{0.96}[][]{\footnotesize $0.96$}
\psfrag{0.06}[][]{\footnotesize $0.06$}
\psfrag{0.08}[][]{\footnotesize $0.08$}
\psfrag{0.10}[][]{\footnotesize $0.10$}
\psfrag{0.12}[][]{\footnotesize $0.12$}
\psfrag{0.14}[][]{\footnotesize $0.14$}
\psfrag{0.16}[][]{\footnotesize $0.16$}
\subfigure[\small $\Delta$ metric.]{\includegraphics[width=6.5cm]{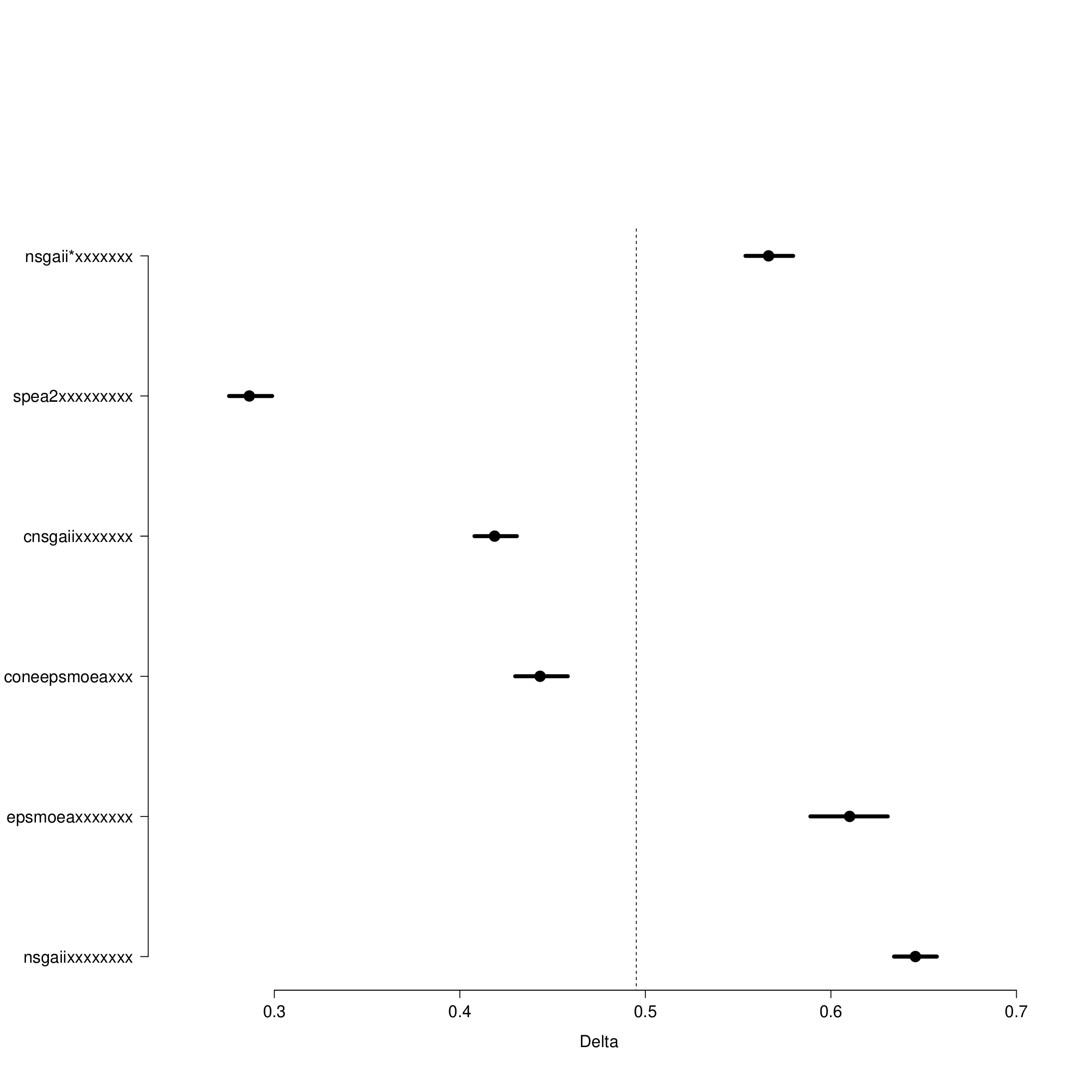}}
\subfigure[\small $\gamma$ metric.]{\includegraphics[width=6.5cm]{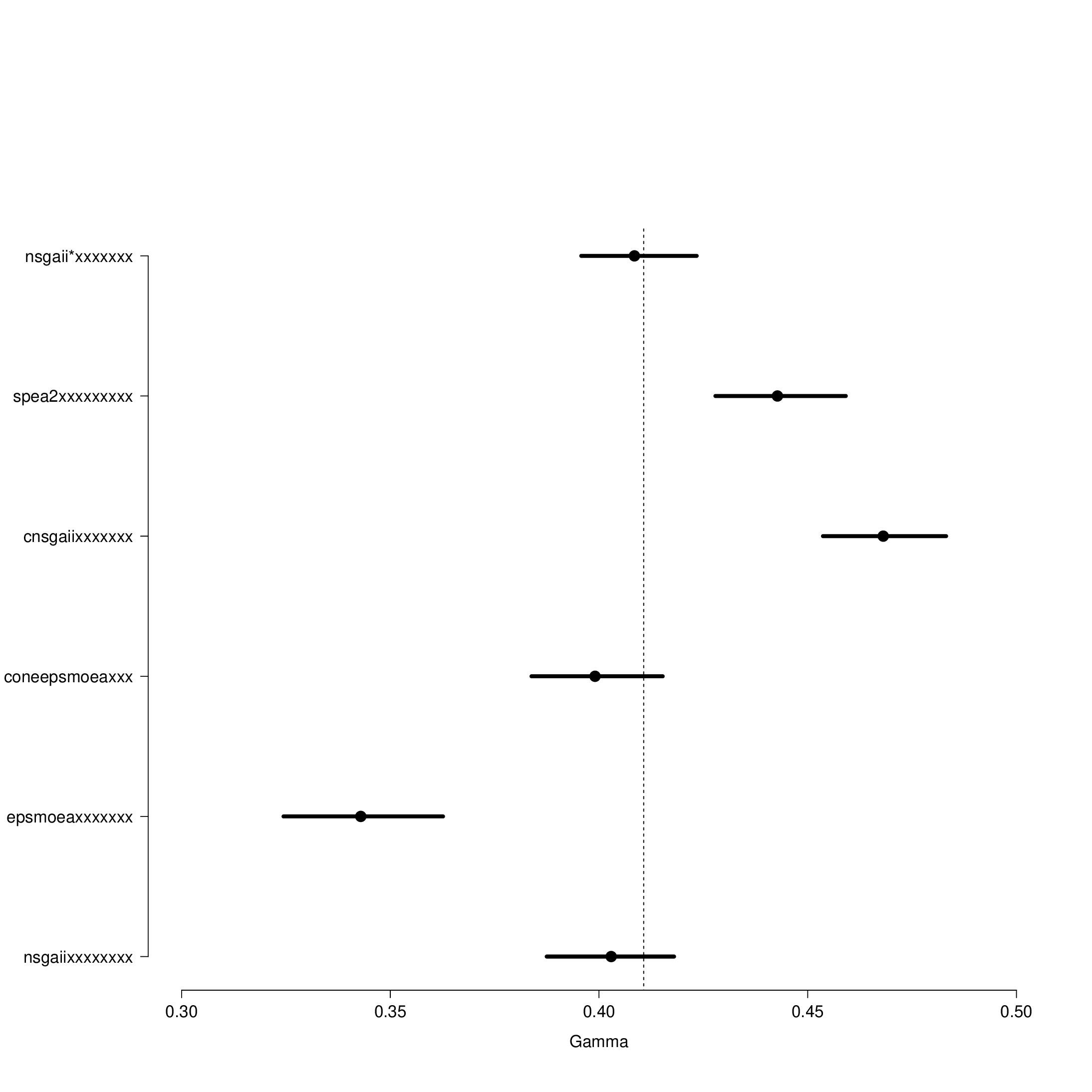}}
\subfigure[\small $HV$ metric.]{\includegraphics[width=6.5cm]{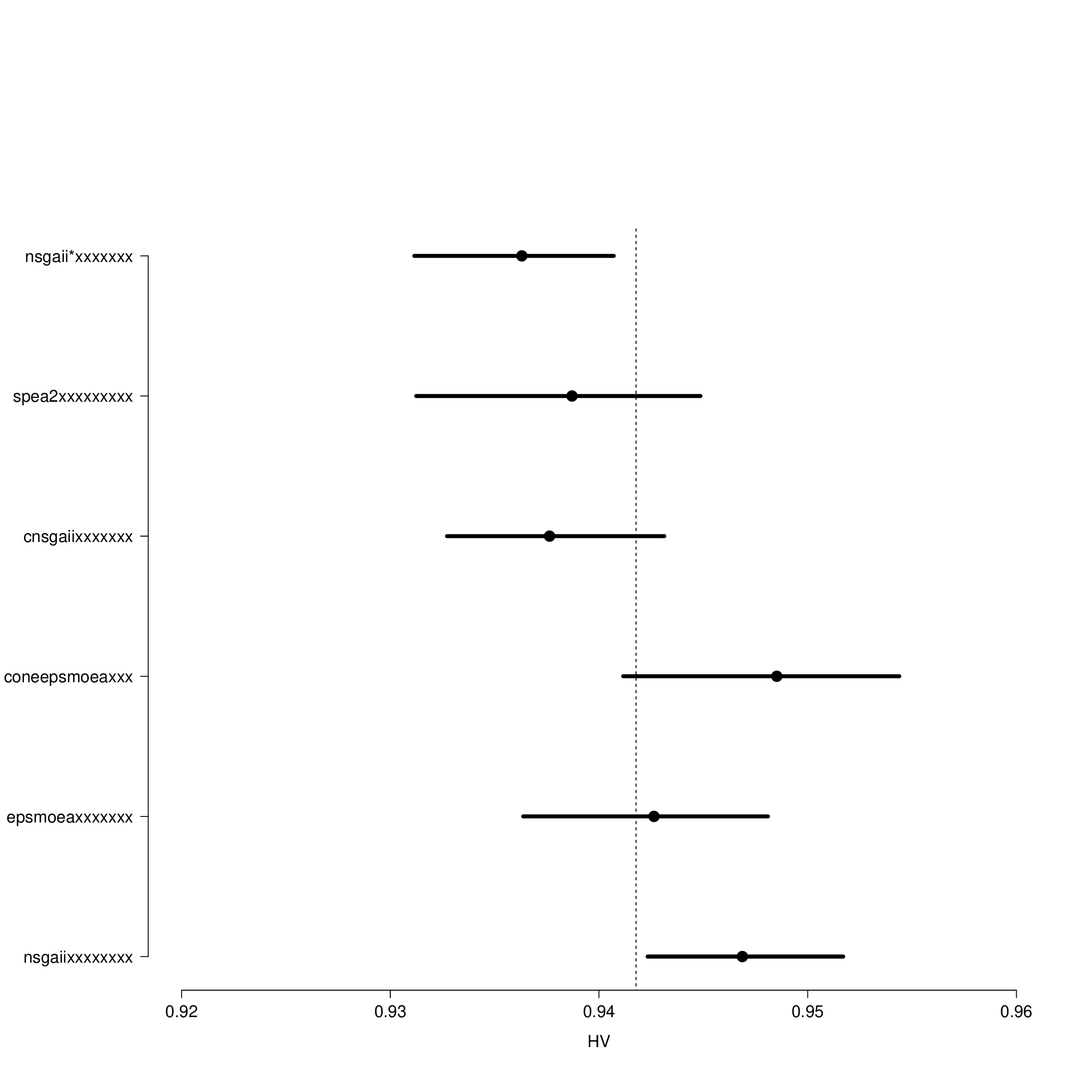}}
\subfigure[\small $CS$ metric.]{\includegraphics[width=6.5cm]{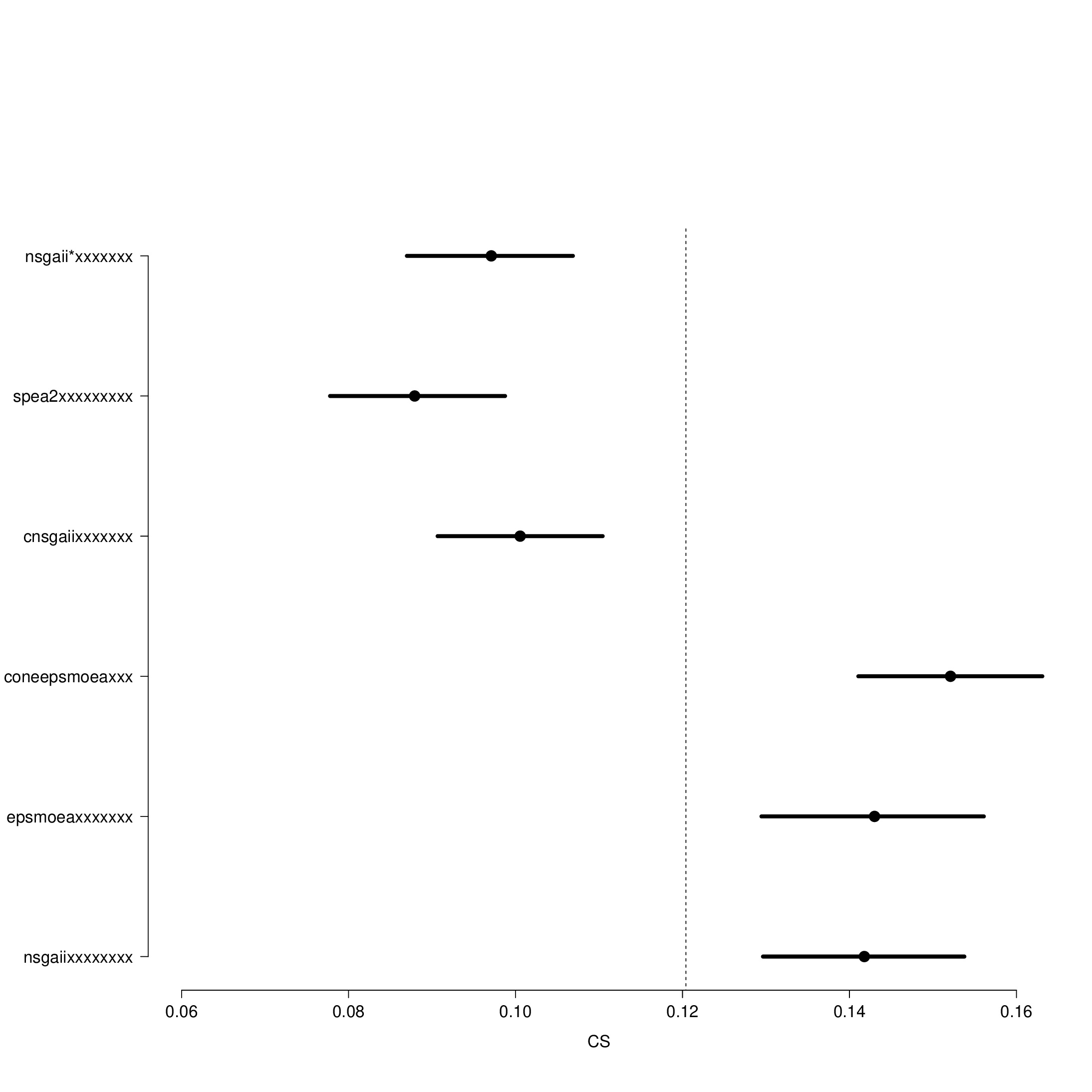}}
\caption{Individual $99\%$ confidence intervals for the estimated average performance of the six algorithms on each metric. The dotted line marks the overall mean result for the metric.}
\label{fig:bootstrp1}
\end{figure*}

The results obtained in the overall analysis are summarized in Tables \ref{tab:overall_results} and \ref{tab:rank_results}, and on Fig. \ref{fig:bootstrp1}. From these results, a few considerations can be made over the relative performance of the six algorithms on the test set used.

\begin{table}
\centering
\caption{Overall pairwise differences. The numbers indicate the point estimator for the difference between the \textit{row} algorithm and the \textit{column} algorithm (after removing problem effects). Differences marked with an asterisk indicate significance at the $99\%$ familywise confidence level, adjusted using Tukey' HSD procedure \citep{book.Montgomery2006}.}
\label{tab:overall_results}
\vspace{2pt}
\begin{tabular*}{13cm}{l|ccccc} \hline
\multicolumn{6}{c}{$\mathbf{\Delta}$ \textbf{metric}}\\
\hline
\textbf{Algorithm}			&NSGA-II	&$\eps$-MOEA	&cone$\eps$-MOEA	&C-NSGA-II	&SPEA2\\
\hline
$\eps$-MOEA					&-0.03*			&						&									&					&\\
cone$\eps$-MOEA			&-0.20*			&-0.17*					&								&					&\\
C-NSGA-II						&-0.23*			&-0.19*				&-0.02*						&					&\\
SPEA2					 		&-0.36*			 &-0.32*				&-0.15*							&-0.13*					&\\
NSGA-II*						&-0.08*			&-0.04*					&0.12*							&0.15*					&0.28*\\
\hline
\end{tabular*}\\
\vspace{2pt}
\begin{tabular*}{13cm}{lc} 
\textbf{Ordering$^{**}$:}	&\footnotesize{ SPEA2\ \ \ \ C-NSGA-II\ \ \ \ cone$\eps$-MOEA\ \ \ \ NSGA-II*\ \ \ \ $\eps$-MOEA\ \ \ \ NSGA-II}\\
\hline\hline\end{tabular*}\\
\vspace{10pt}

\begin{tabular*}{13cm}{l|ccccc} \hline
\multicolumn{6}{c}{$\mathbf{\gamma}$ \textbf{metric}}\\
\hline
\textbf{Algorithm}			&NSGA-II	&$\eps$-MOEA	&cone$\eps$-MOEA	&C-NSGA-II	&SPEA2\\
\hline
$\eps$-MOEA					&-0.06*		&							&									&					&\\
cone$\eps$-MOEA			&-4e-3		 &0.06*					&									&					&\\
C-NSGA-II					   &0.06*		&0.12*					&0.07*							&					&\\
SPEA2					 		  &0.04*		&0.10*					&0.04*							&-0.02			&\\
NSGA-II*						&6e-3		  &0.07*				 &0.01								&-0.06*			&-0.03*\\
\hline
\end{tabular*}\\
\vspace{2pt}
\begin{tabular*}{13cm}{lc} 
\textbf{Ordering$^{**}$:}			&\footnotesize{$\eps$-MOEA\ \ \ \ $\overline{\mbox{cone$\eps$-MOEA\ \ NSGA-II\ \ NSGA-II*}}$\ \ \ \ $\overline{\mbox{SPEA2\ \ C-NSGA-II}}$}\\
\hline\hline\end{tabular*}\\
\vspace{10pt}

\begin{tabular*}{13cm}{l|ccccc} \hline
\multicolumn{6}{c}{\textbf{CS metric}}\\
\hline
\textbf{Algorithm}			&NSGA-II	&$\eps$-MOEA	&cone$\eps$-MOEA	&C-NSGA-II	&SPEA2\\
\hline
$\eps$-MOEA					&1e-3		&							&									&					&\\
cone$\eps$-MOEA			&0.01			&9e-3					&									&					&\\
C-NSGA-II						&-0.04*		&-0.04*					&-0.05*							&					&\\
SPEA2					 			&-0.05*		&-0.06*					&-0.06*							&-0.01			&\\
NSGA-II*							&-0.04*		&-0.04*					&-0.05*							&-3e-3		&9e-3\\
\hline
\end{tabular*}\\
\vspace{2pt}
\begin{tabular*}{13cm}{lc} 
\textbf{Ordering$^{**}$:}			&\footnotesize{$\overline{\mbox{cone$\eps$-MOEA\ \ $\eps$-MOEA\ \ NSGA-II}}$\ \ \ \ $\overline{\mbox{C-NSGA-II\ \ NSGA-II*\ \ SPEA2}}$}\\
\hline\hline\end{tabular*}\\
\vspace{10pt}

\begin{tabular*}{13cm}{l|ccccc} \hline
\multicolumn{6}{c}{\textbf{HV metric}}\\
\hline
\textbf{Algorithm}			&NSGA-II	&$\eps$-MOEA	&cone$\eps$-MOEA	&C-NSGA-II	&SPEA2\\
\hline
$\eps$-MOEA					&-4e-3		&							&									&					&\\
cone$\eps$-MOEA			&2e-3		&6e-3					&									&					&\\
C-NSGA-II						&-9e-3		&-5e-3					&-0.01*							&					&\\
SPEA2					 			&-8e-3		&-4e-3					&-0.01*							&9e-4			&\\
NSGA-II*							&-0.01*		&-6e-3					&-0.01*							&-1e-3			&-2e-3\\
\hline
\end{tabular*}\\
\vspace{2pt}
\begin{tabular*}{13cm}{lc} 
\textbf{Ordering$^{**}$:}			&\footnotesize{$\overline{\mbox{cone$\eps$-MOEA\ }\underline{\mbox{\ NSGA-II\ }\overline{\mbox{\ $\eps$-MOEA\ }}}}\overline{\underline{\mbox{\ SPEA2\ \ C-NSGA-II\ }}\mbox{\ NSGA-II*}}$}\\
\end{tabular*}\\
\begin{tabular*}{13cm}{l|cccc} \hline\hline\end{tabular*}\\
\vspace{1pt}
\footnotesize{** Ordering from best to worst. Horizontal bars indicate the absence of statistically significant differences.}
\end{table}

First of all, the diversity metric $\Delta$ presented an interesting ordering of performance, with the SPEA2 presenting the best performance, the C-NSGA-II in second ($+0.13$)\footnote{The numbers in parentheses represent the difference in performance between a given algorithm and the best average result.}, the cone$\eps$-MOEA in third ($+0.15$), the NSGA-II* in fourth ($+0.28$), the $\eps$-MOEA in fifth ($+0.32$) and the NSGA-II ranking last ($+0.36$). Table \ref{tab:overall_results} shows that the magnitude of the differences for this metric are considerable from a practical point of view, with an absolute difference of $36\%$ between the best (SPEA2) and worst (NSGA-II) algorithms. The difference magnitudes reported in Table \ref{tab:overall_results} also suggest that even though the cone$\eps$-MOEA has occupied the third position in this rank, its difference from the C-NSGA-II is of little effect from a practical point of view. It is also relevant to have in mind that the computational complexity of the cone$\eps$ approach is significantly inferior to those of the two first ones (SPEA2 and C-NSGA-II). Figure \ref{fig:bootstrp1} and Table \ref{tab:rank_results} provide the estimated absolute values and confidence intervals for each algorithm, which provide the necessary information for the derivation of the significance-based ranks given in Table \ref{tab:rank_results}.

The performance ordering of the convergence metric $\gamma$ shows that the $\eps$-MOEA was able to significantly outperform all other approaches in this metric, with the cone$\eps$-MOEA ($+0.06$), the NSGA-II ($+0.06$) and the NSGA-II* ($+0.07$) tied in second, and the SPEA2 ($+0.10$) and C-NSGA-II ($+0.12$) tied in fifth. The effect sizes in this metric are also significant, with an absolute difference between the best ($\eps$-MOEA) and the worst (C-NSGA-II and SPEA2) of around $12\%$. Figure \ref{fig:bootstrp1} and Tables \ref{tab:overall_results} and \ref{tab:rank_results} again provide detailed information on the magnitude of the differences and their statistical significance.

The generalized coverage metric $CS$ presented an ordering with the cone$\eps$-MOEA, $\eps$-MOEA ($-9e-3$), and NSGA-II ($-0.01$) tied in first, and the C-NSGA-II ($-0.05$), NSGA-II* ($-0.05$), and SPEA2 ($-0.06$) tied in fourth. Here the magnitudes of the effects were relatively modest, with an absolute difference of about $6\%$ between the best and worst performing algorithms. Given the relatively low absolute average values for this metric (see Table \ref{tab:rank_results}), it is reasonable to suppose that in most cases all algorithms were finding at least some solutions very close to the true Pareto front, which would contribute to lowering the value of this metric for all others. However, the lack of other works to use as a comparison baseline for this metric makes it difficult to make any conclusive statement about the practical significance of these effect sizes.

Finally, the hypervolume metric $HV$ had a somewhat more complex ordering structure, with no statistically significant difference among the cone$\eps$-MOEA, NSGA-II ($-2e-3$), and $\eps$-MOEA ($-6e-3$); among the NSGA-II, $\eps$-MOEA, SPEA2 ($-0.01$) and C-NSGA-II ($-0.01$) algorithms; and also among the $\eps$-MOEA, SPEA2, C-NSGA-II, and NSGA-II* ($-0.01$) methods. The horizontal bars at the bottom of the HV section of Table \ref{tab:overall_results} illustrate this ordering of algorithms. In any case, the magnitudes of the differences were of very little practical significance, with an average absolute difference of only about $1\%$ between the best and worst-ranking algorithms. It is also important to notice here that, as explained in the next section, this quality metric is not an adequate indicator to be used with problems such as DTLZ8 and DTLZ9, and was not calculated for these two problems.

\begin{table}
\centering
\small
\caption{Aggregation of overall results. The values indicate the estimated global mean and standard error for each algorithm (after removing problem effects), and the value in parentheses is the algorithm rank within each metric. Ties were treated using the mean rank method, so that the sum of ranks is always the same for each metric. The final ranking is then calculated as an ordering of the average ranks.}
\label{tab:rank_results}
\vspace{2pt}
\begin{tabular}{l|cccccc} \hline
\textbf{Algorithm}								&$\Delta$					&$\gamma$					&HV										&CS								&Avg. Rank					&\textbf{Final Rank}\\
\hline
\multirow{2}{*}{NSGA-II}					&0.64$\pm$4e-3		&0.40$\pm$6e-3			&0.95$\pm$2e-3					&0.14$\pm$5e-3		&\multirow{2}{*}{3.5}	&\multirow{2}{*}{\textbf{3}} \\	
															& (6)						&(3)								&(3)										&(2)								& 										&\\
\hline
\multirow{2}{*}{$\eps$-MOEA}			&0.61$\pm$8e-3		&0.34$\pm$7e-3		&0.94$\pm$2e-3					&0.14$\pm$5e-3		&\multirow{2}{*}{2.88}	&\multirow{2}{*}{\textbf{2}} \\	
															& (5)						&(1)								&(3.5)									&(2)							& 										&\\
\hline
\multirow{2}{*}{cone$\eps$-MOEA}	&0.44$\pm$5e-3		&0.40$\pm$6e-3			&0.95$\pm$2e-3				&0.15$\pm$4e-3		&\multirow{2}{*}{2.5}	&\multirow{2}{*}{\textbf{1}} \\	
															& (3)						&(3)								&(2)									&(2)							& 										&\\
\hline	
\multirow{2}{*}{C-NSGA-II}					&0.42$\pm$5e-3		&0.47$\pm$6e-3			&0.94$\pm$2e-3				&0.10$\pm$4e-3		&\multirow{2}{*}{4.13}	&\multirow{2}{*}{\textbf{5.5}} \\	
															& (2)							&(5.5)							&(4)									&(5)								& 										&\\
\hline
\multirow{2}{*}{SPEA2}							&0.29$\pm$4e-3		&0.44$\pm$6e-3			&0.94$\pm$2e-3				&0.09$\pm$4e-3		&\multirow{2}{*}{3.88}	&\multirow{2}{*}{\textbf{4}} \\	
															&(1)								&(5.5)							&(4)									&(5)									& 										&\\
\hline
\multirow{2}{*}{NSGA-II*}					&0.57$\pm$5e-3		&0.41$\pm$5e-3			&0.94$\pm$2e-3					&0.10$\pm$4e-3		&\multirow{2}{*}{4.13}	&\multirow{2}{*}{\textbf{5.5}} \\	
															& (4)							&(3)							&(4.5)										&(5)								& 										&\\
\hline\end{tabular}\\
\end{table}
\begin{figure*}[!htb]
\centering
\psfrag{Delta}[][]{$\mathbf{\Delta}$}
\psfrag{Gamma}[][]{$\mathbf{\gamma}$}
\psfrag{NSGA-II}[][]{\footnotesize\textbf{NSGA-II}}
\psfrag{C-NSGA-II}[][]{\footnotesize\textbf{C-NSGA-II}}
\psfrag{SPEA2}[][]{\footnotesize\textbf{SPEA2}}
\psfrag{eps-MOEA}[][]{\footnotesize\textbf{$\mathbf{\eps}$-MOEA}}
\psfrag{coneeps-MOEA}[][]{\footnotesize\textbf{cone$\mathbf{\eps}$-MOEA}}
\psfrag{NSGA-II*}[][]{\footnotesize\textbf{NSGA-II*}}
\psfrag{0.3}[][]{$0.3$}
\psfrag{0.4}[][]{$0.4$}
\psfrag{0.5}[][]{$0.5$}
\psfrag{0.6}[][]{$0.6$}
\psfrag{0.7}[][]{$0.7$}
\psfrag{0.30}[][]{$0.30$}
\psfrag{0.35}[][]{$0.35$}
\psfrag{0.40}[][]{$0.40$}
\psfrag{0.45}[][]{$0.45$}
\psfrag{0.50}[][]{$0.50$}
\includegraphics[width=0.7\textwidth]{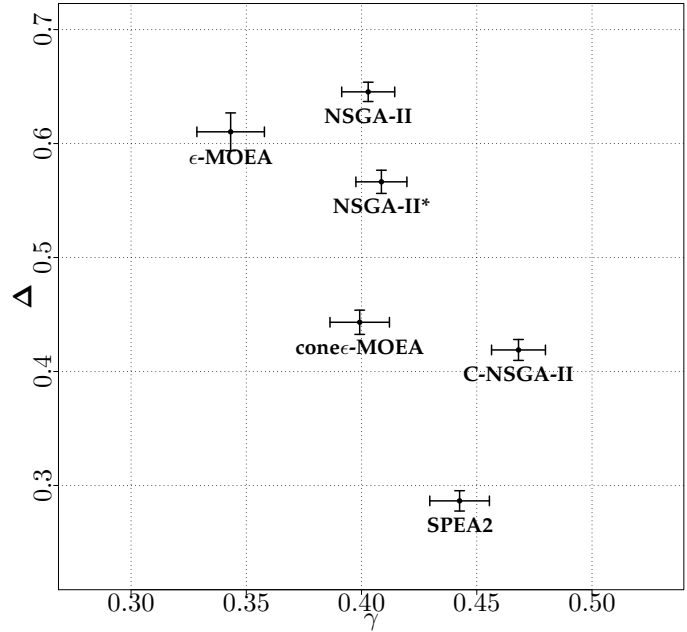}
\caption{Compromise between convergence ($\gamma$) and diversity ($\Delta$) for each algorithm. The bars represent $99\%$ confidence intervals.}
\label{fig:GammaXDelta}
\end{figure*}

Table \ref{tab:rank_results} shows a rank analysis of the algorithms, in which the results of the four metrics are ranked and aggregated using a simple averaging criterion, and an overall rank is calculated for the algorithms. No statistical analysis of significance was possible for the rank analysis, due to the non-independence of the four performance metrics, which could bias the results of most inferential procedures. However, the final ranks suggest that the cone$\eps$-MOEA was capable of presenting an efficient and balanced performance over all the performance metrics considered. The $\eps$-MOEA comes in a close second, mainly due to its poor performance on the $\Delta$ metric. The NSGA-II ranked third, also mainly as a consequence of its very poor diversity performance. The SPEA2 was ranked fourth, and the C-NSGA-II and NSGA-II* were tied in the last place. From the observation of their performance rankings on all metrics other than the $\Delta$, it is possible to observe that both the SPEA2 and the C-NSGA-II seem to have attained good diversity performance at the cost of their convergence properties. As for the NSGA-II*, the modification of the crowding distance procedure seems to have indeed improved its diversity when compared against the original NSGA-II, but not enough to be competitive with the cone$\eps$-MOEA, C-NSGA-II or SPEA2 algorithms.

To conclude this overall analysis, Fig. \ref{fig:GammaXDelta} illustrates the compromise between the convergence and diversity obtained for each algorithm. From this figure, it is possible to notice that the best tradeoffs with respect to these two performance indicators are obtained by the SPEA2, cone$\eps$-MOEA and $\eps$-MOEA. As expected from the data shown in Table \ref{tab:rank_results}, the cone$\eps$-MOEA is shown to provide an efficient balance between convergence and diversity, which makes it an interesting and competitive approach for the solutions of MOPs in which a good sampling of the Pareto optimal front is desired.

\subsection{Problemwise Analysis}

The discussion presented in this section consists of an analysis of the behavior of each algorithm for the individual problems. The results obtained for each problem are summarized in Tables \ref{tab:problemwise1} and \ref{tab:problemwise2}.

\begin{table}
\centering
\caption{Problemwise comparison of the algorithms on the four performance metrics used, for problems Deb52, Pol and the ZDT family. The values reported represent the mean and standard error obtained for each combination of algorithm, problem and performance metric.}
\label{tab:problemwise1}
\resizebox{\textwidth}{!}{%
\begin{tabular}{l|c|c|c|c|c|c|c}
\hline
Problem	 							&Metric 			&NSGA-II 	&$\eps$-MOEA	&cone$\eps$-MOEA 	&C-NSGA-II 		&SPEA2		&NSGA-II*\\
\hline
\multirow{4}{*}{Deb52}		&$\gamma$		&0.58$\pm$7e-3		&0.56$\pm$3e-3		&0.56$\pm$2e-3		&0.58$\pm$1e-2		&0.55$\pm$7e-3		&0.55$\pm$6e-3		\\
							&$\Delta$		&0.53$\pm$8e-3		&0.99$\pm$6e-4		&0.32$\pm$3e-4		&0.33$\pm$6e-3		&0.20$\pm$3e-3		&0.41$\pm$8e-3		\\
							&HV		&0.99$\pm$7e-5		&0.99$\pm$6e-5		&0.99$\pm$0		&0.99$\pm$1e-4		&0.99$\pm$3e-5		&0.99$\pm$7e-5		\\
							&CS		&0.02$\pm$7e-4		&0.03$\pm$10e-4		&0.03$\pm$8e-4		&0.02$\pm$8e-4		&0.03$\pm$9e-4		&0.02$\pm$8e-4		\\
\hline\hline
\multirow{4}{*}{Pol}		&$\gamma$		&0.20$\pm$2e-2		&0.13$\pm$6e-4		&0.19$\pm$2e-2		&0.19$\pm$9e-3		&0.15$\pm$2e-3		&0.16$\pm$2e-3		\\
							&$\Delta$		&0.58$\pm$1e-2		&0.98$\pm$9e-4		&0.29$\pm$6e-3		&0.38$\pm$7e-3		&0.24$\pm$3e-3		&0.36$\pm$8e-3		\\
							&HV		&1.00$\pm$1e-5		&1.00$\pm$5e-6		&1.00$\pm$4e-6		&1.00$\pm$3e-5		&1.00$\pm$4e-6		&1.00$\pm$8e-6		\\
							&CS		&0.04$\pm$1e-3		&0.04$\pm$1e-3		&0.04$\pm$2e-3		&0.04$\pm$10e-4		&0.06$\pm$1e-3		&0.05$\pm$1e-3		\\
\hline\hline
\multirow{4}{*}{Zdt1}		&$\gamma$		&0.16$\pm$2e-2		&0.30$\pm$3e-2		&0.23$\pm$3e-2		&0.18$\pm$2e-2		&0.30$\pm$2e-2		&0.19$\pm$2e-2		\\
							&$\Delta$		&0.79$\pm$1e-2		&0.70$\pm$4e-3		&0.37$\pm$6e-3		&0.50$\pm$8e-3		&0.29$\pm$7e-3		&0.56$\pm$1e-2		\\
							&HV		&0.99$\pm$10e-4		&0.98$\pm$2e-3		&0.98$\pm$2e-3		&0.98$\pm$9e-4		&0.98$\pm$1e-3		&0.98$\pm$1e-3		\\
							&CS		&0.33$\pm$2e-2		&0.20$\pm$3e-2		&0.27$\pm$3e-2		&0.30$\pm$2e-2		&0.15$\pm$2e-2		&0.27$\pm$2e-2		\\
\hline\hline
\multirow{4}{*}{Zdt2}		&$\gamma$		&0.43$\pm$9e-3		&0.65$\pm$8e-3		&0.30$\pm$4e-3		&0.80$\pm$1e-2		&0.41$\pm$8e-3		&0.41$\pm$8e-3		\\
							&$\Delta$		&0.76$\pm$1e-2		&0.56$\pm$3e-3		&0.38$\pm$4e-3		&0.50$\pm$7e-3		&0.28$\pm$4e-3		&0.58$\pm$1e-2		\\
							&HV		&0.99$\pm$1e-4		&0.99$\pm$7e-5		&0.99$\pm$4e-5		&0.98$\pm$1e-4		&0.99$\pm$8e-5		&0.99$\pm$9e-5		\\
							&CS		&0.07$\pm$3e-3		&0.04$\pm$2e-3		&0.13$\pm$3e-3		&0.01$\pm$6e-4		&0.08$\pm$3e-3		&0.07$\pm$3e-3		\\
\hline\hline
\multirow{4}{*}{Zdt3}		&$\gamma$		&0.16$\pm$2e-2		&0.15$\pm$1e-2		&0.17$\pm$10e-3		&0.19$\pm$1e-2		&0.35$\pm$3e-2		&0.24$\pm$2e-2		\\
							&$\Delta$		&0.67$\pm$1e-2		&0.85$\pm$1e-2		&0.57$\pm$2e-2		&0.49$\pm$1e-2		&0.33$\pm$7e-3		&0.50$\pm$2e-2		\\
							&HV		&0.98$\pm$2e-3		&0.97$\pm$3e-3		&0.97$\pm$2e-3		&0.97$\pm$2e-3		&0.95$\pm$4e-3		&0.97$\pm$3e-3		\\
							&CS		&0.41$\pm$3e-2		&0.36$\pm$3e-2		&0.33$\pm$2e-2		&0.29$\pm$2e-2		&0.12$\pm$2e-2		&0.24$\pm$3e-2		\\
\hline\hline
\multirow{4}{*}{Zdt4}		&$\gamma$		&0.27$\pm$2e-2		&0.32$\pm$2e-2		&0.35$\pm$2e-2		&0.47$\pm$2e-2		&0.69$\pm$2e-2		&0.57$\pm$2e-2		\\
							&$\Delta$		&0.61$\pm$9e-3		&0.59$\pm$1e-2		&0.48$\pm$1e-2		&0.58$\pm$1e-2		&0.53$\pm$1e-2		&0.68$\pm$1e-2		\\
							&HV		&0.84$\pm$9e-3		&0.80$\pm$1e-2		&0.79$\pm$1e-2		&0.73$\pm$1e-2		&0.62$\pm$1e-2		&0.66$\pm$2e-2		\\
							&CS		&0.73$\pm$3e-2		&0.58$\pm$4e-2		&0.57$\pm$3e-2		&0.41$\pm$4e-2		&0.15$\pm$2e-2		&0.21$\pm$4e-2		\\
\hline\hline
\multirow{4}{*}{Zdt6}		&$\gamma$		&0.06$\pm$3e-2		&0.04$\pm$1e-2		&0.04$\pm$2e-2		&0.04$\pm$1e-2		&0.01$\pm$3e-3		&0.03$\pm$2e-2		\\
							&$\Delta$		&0.52$\pm$1e-2		&0.24$\pm$1e-2		&0.27$\pm$1e-2		&0.37$\pm$9e-3		&0.18$\pm$8e-3		&0.37$\pm$2e-2		\\
							&HV		&0.96$\pm$1e-2		&0.98$\pm$7e-3		&0.98$\pm$10e-3		&0.97$\pm$7e-3		&0.99$\pm$1e-3		&0.98$\pm$10e-3		\\
							&CS		&0.18$\pm$2e-2		&0.11$\pm$2e-2		&0.18$\pm$2e-2		&0.10$\pm$1e-2		&0.21$\pm$2e-2		&0.21$\pm$2e-2		\\
\hline
\end{tabular}}
\end{table}

\begin{table}
\centering
\caption{Problemwise comparison of the algorithms on the four performance metrics used, for the DTLZ family. The values reported represent the mean and standard error obtained for each combination of algorithm, problem and performance metric.}
\label{tab:problemwise2}
\resizebox{\textwidth}{!}{%
\begin{tabular}{l|c|c|c|c|c|c|c}
\hline
Problem	 							&Metric 			&NSGA-II 	&$\eps$-MOEA	&cone$\eps$-MOEA 	&C-NSGA-II 		&SPEA2		&NSGA-II*\\
\hline
\multirow{4}{*}{Dtlz1}		&$\gamma$		&0.23$\pm$7e-3		&0.13$\pm$1e-3		&0.17$\pm$2e-2		&0.39$\pm$1e-2		&0.18$\pm$2e-3		&0.17$\pm$2e-3		\\
							&$\Delta$		&0.34$\pm$3e-3		&0.12$\pm$2e-3		&0.05$\pm$1e-2		&0.20$\pm$2e-2		&0.08$\pm$1e-3		&0.34$\pm$4e-3		\\
							&HV		&0.95$\pm$6e-4		&0.92$\pm$4e-4		&0.95$\pm$2e-4		&0.96$\pm$5e-4		&0.97$\pm$1e-4		&0.96$\pm$5e-4		\\
							&CS		&0.02$\pm$8e-4		&0.01$\pm$9e-4		&0.03$\pm$1e-3		&0.00$\pm$5e-4		&0.02$\pm$1e-3		&0.02$\pm$10e-4		\\
\hline\hline
\multirow{4}{*}{Dtlz2}		&$\gamma$		&0.62$\pm$10e-3		&0.70$\pm$7e-3		&0.48$\pm$8e-3		&0.75$\pm$1e-2		&0.55$\pm$8e-3		&0.48$\pm$6e-3		\\
							&$\Delta$		&0.81$\pm$10e-3		&0.42$\pm$4e-3		&0.42$\pm$6e-3		&0.29$\pm$5e-3		&0.16$\pm$3e-3		&0.83$\pm$9e-3		\\
							&HV		&0.89$\pm$9e-4		&0.92$\pm$4e-4		&0.94$\pm$1e-4		&0.90$\pm$7e-4		&0.93$\pm$4e-4		&0.89$\pm$9e-4		\\
							&CS		&0.03$\pm$1e-3		&0.02$\pm$8e-4		&0.06$\pm$2e-3		&0.01$\pm$6e-4		&0.03$\pm$1e-3		&0.04$\pm$1e-3		\\
\hline\hline
\multirow{4}{*}{Dtlz3}		&$\gamma$		&0.35$\pm$2e-2		&0.25$\pm$1e-2		&0.42$\pm$3e-2		&0.50$\pm$3e-2		&0.32$\pm$2e-2		&0.26$\pm$1e-2		\\
							&$\Delta$		&0.33$\pm$9e-3		&0.19$\pm$1e-2		&0.29$\pm$2e-2		&0.22$\pm$3e-2		&0.15$\pm$2e-2		&0.34$\pm$8e-3		\\
							&HV		&0.90$\pm$10e-4		&0.91$\pm$2e-2		&0.91$\pm$1e-2		&0.91$\pm$1e-3		&0.93$\pm$3e-4		&0.90$\pm$9e-4		\\
							&CS		&0.02$\pm$1e-3		&0.03$\pm$2e-3		&0.04$\pm$2e-3		&0.00$\pm$5e-4		&0.02$\pm$2e-3		&0.02$\pm$2e-3		\\
\hline\hline
\multirow{4}{*}{Dtlz4}		&$\gamma$		&0.32$\pm$8e-3		&0.41$\pm$2e-2		&0.53$\pm$3e-2		&0.34$\pm$1e-2		&0.33$\pm$1e-2		&0.30$\pm$3e-3		\\
							&$\Delta$		&0.67$\pm$2e-2		&0.37$\pm$3e-2		&0.43$\pm$2e-2		&0.36$\pm$4e-2		&0.22$\pm$3e-2		&0.66$\pm$8e-3		\\
							&HV		&0.88$\pm$1e-2		&0.86$\pm$2e-2		&0.86$\pm$2e-2		&0.84$\pm$2e-2		&0.87$\pm$2e-2		&0.90$\pm$7e-4		\\
							&CS		&0.04$\pm$2e-3		&0.02$\pm$1e-3		&0.03$\pm$2e-3		&0.02$\pm$2e-3		&0.03$\pm$2e-3		&0.03$\pm$1e-3		\\
\hline\hline
\multirow{4}{*}{Dtlz5}		&$\gamma$		&0.14$\pm$2e-3		&0.26$\pm$3e-3		&0.56$\pm$3e-2		&0.22$\pm$5e-3		&0.15$\pm$2e-3		&0.13$\pm$1e-3		\\
							&$\Delta$		&0.74$\pm$2e-2		&0.78$\pm$5e-3		&0.83$\pm$9e-3		&0.43$\pm$6e-3		&0.26$\pm$4e-3		&0.61$\pm$1e-2		\\
							&HV		&0.99$\pm$1e-4		&0.99$\pm$7e-5		&0.98$\pm$3e-5		&0.98$\pm$1e-4		&0.99$\pm$4e-5		&0.99$\pm$1e-4		\\
							&CS		&0.06$\pm$2e-3		&0.02$\pm$1e-3		&0.04$\pm$1e-3		&0.03$\pm$1e-3		&0.05$\pm$2e-3		&0.07$\pm$2e-3		\\
\hline\hline
\multirow{4}{*}{Dtlz6}		&$\gamma$		&0.84$\pm$8e-3		&0.94$\pm$3e-3		&0.83$\pm$3e-3		&0.83$\pm$5e-3		&0.84$\pm$7e-3		&0.84$\pm$8e-3		\\
							&$\Delta$		&0.78$\pm$1e-2		&0.99$\pm$6e-4		&0.45$\pm$6e-4		&0.51$\pm$7e-3		&0.32$\pm$5e-3		&0.62$\pm$1e-2		\\
							&HV		&0.99$\pm$1e-4		&0.99$\pm$4e-5		&0.99$\pm$2e-5		&0.99$\pm$9e-5		&0.99$\pm$3e-5		&0.99$\pm$1e-4		\\
							&CS		&0.02$\pm$6e-4		&0.03$\pm$8e-4		&0.04$\pm$8e-4		&0.03$\pm$8e-4		&0.03$\pm$8e-4		&0.01$\pm$7e-4		\\
\hline\hline
\multirow{4}{*}{Dtlz7}		&$\gamma$		&0.74$\pm$9e-3		&0.47$\pm$2e-3		&0.74$\pm$10e-3		&0.86$\pm$9e-3		&0.67$\pm$8e-3		&0.74$\pm$10e-3		\\
							&$\Delta$		&0.78$\pm$1e-2		&0.55$\pm$9e-3		&0.71$\pm$9e-3		&0.56$\pm$1e-2		&0.52$\pm$8e-3		&0.78$\pm$8e-3		\\
							&HV		&0.92$\pm$1e-3		&0.91$\pm$1e-3		&0.93$\pm$7e-4		&0.91$\pm$1e-3		&0.94$\pm$6e-4		&0.92$\pm$9e-4		\\
							&CS		&0.06$\pm$2e-3		&0.08$\pm$2e-3		&0.06$\pm$2e-3		&0.02$\pm$1e-3		&0.09$\pm$3e-3		&0.07$\pm$2e-3		\\
\hline\hline
\multirow{4}{*}{Dtlz8}		&$\gamma$		&0.45$\pm$2e-2		&0.03$\pm$10e-4		&0.24$\pm$5e-3		&0.52$\pm$1e-2		&0.83$\pm$9e-3		&0.58$\pm$1e-2		\\
							&$\Delta$		&0.82$\pm$9e-3		&0.69$\pm$9e-3		&0.70$\pm$9e-3		&0.49$\pm$8e-3		&0.36$\pm$8e-3		&0.80$\pm$1e-2		\\
							&CS		&0.11$\pm$4e-3		&0.20$\pm$3e-3		&0.24$\pm$4e-3		&0.10$\pm$3e-3		&0.10$\pm$3e-3		&0.09$\pm$3e-3		\\
\hline\hline
\multirow{4}{*}{Dtlz9}		&$\gamma$		&0.90$\pm$5e-3		&0.16$\pm$7e-3		&0.57$\pm$5e-3		&0.64$\pm$7e-3		&0.77$\pm$7e-3		&0.90$\pm$5e-3		\\
							&$\Delta$		&0.62$\pm$5e-3		&0.71$\pm$1e-2		&0.52$\pm$6e-3		&0.49$\pm$6e-3		&0.45$\pm$4e-3		&0.61$\pm$6e-3		\\
							&CS		&0.11$\pm$8e-3		&0.51$\pm$1e-2		&0.34$\pm$7e-3		&0.23$\pm$8e-3		&0.24$\pm$8e-3		&0.13$\pm$7e-3		\\
\hline
\end{tabular}}
\end{table}

The Deb52 problem tests the ability of the MOEA to find non-convex Pareto-optimal solutions \citep{journal.Deb1999}. As this simple and continuous problem has only two decision variables, all algorithms presented a similar performance in terms of the $\gamma$, $HV$ and $CS$ measures. On the other hand, the SPEA2 achieved the best $\Delta$ value, followed by cone$\eps$-MOEA, C-NSGA-II, NSGA-II* and NSGA-II, respectively. Since the optimal frontier of this problem is represented by a ``strong'' concave curve, and the $\eps$-mechanism is sensitive to the loss of feasible solutions, the $\eps$-MOEA ranked last in this metric. 

The Pol problem, although simple, provides difficulties by introducing discontinuities in the non-convex Pareto front \citep{proc.Poloni1995}. As the parameter space is also mapped with two variables, very small differences were observed for the $HV$ and $CS$ indicators. The cone$\eps$-MOEA found the second best $\Delta$ value, and a convergence measure similar to that of the approaches based on the NSGA-II. As an illustration, Fig. \ref{fig:viewbox} gives a good idea of the convergence and distribution of the solutions accomplished by both $\eps$-approaches in a particular case.

The ZDT1 30-variable problem has a convex Pareto-optimal front \citep{journal.Zitzler2000}. The NSGA-II approaches achieved the best $\gamma$ and $CS$ values. The cone$\eps$-MOEA also ranked well in these metrics. Because of the cone$\eps$-concept, this method achieved the second best spread of solutions. The SPEA2 obtained the overall best value in this metric. The ZDT2 30-variable problem represents the counterpart of ZDT1, and tests the ability of the MOEA to approximate non-convex Pareto-optimal solutions \citep{journal.Zitzler2000}. The cone$\eps$-MOEA has significantly outperformed the other methods in the $\gamma$ and $CS$ indicators. Regarding the diversity found, the SPEA2 and the cone$\eps$-MOEA were the two best ones. No significant differences were observed for the $HV$ indicator in both problems.

As we have observed for the Pol test, the ZDT3 problem provides difficulties by introducing discontinuities in the objective space, with a Pareto front consisting of several disconnected convex parts. However, the Pareto-optimal set in the 30-variable parameter space has no discontinuity \citep{journal.Zitzler2000}. The cone$\eps$-MOEA ranked third in the $\gamma$ and $CS$ measures. The $\eps$-MOEA and the SPEA2 presented either a good convergence or diversity: the $\eps$-MOEA found the best $\gamma$ value and the worst $\Delta$ measure; on the other hand, the SPEA2 ranked first for the diversity measure at the cost of the worst $\gamma$ value.

The ZDT4 10-variable problem contains a number of local Pareto-optimal fronts and, therefore, tests the ability of the MOEA to deal with multimodality, providing challenges for a MOEA to converge to the global Pareto front \citep{journal.Zitzler2000}. In this problem, the cone$\eps$-MOEA has achieved the best spread of solutions. On the other hand, the NSGA-II performed better in terms of convergence, $HV$ and $CS$, followed by the $\eps$-MOEA and cone$\eps$-MOEA in these metrics. 

In the ZDT6 10-variable problem, the Pareto-optimal solutions are nonuniformly distributed along the global Pareto front \citep{journal.Zitzler2000}. Essentially, very small differences were observed for both convergence and hypervolume metrics. The SPEA2 presented the best diversity, followed by the $\eps$-MOEA and the cone$\eps$-MOEA. For the $CS$ indicator, the $\eps$-MOEA and the C-NSGA-II ranked last.

\begin{figure*}[!ht]
\centering
\psfrag{f1}[][]{$f_{1}$}
\psfrag{f2}[][]{$f_{2}$}
\psfrag{f3}[][]{$f_{3}$}
\psfrag{Fronteira Pareto Global}[][]{\scriptsize Pareto front}
\psfrag{cone-MOEAMOEA}[][]{\tiny c$\eps$-MOEA}
\psfrag{e-MOEAMOEA}[][]{\tiny $\eps$-MOEA}
\psfrag{NSGANSGA-II}[][]{\tiny NSGA-II}
\psfrag{NSGANSGA-II-s}[][]{\tiny NSGA-II*}
\psfrag{CNSGANSGA-II}[][]{\tiny C-NSGA-II}
\psfrag{SPEA2-SPEA2}[][]{\tiny SPEA2}
\centering\subfigure[\small NSGA-II.]{\includegraphics[width=4.4cm]{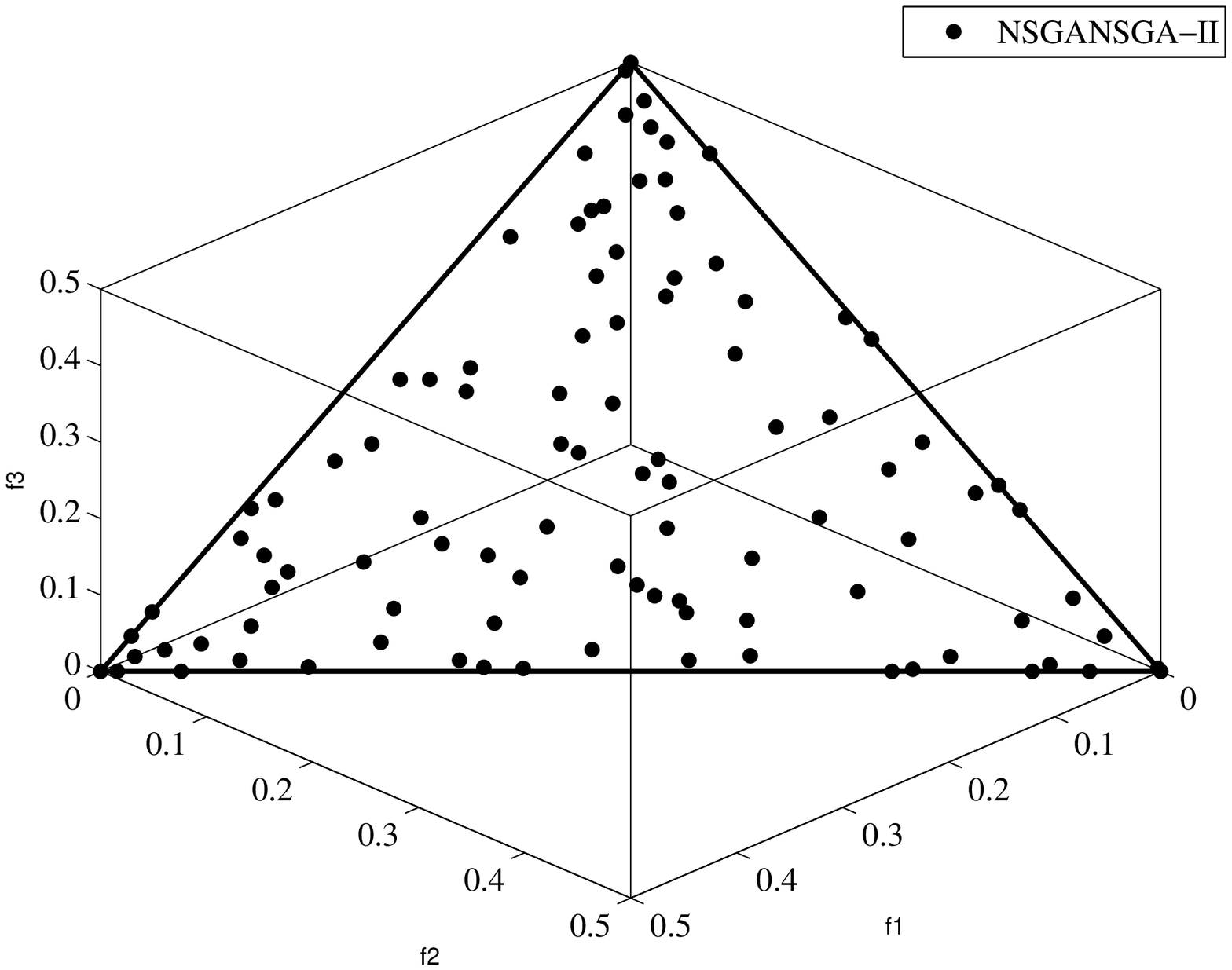}}
\centering\subfigure[\small NSGA-II*.]{\includegraphics[width=4.4cm]{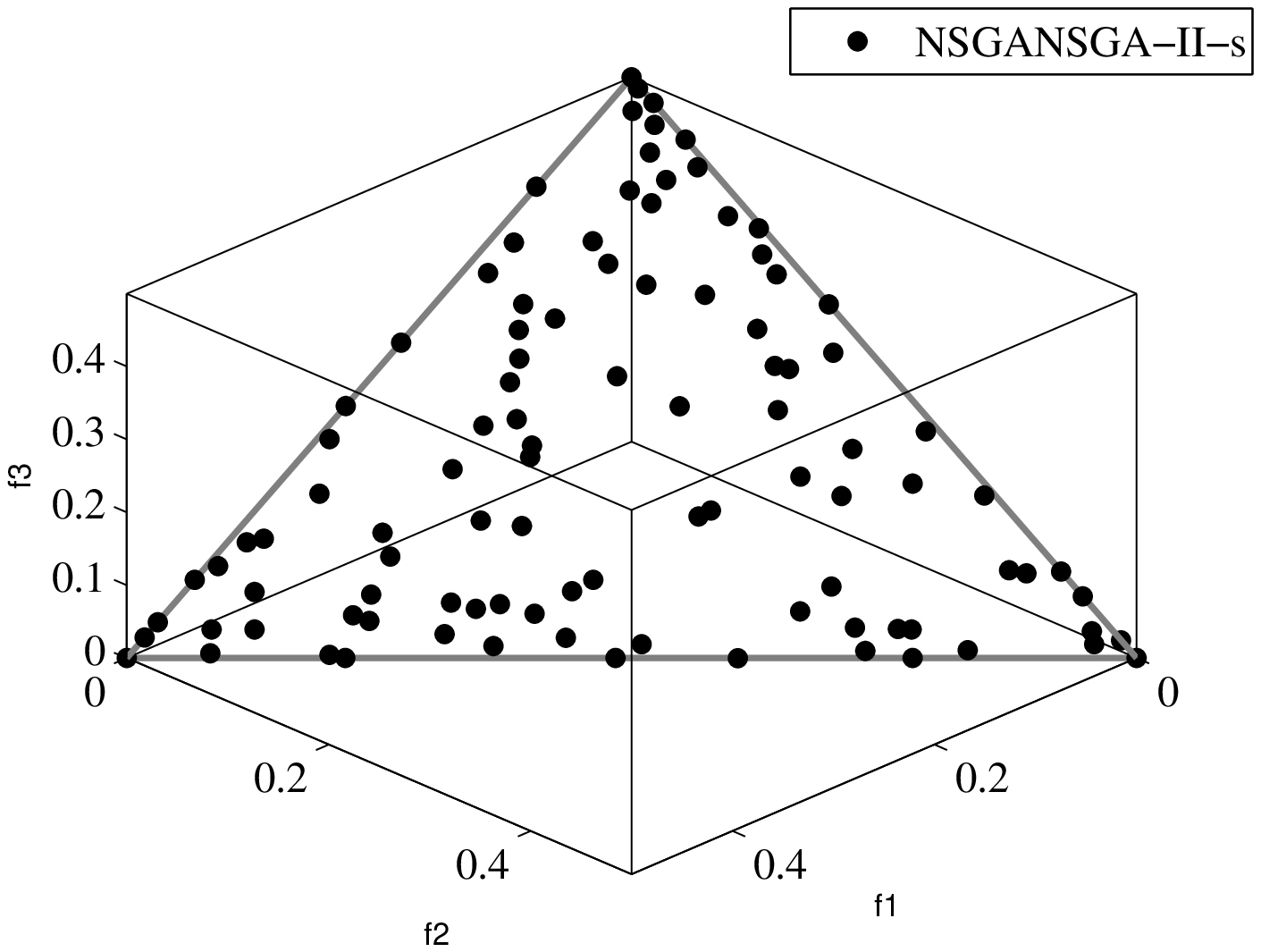}}
\centering\subfigure[\small C-NSGA-II.]{\includegraphics[width=4.4cm]{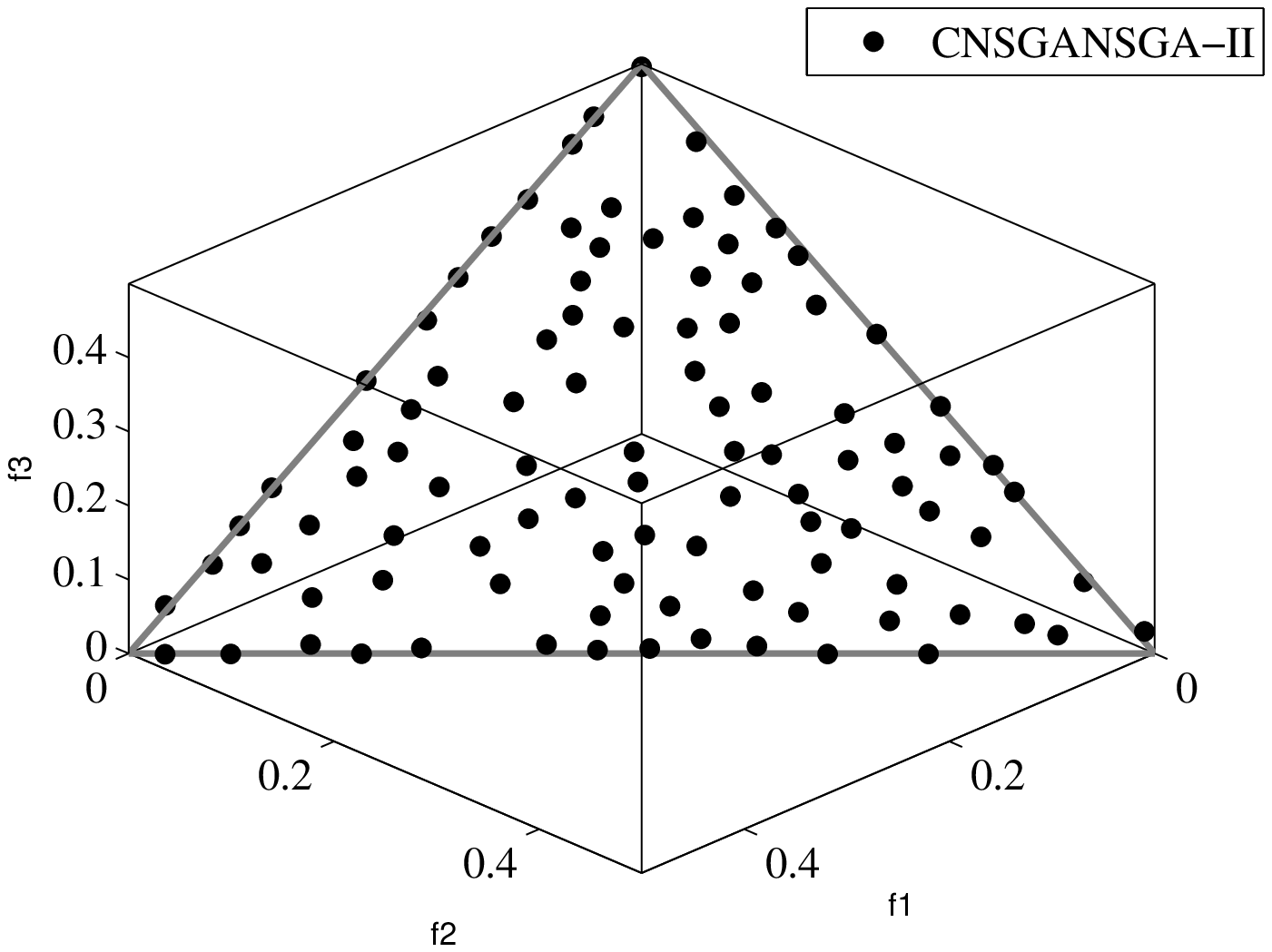}}
\centering\subfigure[\small SPEA2.]{\includegraphics[width=4.4cm]{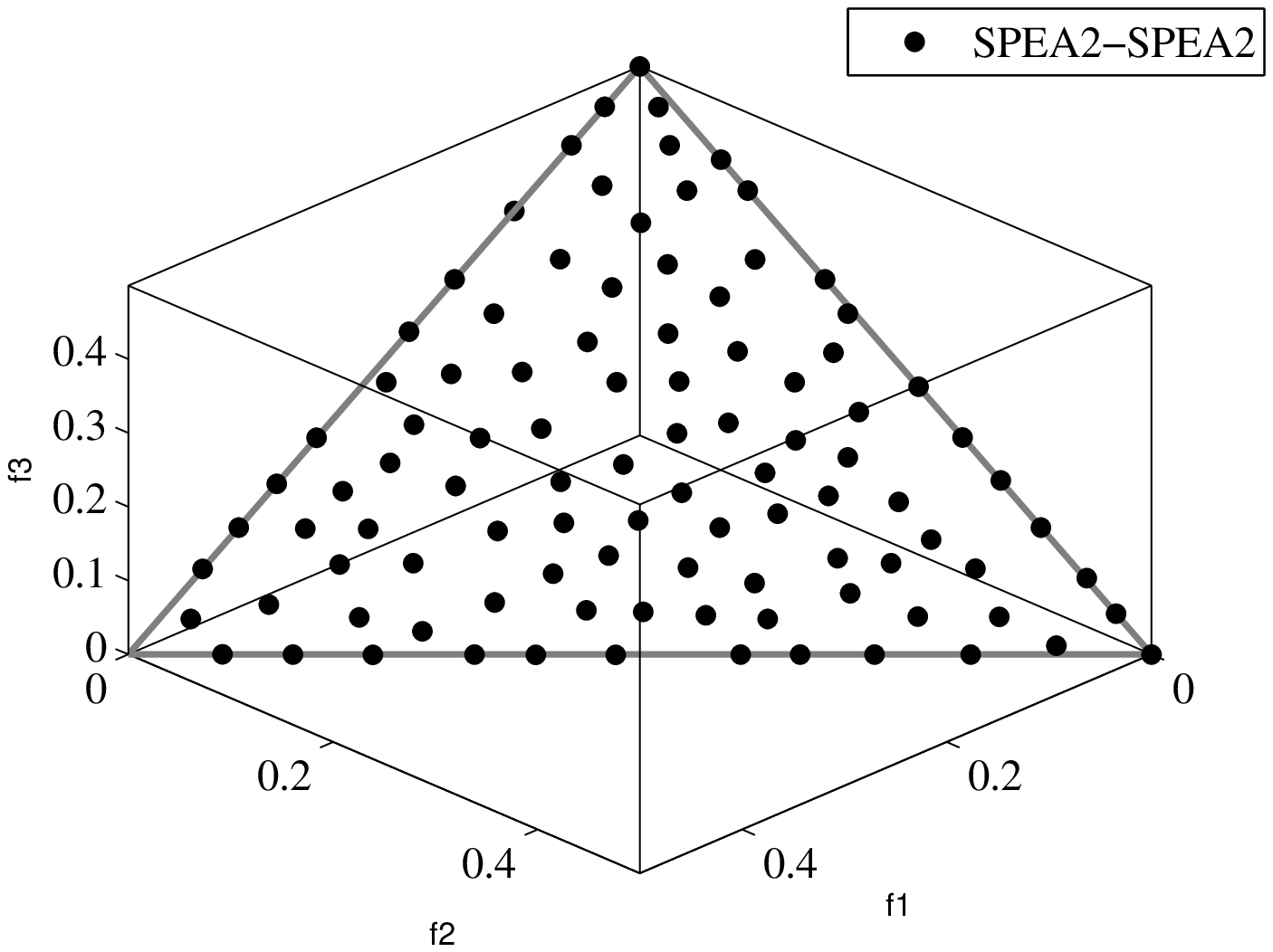}}
\centering\subfigure[\small $\eps$-MOEA.]{\includegraphics[width=4.4cm]{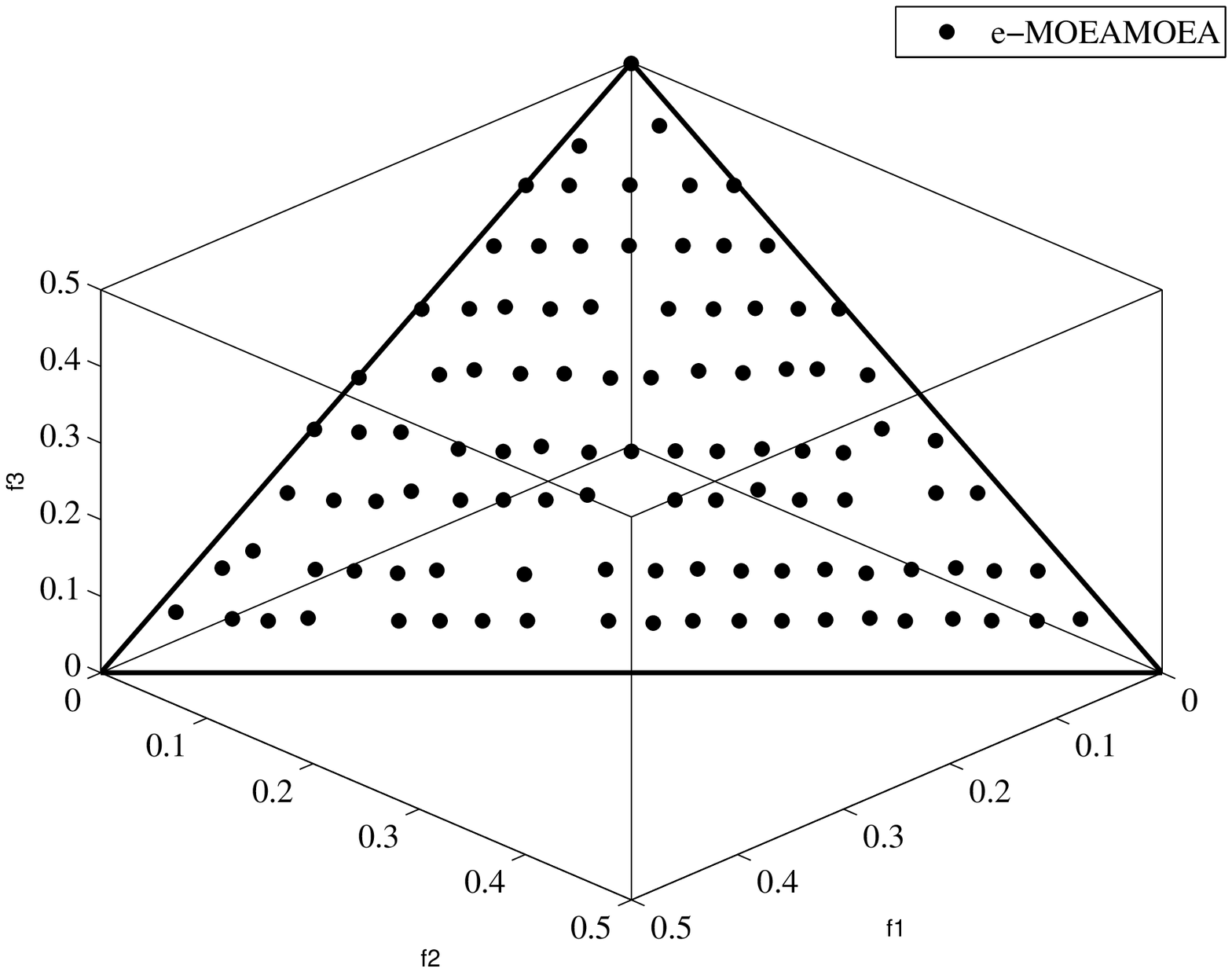}}
\centering\subfigure[\small cone$\eps$-MOEA.]{\includegraphics[width=4.4cm]{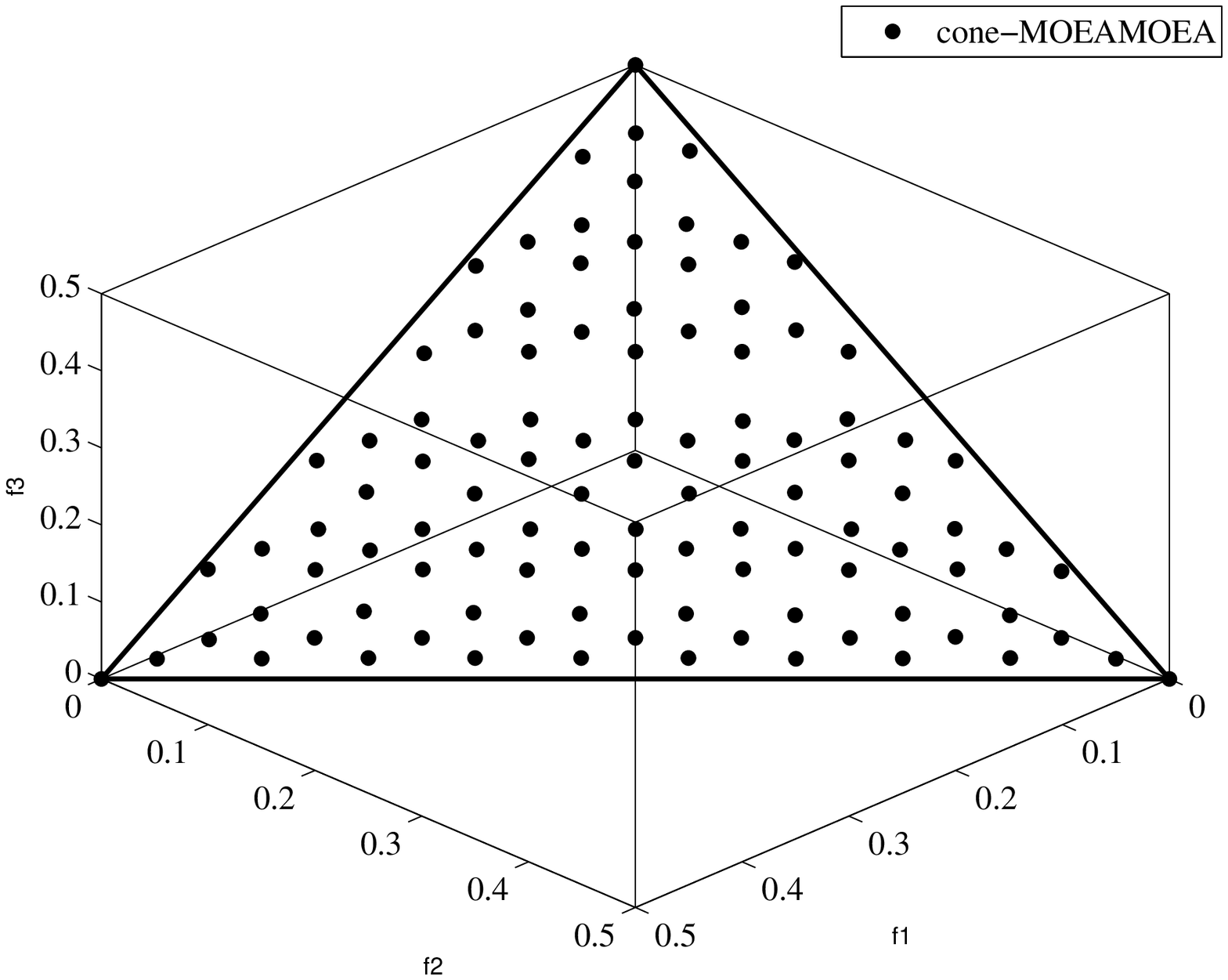}}
\caption{Efficient solutions generated for problem DTLZ1, considering the estimated $\pmb{\eps}$ values provided in Table \ref{tab:eps_values}. The fronts presented are the outcome of a typical run.}
\label{fig:viewDTLZ1}
\end{figure*}

The three-objective DTLZ1 test problem has seven decision variables, and the Pareto-optimal solutions lie on a plane satisfying $\sum_{i=1}^{m} f_{i} = 0.5$, such that $f_{i} \in [0,0.5]$ for all $i = \left\{1,2,3\right\}$ \citep{proc.Deb2005}. Small differences were observed for the algorithms regarding the $HV$ and $CS$ metrics. The $\eps$-MOEA has achieved the best convergence, followed by the cone$\eps$-MOEA and the NSGA-II*, and the cone$\eps$-method has outperformed the other algorithms in finding a well-distributed set of solutions. The $\eps$-MOEA was better than the NSGA-II -based methods in terms of diversity, possibly due to the use of the $\eps$-dominance, which tends to avoid the clumping of solutions. Figure \ref{fig:viewDTLZ1} presents a visual comparison between the outcome of a typical run of each MOEA considered in this study. It can be seen that the distribution of solutions obtained by the NSGA-II and NSGA-II* are poorer than the other MOEAs.

\begin{figure*}[!ht]
\centering
\psfrag{f1}[][]{$f_{1}$}
\psfrag{f2}[][]{$f_{2}$}
\psfrag{f3}[][]{$f_{3}$}
\psfrag{Fronteira Pareto Global}[][]{\scriptsize Pareto front}
\psfrag{cone-MOEAMOEA}[][]{\tiny c$\eps$-MOEA}
\psfrag{e-MOEAMOEA}[][]{\tiny $\eps$-MOEA}
\psfrag{NSGANSGA-II}[][]{\tiny NSGA-II}
\psfrag{NSGANSGA-II-s}[][]{\tiny NSGA-II*}
\psfrag{CNSGANSGA-II}[][]{\tiny C-NSGA-II}
\psfrag{SPEA2-SPEA2}[][]{\tiny SPEA2}
\subfigure[\small NSGA-II.]{\includegraphics[width=4.4cm]{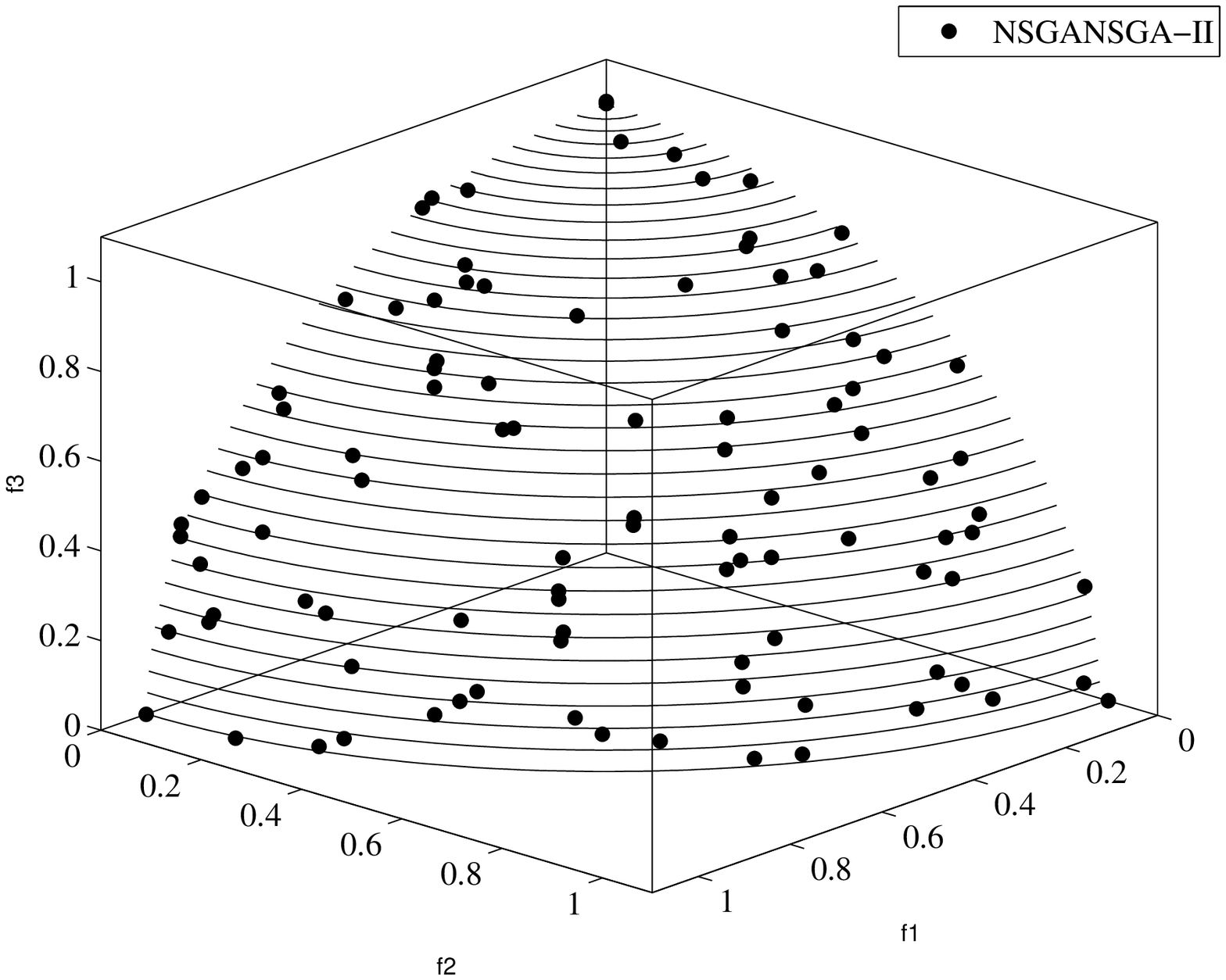}}
\subfigure[\small NSGA-II*.]{\includegraphics[width=4.4cm]{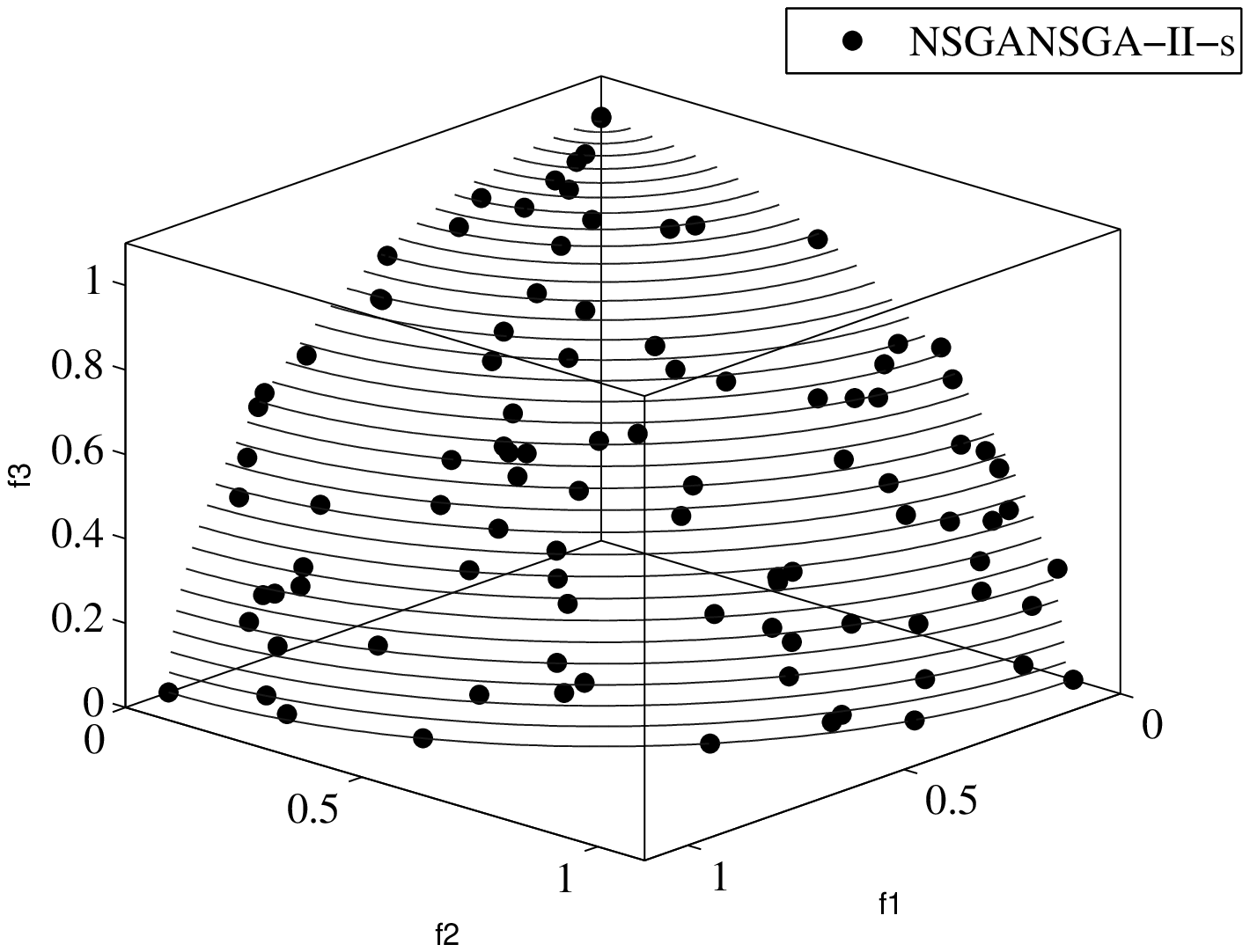}}
\subfigure[\small C-NSGA-II.]{\includegraphics[width=4.4cm]{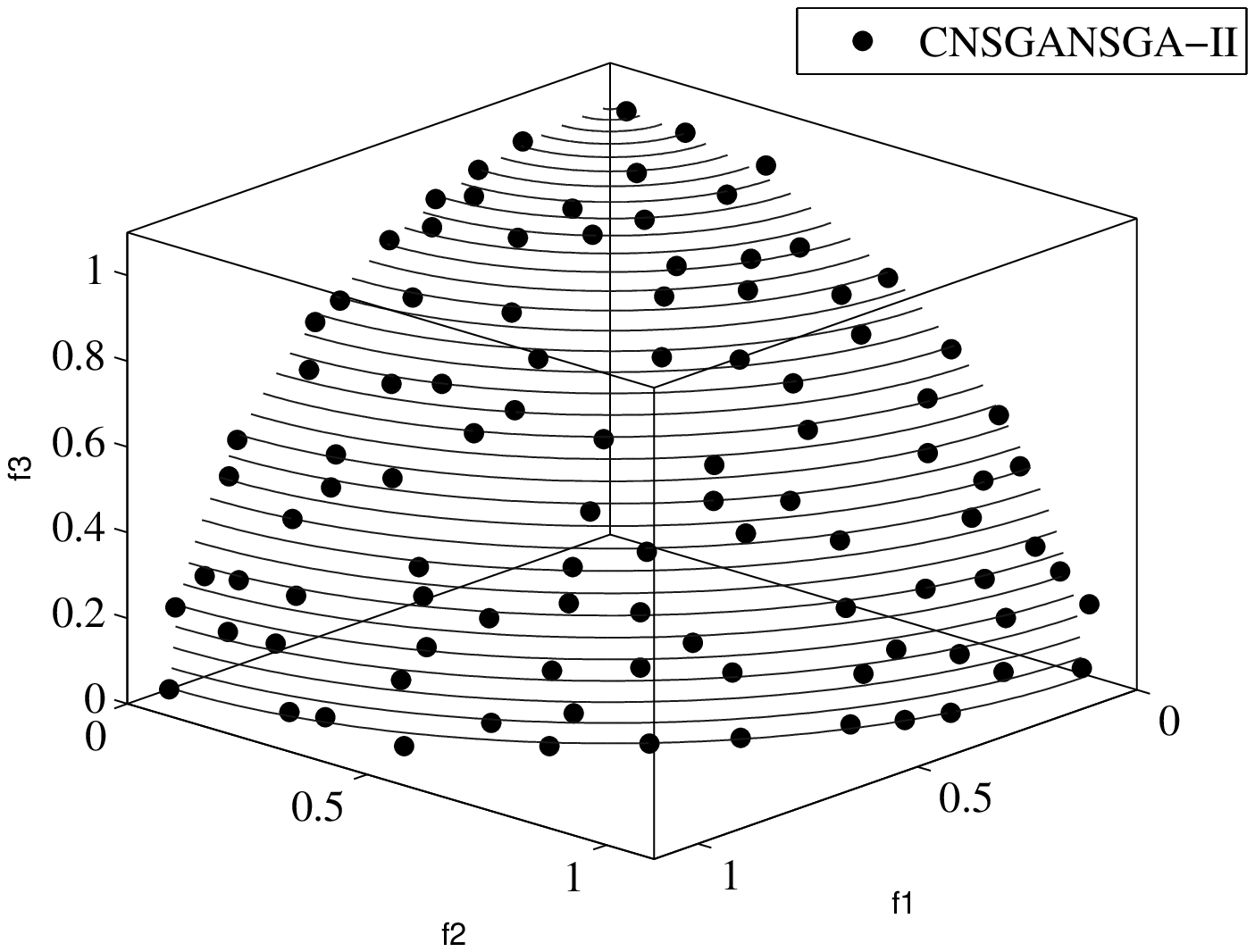}}
\subfigure[\small SPEA2.]{\includegraphics[width=4.4cm]{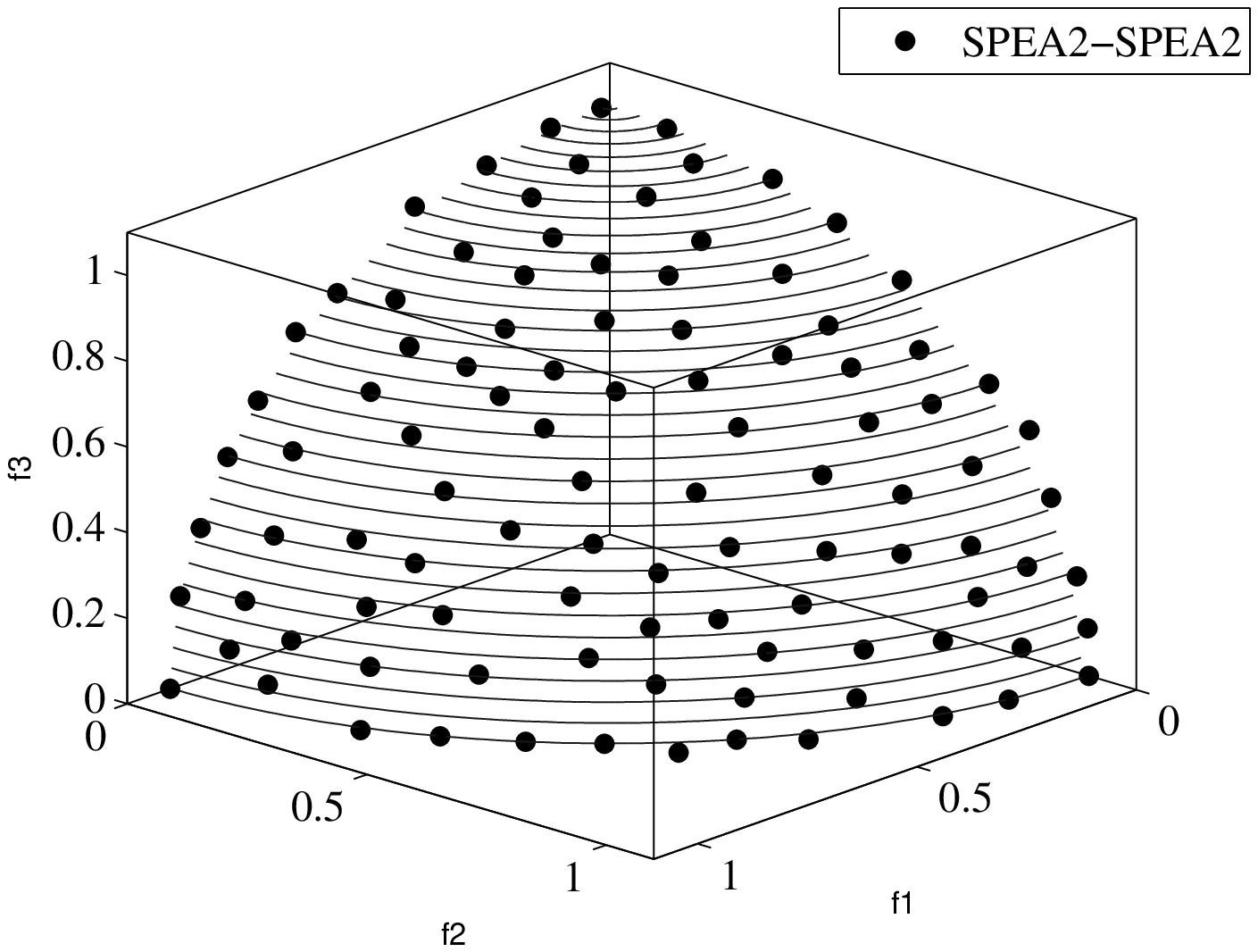}}
\subfigure[\small $\eps$-MOEA.]{\includegraphics[width=4.4cm]{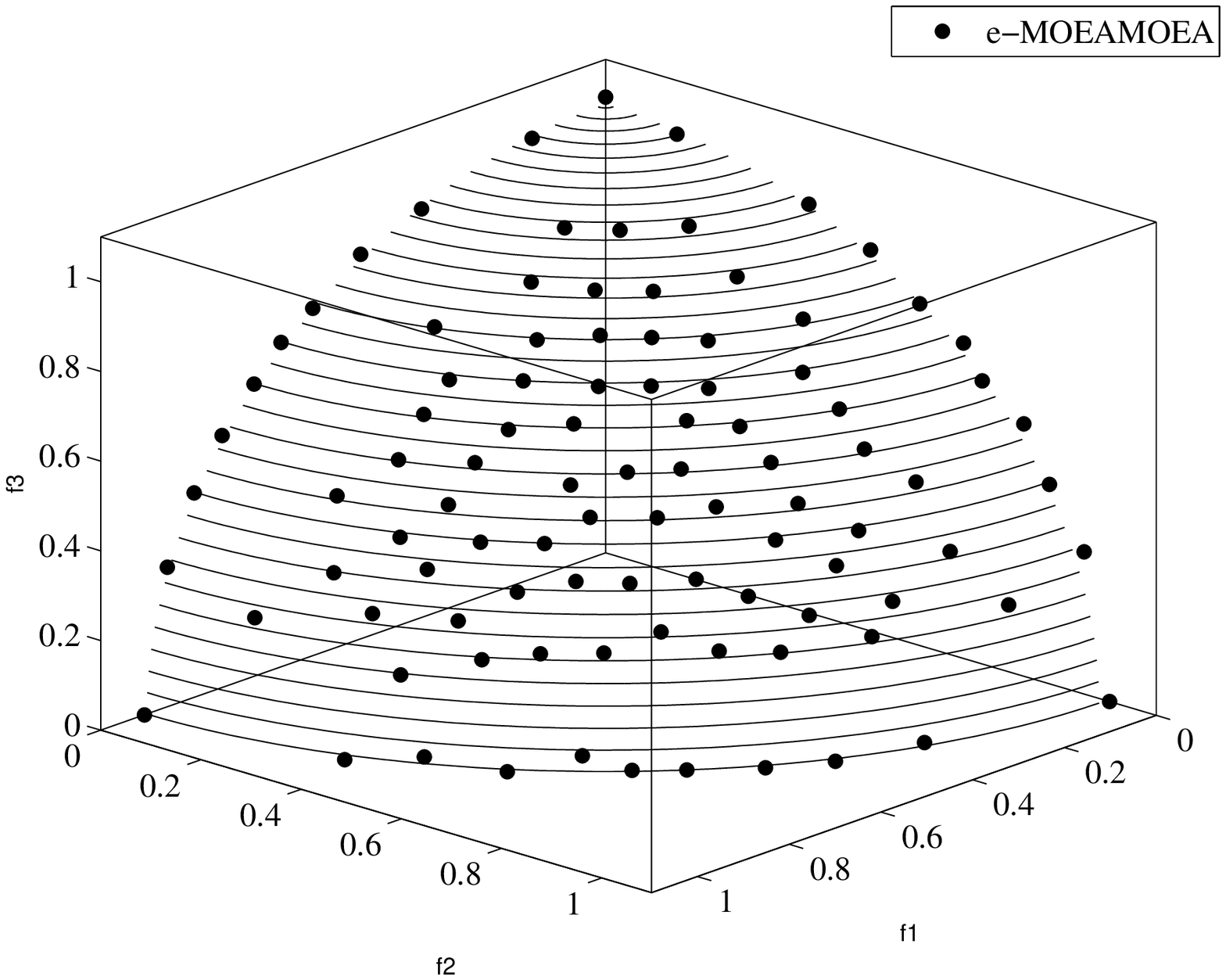}}
\subfigure[\small cone$\eps$-MOEA.]{\includegraphics[width=4.4cm]{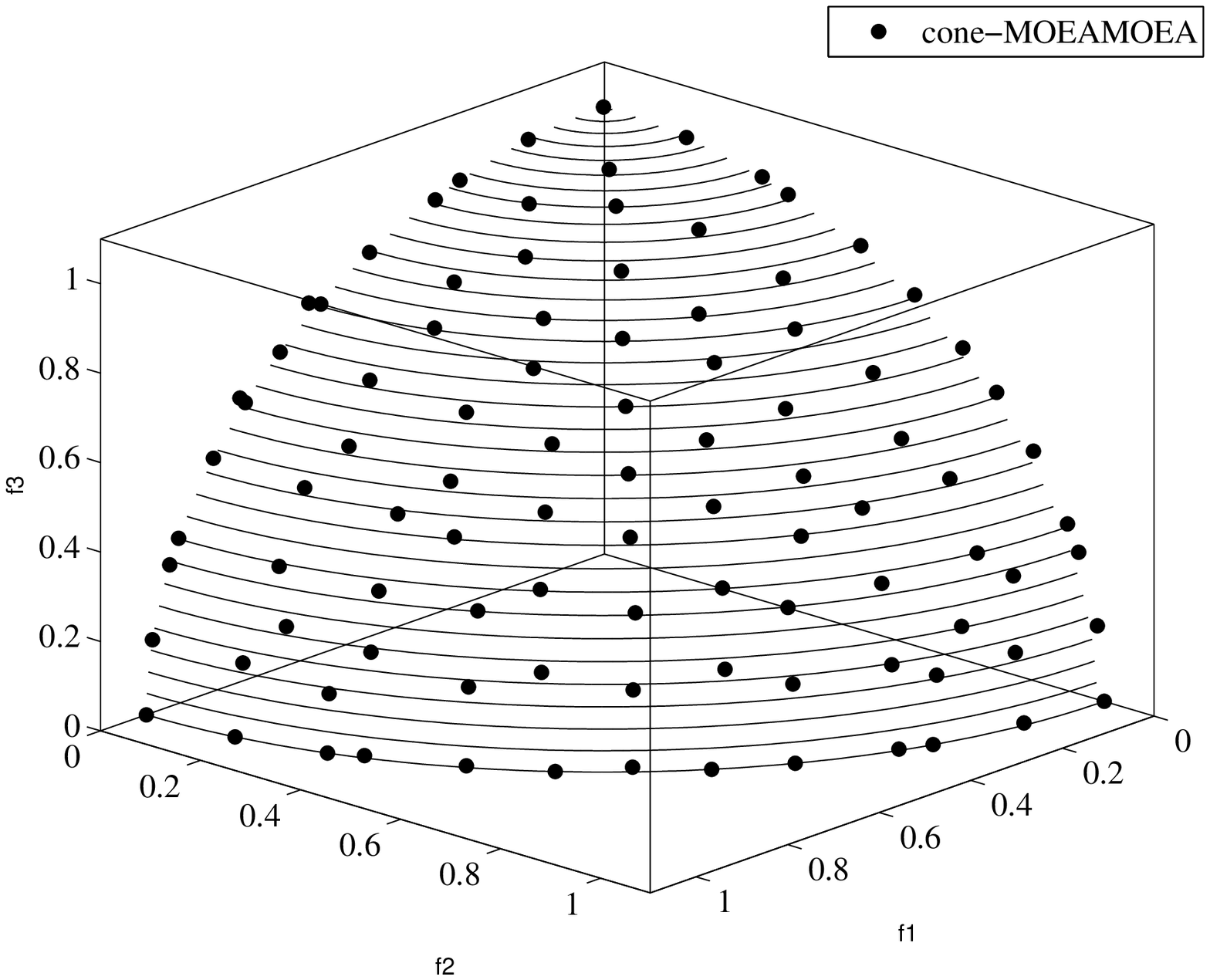}}
\caption{Efficient solutions generated for problem DTLZ2, considering the estimated $\pmb{\eps}$ values provided in Table \ref{tab:eps_values}. The fronts presented are the outcome of a typical run.}
\label{fig:viewDTLZ2}
\end{figure*}

The DTLZ2 12-variable test problem has a spherical Pareto-optimal front satisfying $\sum_{i=1}^{m} f_{i}^{2} = 1$, such that $f_{i} \in [0,1]$ for all $i = \left\{1,2,3\right\}$ \citep{proc.Deb2005}. For the hypervolume and coverage of many sets quality indicators, the use of cone$\eps$-dominance was able to provide a small advantage over the other approaches. Additionally, the cone$\eps$-MOEA and NSGA-II* have achieved the best convergence, and the clustering techniques employed in SPEA2 and C-NSGA-II provided the overall best $\Delta$ values. Although no significant difference was found between the cone$\eps$-MOEA and the $\eps$-MOEA for the diversity metric, it can be seen from Fig. \ref{fig:viewDTLZ2} that the border regions of the Pareto front were better mapped by the cone$\eps$-method. Notice that, even though these fronts are the outcome of a single run, both the $\eps$ and the cone$\eps$-concepts yielded very similar distributions of points in all runs. In the case of the $\eps$-MOEA, there seems to be a considerable gap between the boundary solutions and their nearest neighbors. This happens because of the gentle slope near the boundary solutions on a spherical surface, and the fact that the $\eps$-dominance does not allow any point to be nondominated within an $\eps_{i}$ in the $i^{th}$ objective \citep{proc.Deb2003,journal.Deb2005a}.

The DTLZ3 12-variable test problem has a large number of local Pareto-optimal fronts and investigates the ability of the MOEA to converge to the global Pareto front. As in the previous problem, the global front satisfies $\sum_{i=1}^{m} f_{i}^{2} = 1$, such that $f_{i} \in [0,1]$ for all $i = \left\{1,2,3\right\}$ \citep{proc.Deb2005}. In this problem, the cone$\eps$-MOEA ranked fourth in the $\Delta$ and fifth in the $\gamma$ metric. The SPEA2 and $\eps$-MOEA presented the best overall values in these indicators, respectively. Small differences were observed for the other two quality measures.

\begin{figure*}[!htb]
\centering
\psfrag{f1}[][]{$f_{1}$}
\psfrag{f2}[][]{$f_{2}$}
\psfrag{f3}[][]{$f_{3}$}
\psfrag{Fronteira Pareto Global}[][]{\scriptsize Pareto front}
\psfrag{cone-MOEAMOEA}[][]{\tiny c$\eps$-MOEA}
\psfrag{e-MOEAMOEA}[][]{\tiny $\eps$-MOEA}
\psfrag{NSGANSGA-II}[][]{\tiny NSGA-II}
\psfrag{NSGANSGA-II-s}[][]{\tiny NSGA-II*}
\psfrag{CNSGANSGA-II}[][]{\tiny C-NSGA-II}
\psfrag{SPEA2-SPEA2}[][]{\tiny SPEA2}
\subfigure[\small NSGA-II.]{\includegraphics[width=4.4cm]{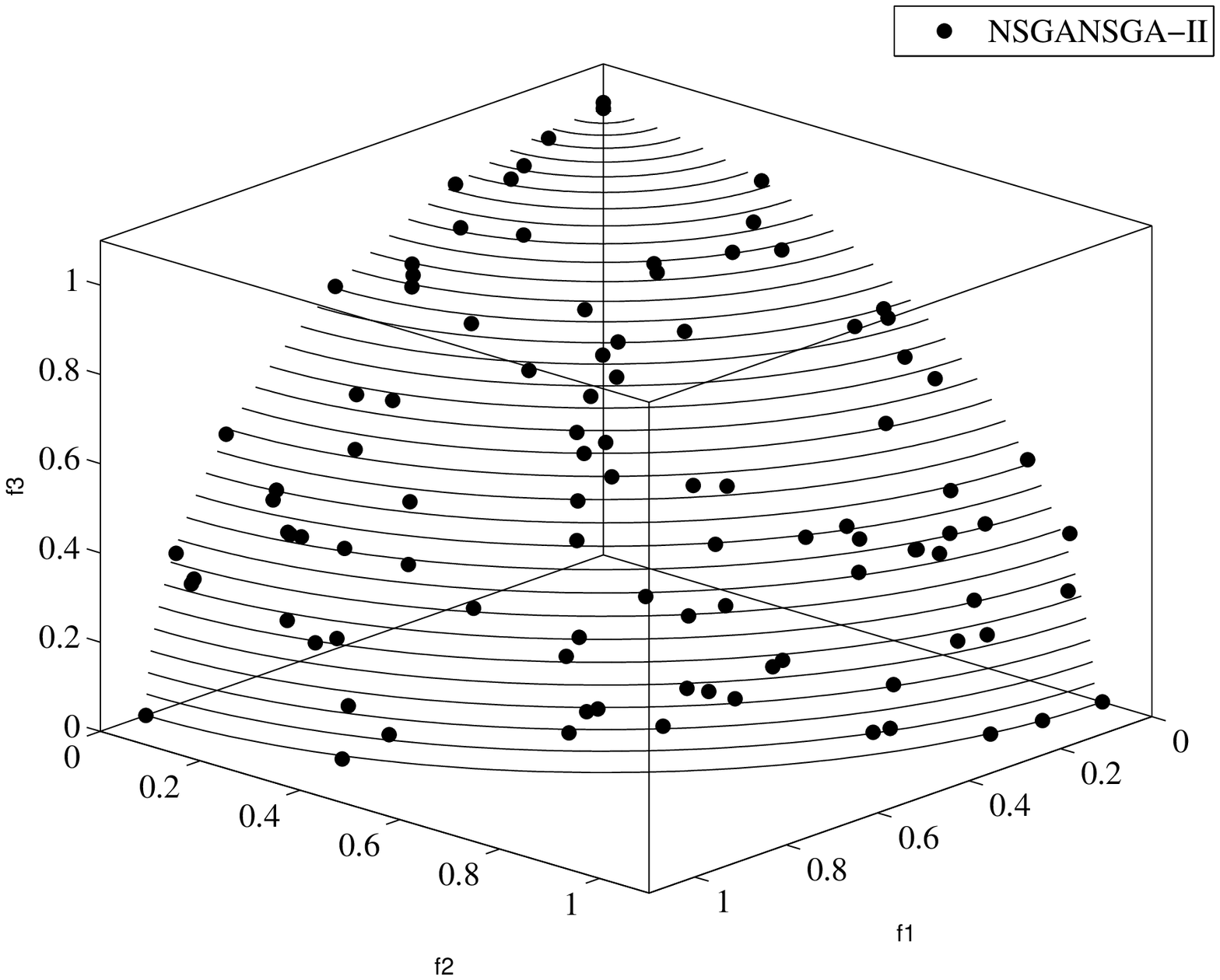}}
\subfigure[\small NSGA-II*.]{\includegraphics[width=4.4cm]{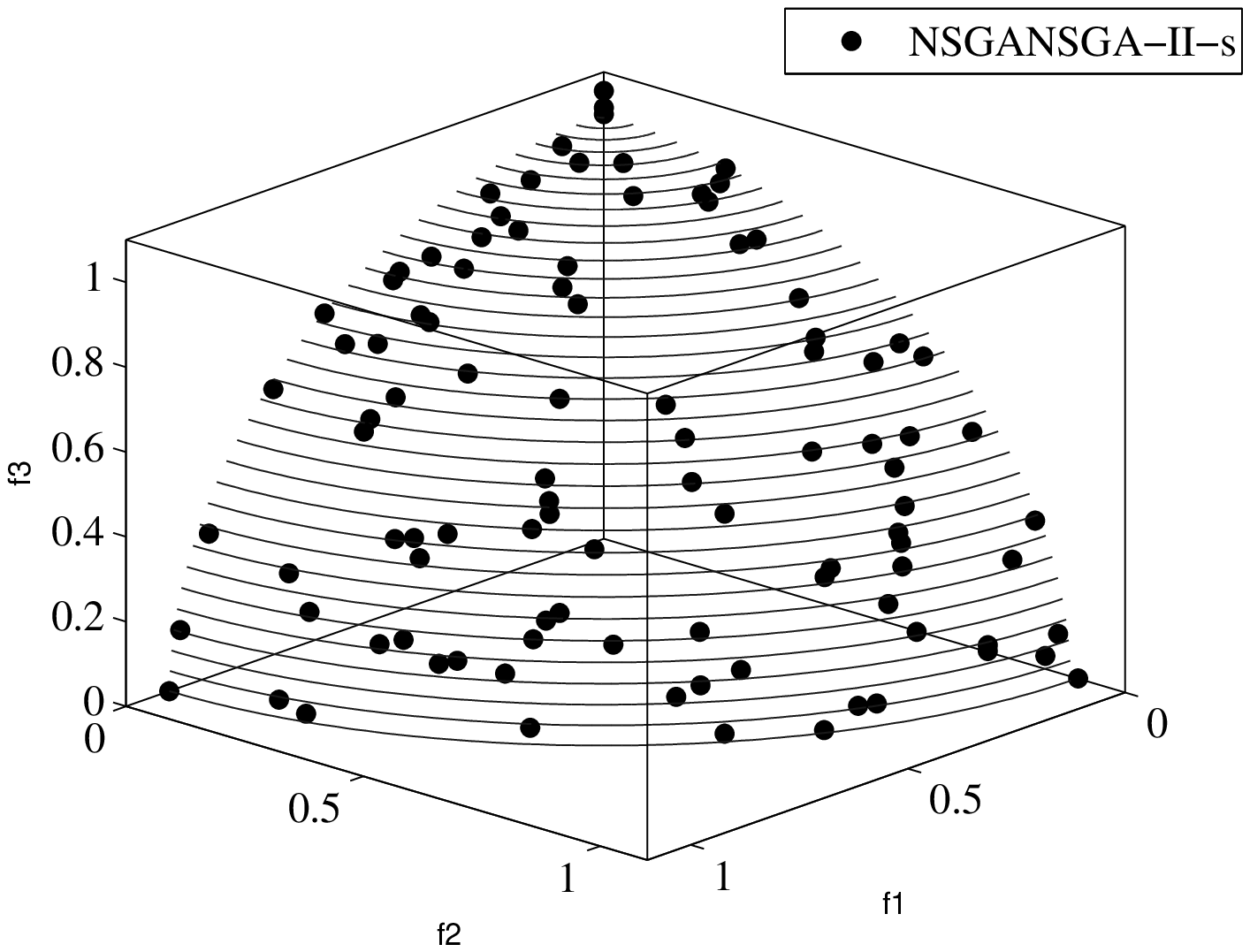}}
\subfigure[\small C-NSGA-II.]{\includegraphics[width=4.4cm]{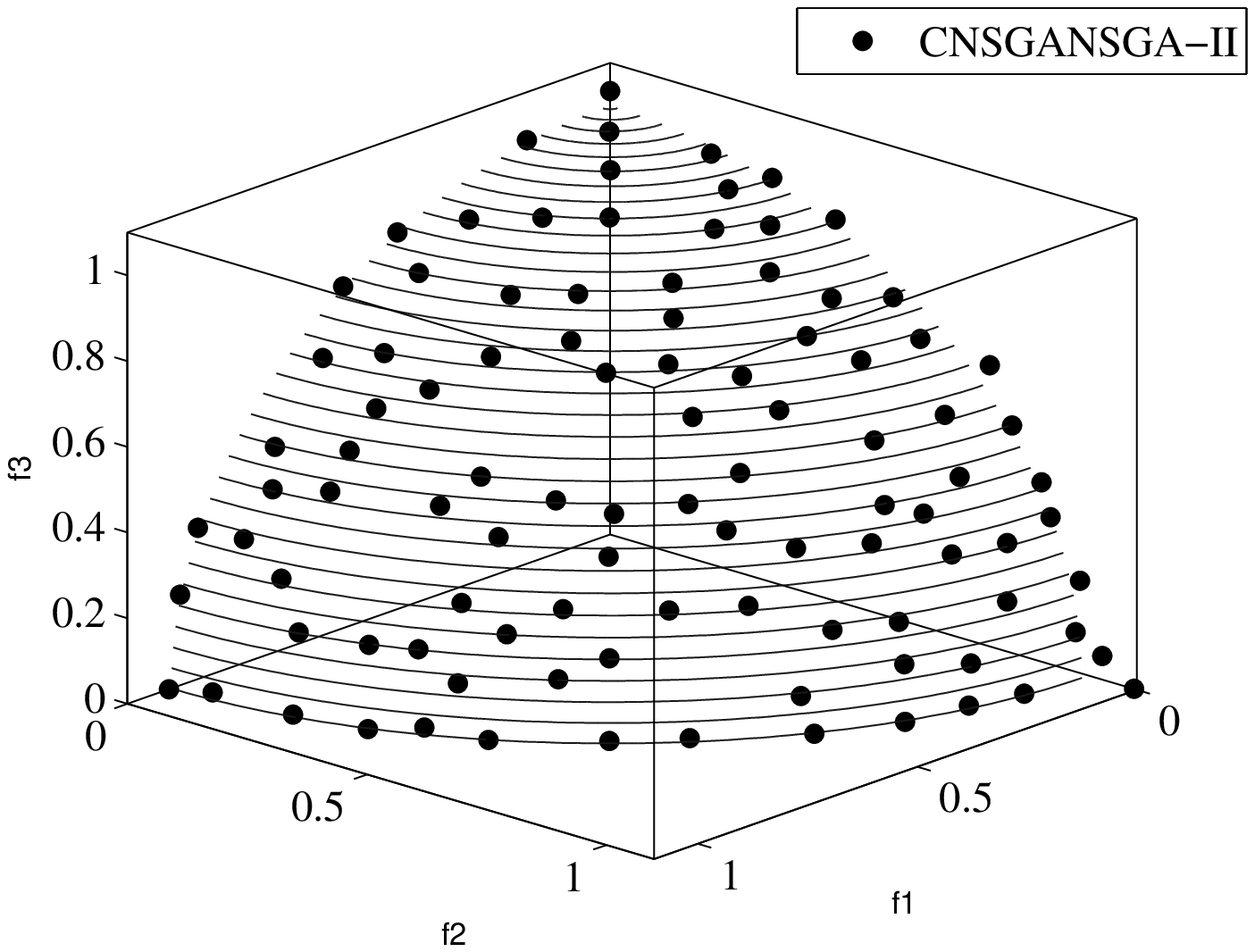}}
\subfigure[\small SPEA2.]{\includegraphics[width=4.4cm]{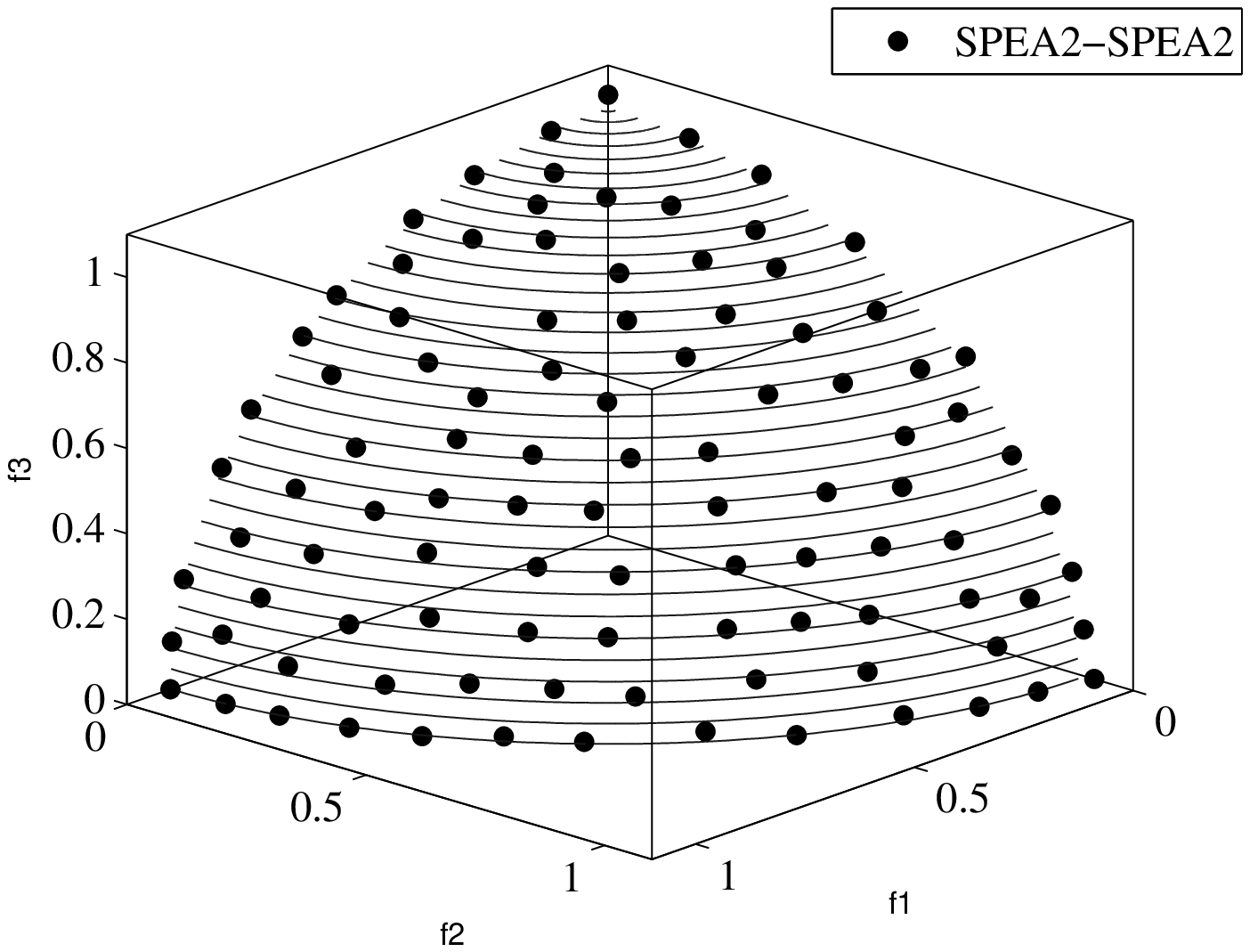}}
\subfigure[\small $\eps$-MOEA.]{\includegraphics[width=4.4cm]{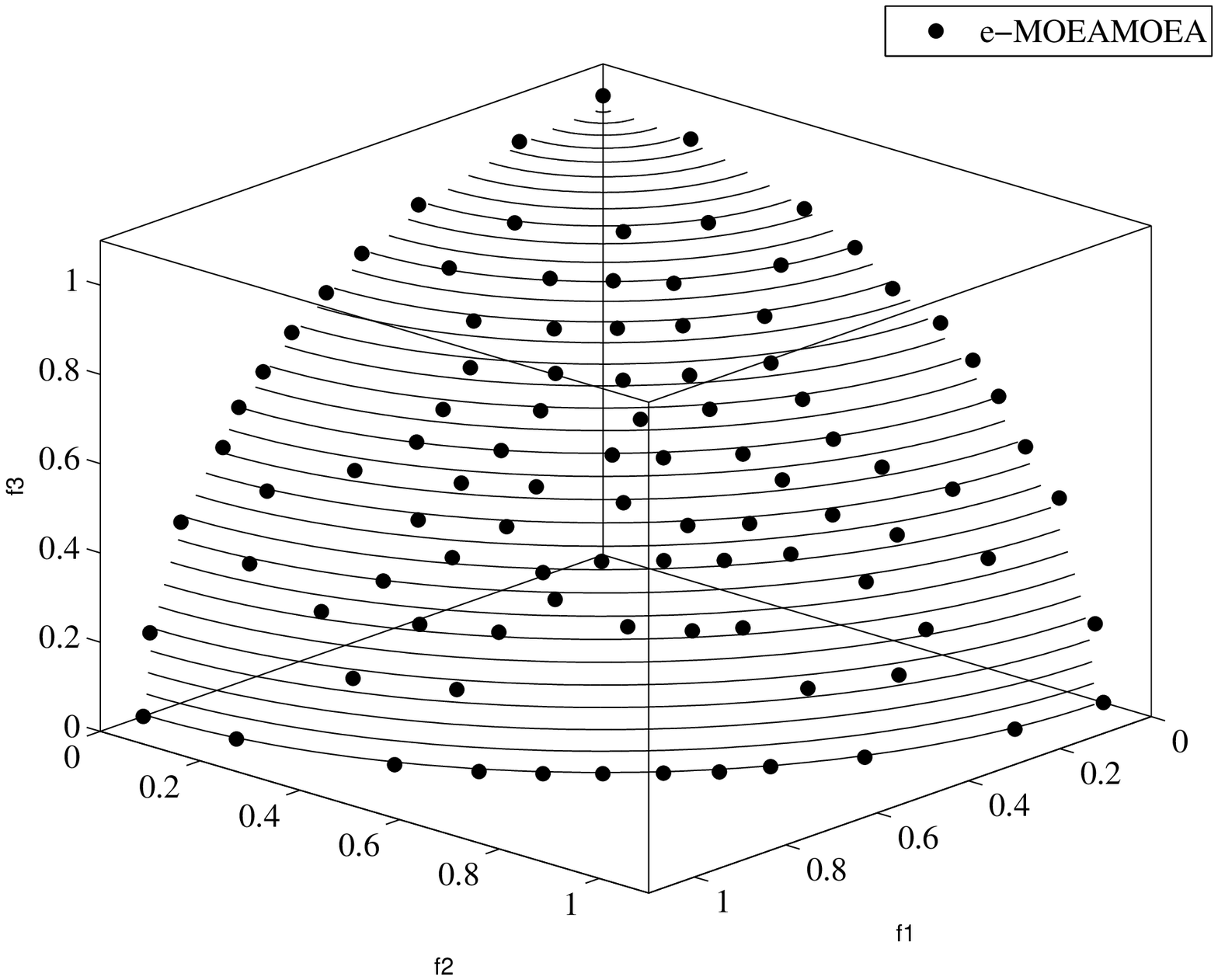}}
\subfigure[\small cone$\eps$-MOEA.]{\includegraphics[width=4.4cm]{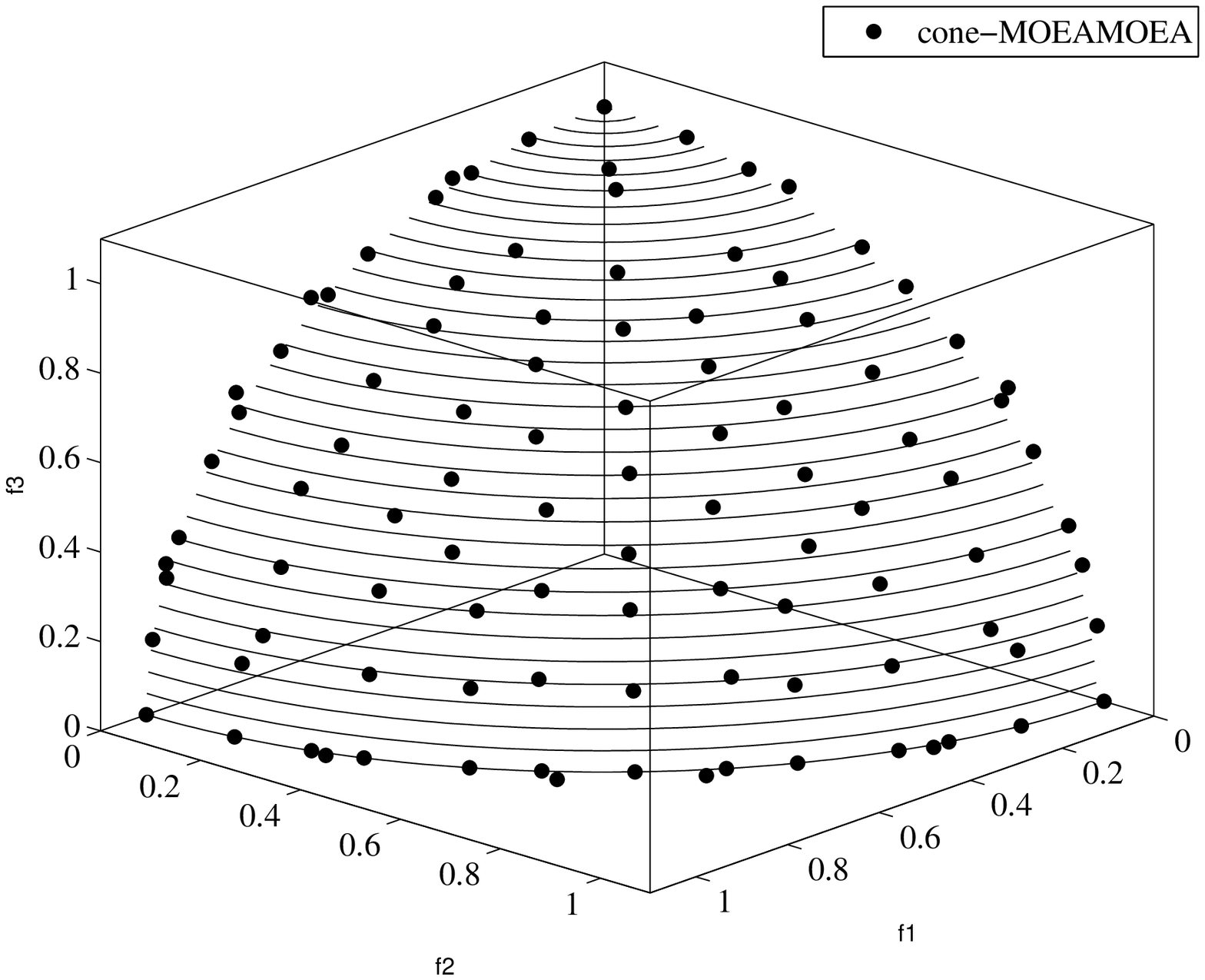}}
\caption{Efficient solutions generated for problem DTLZ4, considering the estimated $\pmb{\eps}$ values provided in Table \ref{tab:eps_values}. The fronts presented are the outcome of a typical run.}
\label{fig:viewDTLZ4}
\end{figure*}

The DTLZ4 12-variable test problem introduces a nonuniform density of solutions on the three-objective Pareto-optimal set, and so it investigates the ability of the MOEA to maintain a good distribution of solutions on the corresponding front. The global optimal front is the surface represented by $\sum_{i=1}^{m} f_{i}^{2} = 1$, in which $f_{i} \in [0,1]$ for all $i = \left\{1,2,3\right\}$ \citep{proc.Deb2005}. As this problem has a greater density of solutions near the $f_{1}$--$f_{2}$ and $f_{1}$--$f_{3}$ planes, some runs of all the algorithms achieved solutions only in these regions. In this problem, the NSGA-II* was the best in terms of convergence, followed by NSGA-II, SPEA2, C-NSGA-II, and $\eps$-MOEA, which presented an advantage over the cone$\eps$-method. Concerning the spread of solutions, the best methods were the SPEA2 and C-NSGA-II, followed by both $\eps$-approaches. Although there is a difference in $\Delta$ in favor of the $\eps$-MOEA when compared with the cone$\eps$-MOEA, it can be seen from Fig. \ref{fig:viewDTLZ4} that the border regions of the Pareto front were better mapped by the cone$\eps$-method. Note again that, even though these fronts are the outcome of a single run, both the $\eps$ and the cone$\eps$-concepts yielded similar distributions of solutions in all runs. Regarding the $HV$ and $CS$ metrics, only small differences were observed between the algorithms.

The three-objective 12-variable DTLZ5 problem has a Pareto-optimal front satisfying $\sum_{i=1}^{m} f_{i}^{2} = 1$, in which $f_{i} \in [0,1]$ for all $i = \left\{1,2,3\right\}$. This problem tests the ability of the MOEA to find a lower-dimensional Pareto curve while dealing with a higher-dimensional objective space \citep{proc.Deb2005}. In this problem, the NSGA-II* achieved the best convergence, closely followed by the NSGA-II and SPEA2. For the diversity indicator, the clustering-based methods (SPEA2 and C-NSGA-II) provided the best overall $\Delta$ values. The cone$\eps$-MOEA performed worst in these metrics. The DTLZ6 problem is basically a harder version of the DTLZ5 \citep{proc.Deb2005}. In this case, the cone$\eps$-MOEA presented the best and the second best performances in finding well-converged and well-distributed solutions, respectively. Once more, the $HV$ and $CS$ metrics presented very small differences between the algorithms tested.

\begin{figure*}[!htb]
\centering
\psfrag{f1}[][]{$f_{1}$}
\psfrag{f2}[][]{$f_{2}$}
\psfrag{f3}[][]{$f_{3}$}
\psfrag{Fronteira Pareto Global}[][]{\tiny Pareto front}
\psfrag{cone-MOEAMOEA}[][]{\tiny c$\eps$-MOEA}
\psfrag{e-MOEAMOEA}[][]{\tiny $\eps$-MOEA}
\psfrag{NSGANSGA-II}[][]{\tiny NSGA-II}
\psfrag{NSGANSGA-II-s}[][]{\tiny NSGA-II*}
\psfrag{CNSGANSGA-II}[][]{\tiny C-NSGA-II}
\psfrag{SPEA2-SPEA2}[][]{\tiny SPEA2}
\subfigure[\small NSGA-II.]{\includegraphics[width=4.4cm]{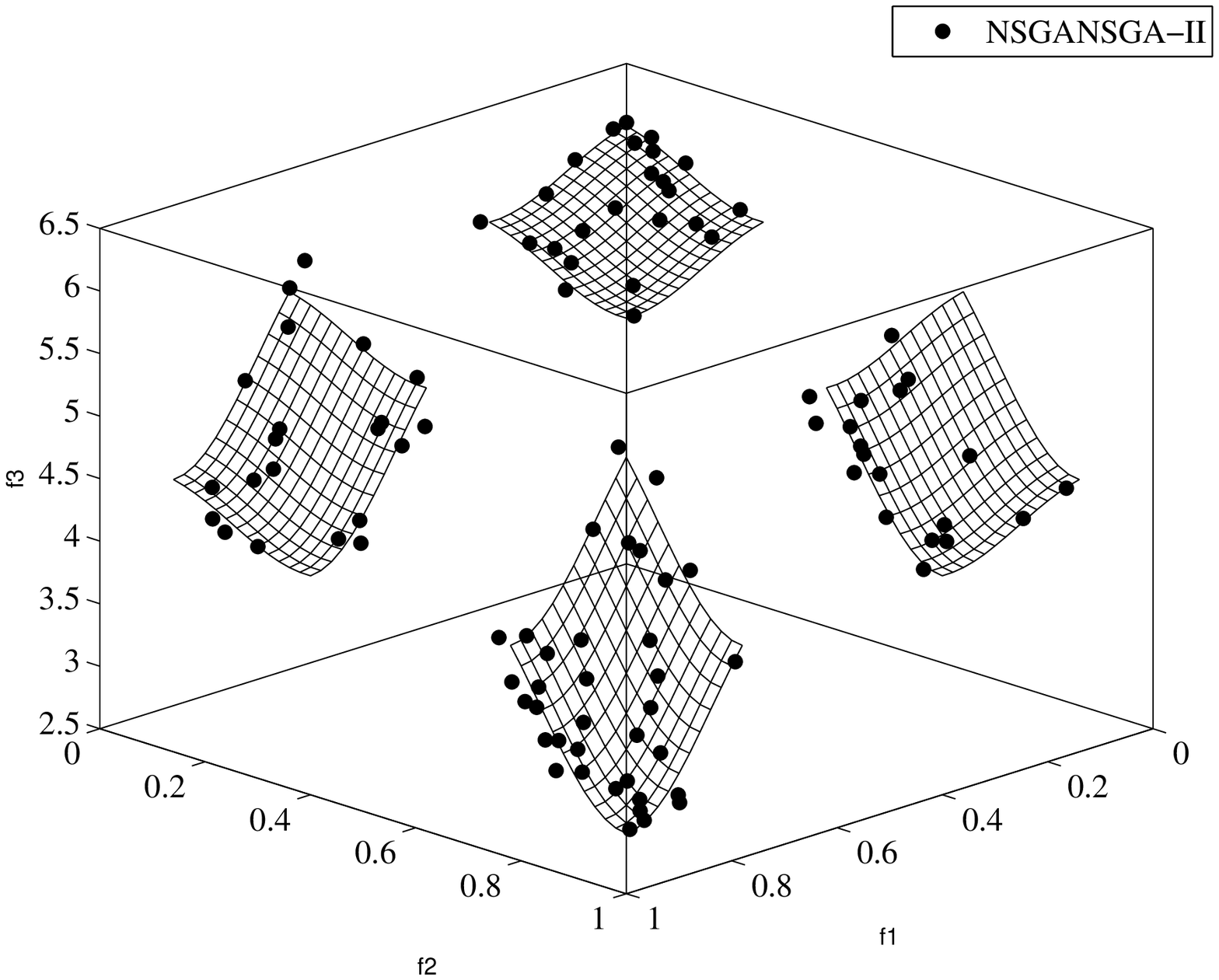}}
\subfigure[\small NSGA-II*.]{\includegraphics[width=4.4cm]{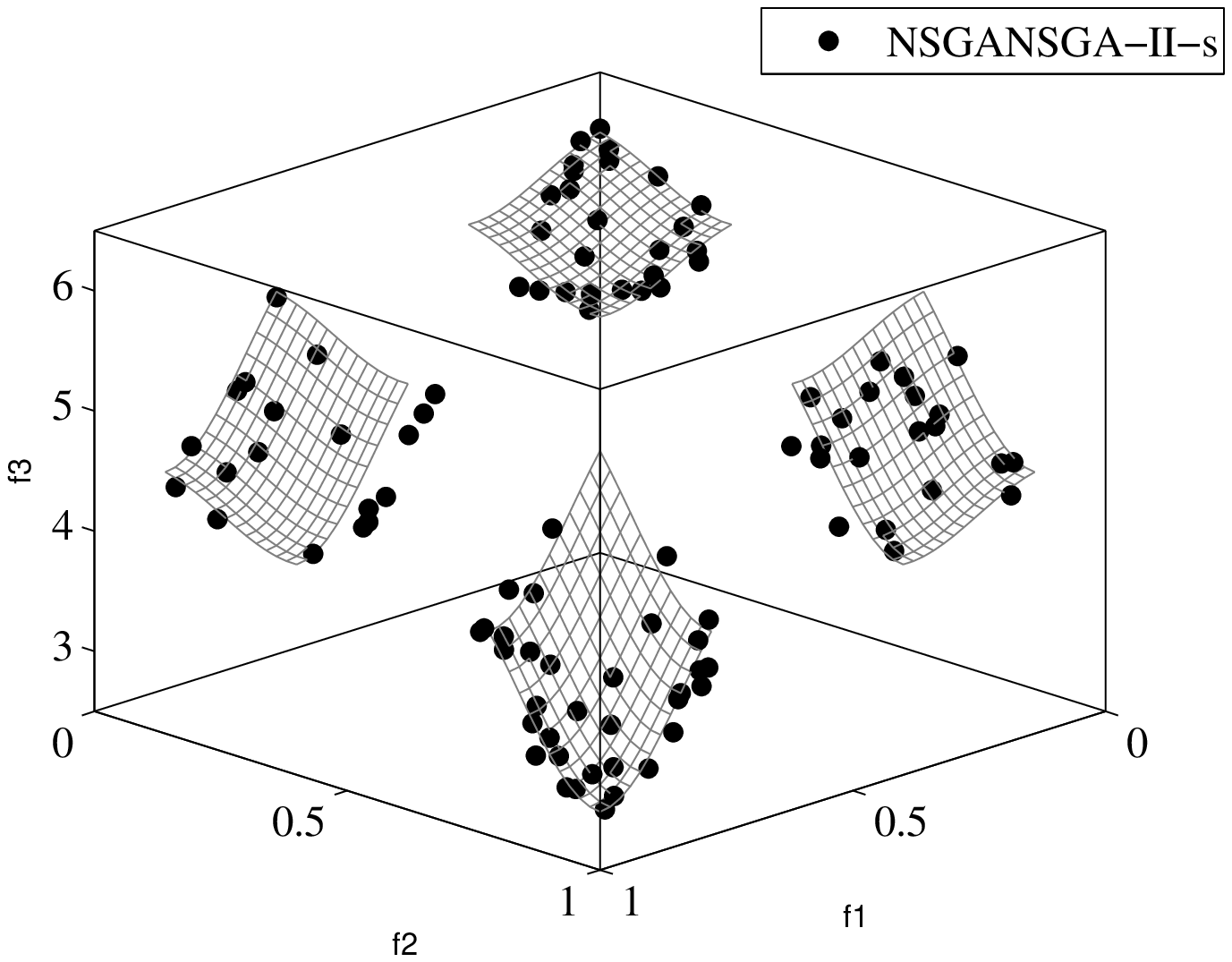}}
\subfigure[\small C-NSGA-II.]{\includegraphics[width=4.4cm]{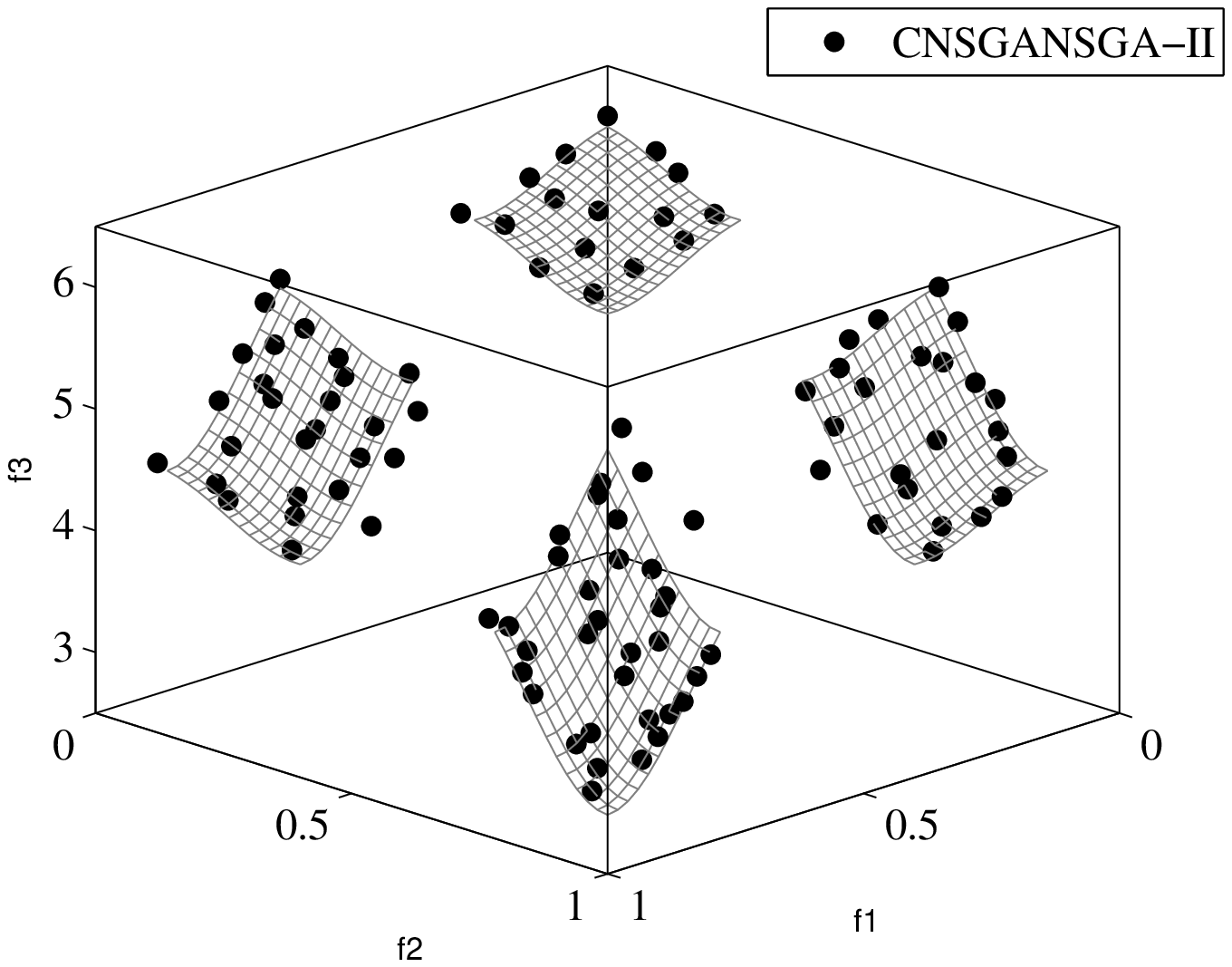}}
\subfigure[\small SPEA2.]{\includegraphics[width=4.4cm]{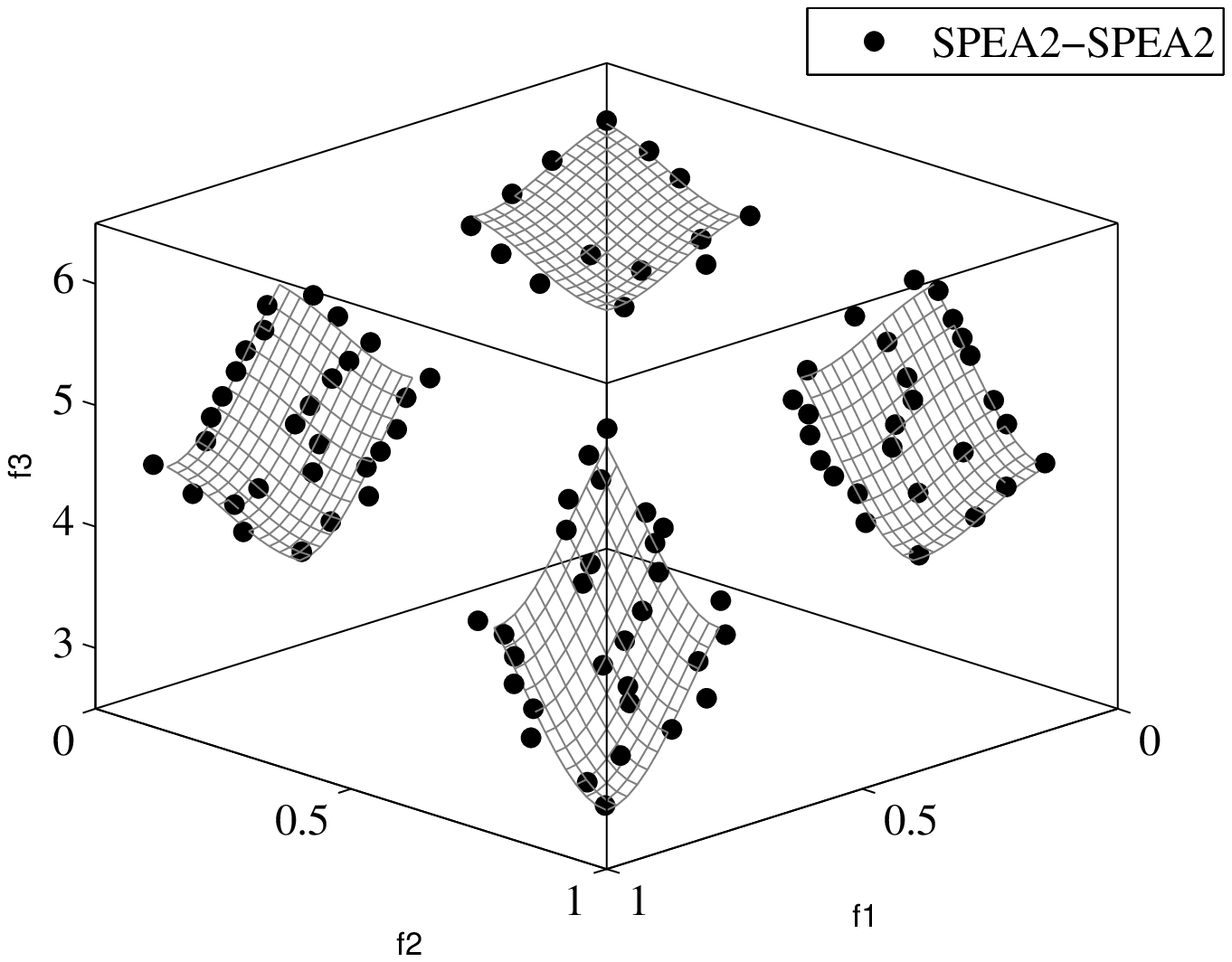}}
\subfigure[\small $\eps$-MOEA.]{\includegraphics[width=4.4cm]{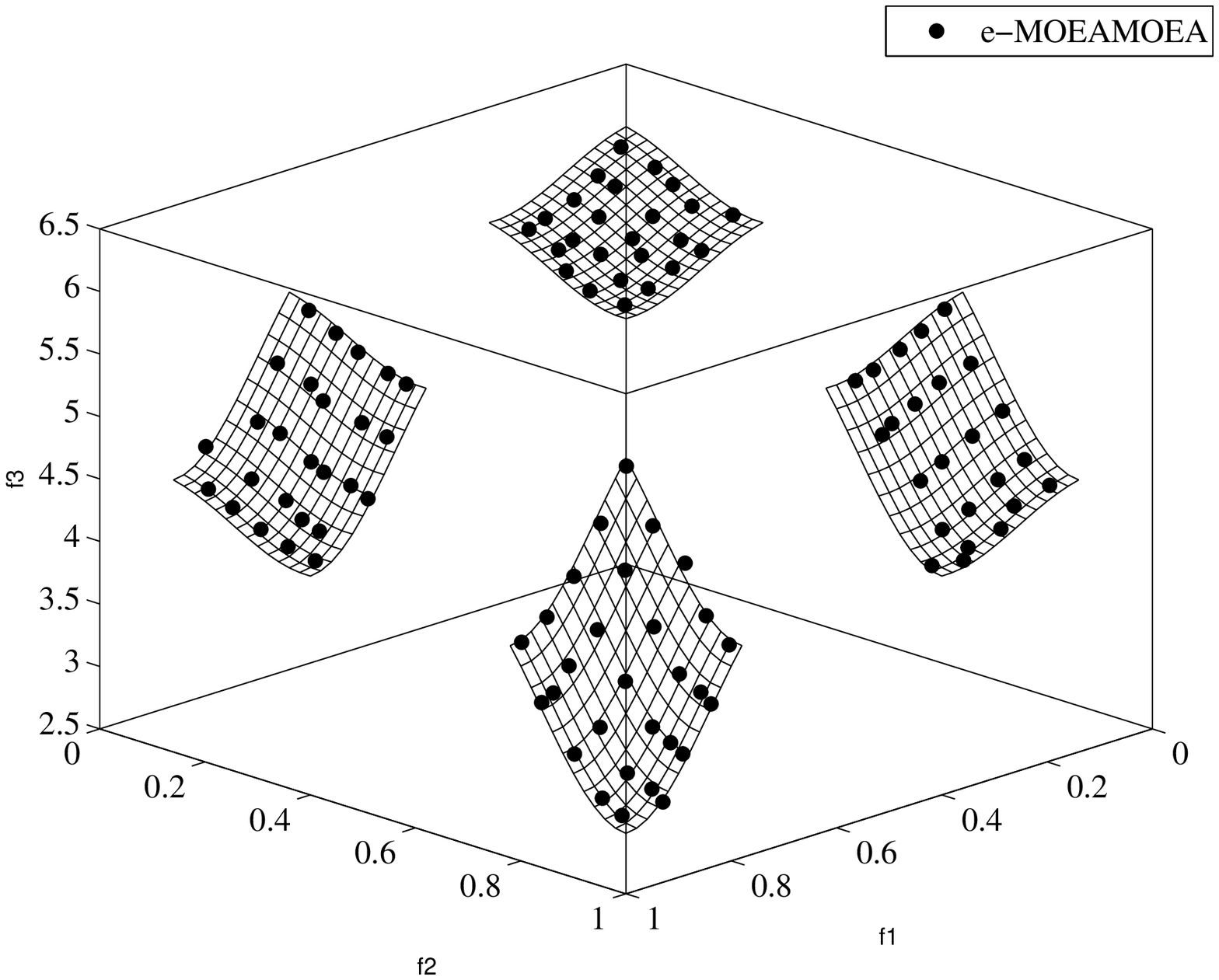}}
\subfigure[\small cone$\eps$-MOEA.]{\includegraphics[width=4.4cm]{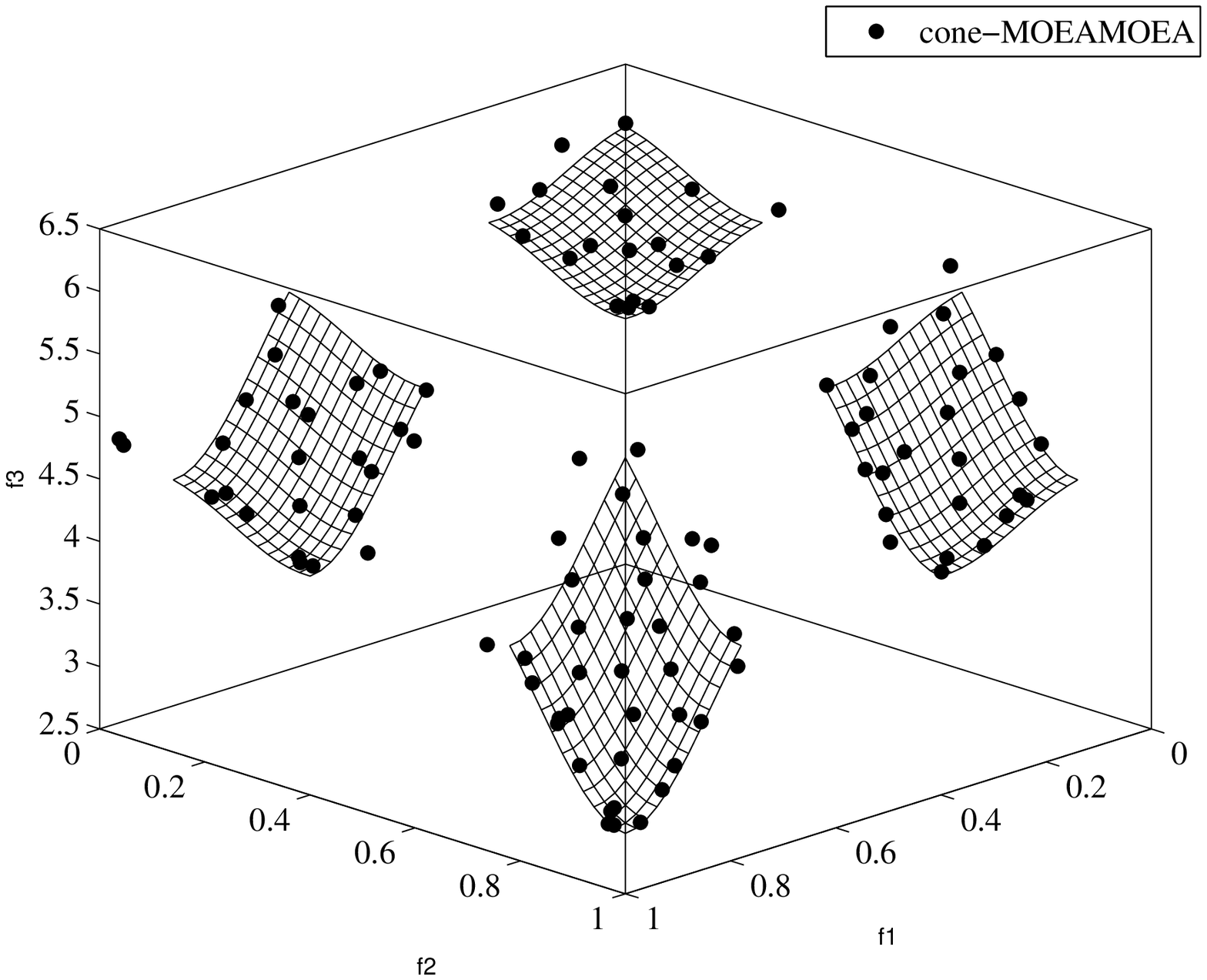}\label{fig:viewDTLZ7a}}
\caption{Efficient solutions generated for problem DTLZ7, considering the estimated $\pmb{\eps}$ values provided in Table \ref{tab:eps_values}. The fronts presented are the outcome of a typical run.}
\label{fig:viewDTLZ7}
\end{figure*}

The 22-variable DTLZ7 problem has a disconnected set of Pareto-optimal areas in the search space, and tests the ability of the algorithm to maintain subpopulations in different Pareto front regions \citep{proc.Deb2005}. The $\eps$-MOEA has outperformed the other methods in the convergence indicator, and the SPEA2 achieved the best $\Delta$ value, closely followed by $\eps$-MOEA and C-NSGA-II. The cone$\eps$-method ranked fourth in this measure. Again, the $HV$ and $CS$ metrics presented very small differences between the algorithms, with positive gains in favor of the SPEA2 in both cases. Figure \ref{fig:viewDTLZ7} shows the outcome of a typical run of each algorithm considered in this study. Since the influence on the ordering of points performed by the cone of dominance was limited to a local neighborhood in the objective space when using $\kappa = 0.5$, the cone$\eps$-method was not able to estimate a well-spread approximation of the global Pareto front (see Fig. \ref{fig:viewDTLZ7a}), which may also have adversely affected the convergence of the approach. Due to this feature, both the cone$\eps$-MOEA and the NSGA-II presented a very similar distribution of solutions on this test. Note, however, that the performance of the cone$\eps$-approach should be improved by considering a smaller $\kappa$ value, and a more desirable quality for the convergence and diversity of the solutions could be found.

\begin{figure*}[!htb]
\centering
\psfrag{f1}[][]{$f_{1}$}
\psfrag{f2}[][]{$f_{2}$}
\psfrag{f3}[][]{$f_{3}$}
\psfrag{Fronteira Pareto Global}[][]{\tiny Pareto front}
\psfrag{cone-MOEAMOEA}[][]{\tiny c$\eps$-MOEA}
\psfrag{e-MOEAMOEA}[][]{\tiny $\eps$-MOEA}
\psfrag{NSGANSGA-II}[][]{\tiny NSGA-II}
\psfrag{NSGANSGA-II-s}[][]{\tiny NSGA-II*}
\psfrag{CNSGANSGA-II}[][]{\tiny C-NSGA-II}
\psfrag{SPEA2-SPEA2}[][]{\tiny SPEA2}
\subfigure[\small NSGA-II.]{\includegraphics[width=4.4cm]{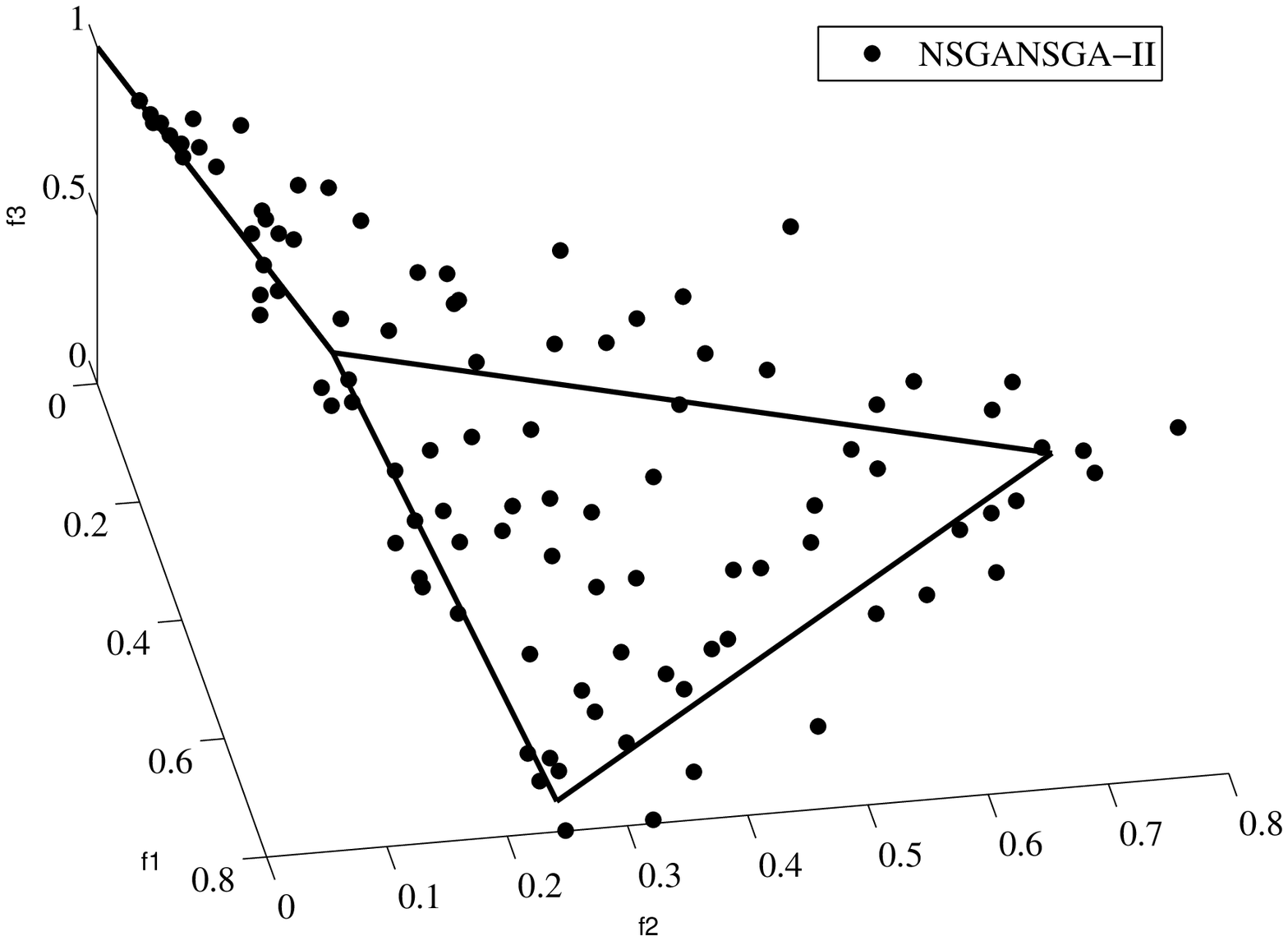}\label{fig:viewDTLZ8a}}
\subfigure[\small NSGA-II*.]{\includegraphics[width=4.4cm]{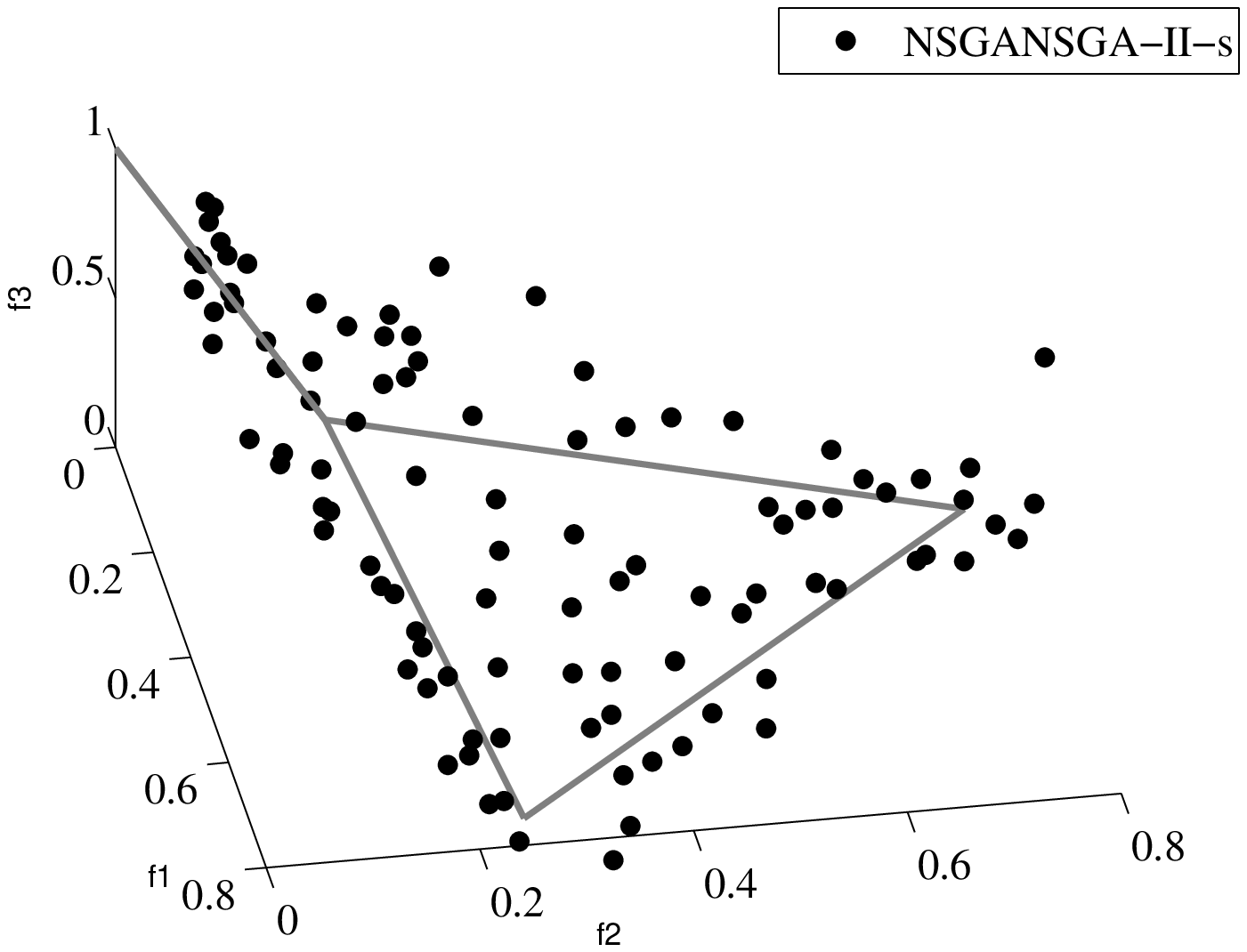}\label{fig:viewDTLZ8ab}}
\subfigure[\small C-NSGA-II.]{\includegraphics[width=4.4cm]{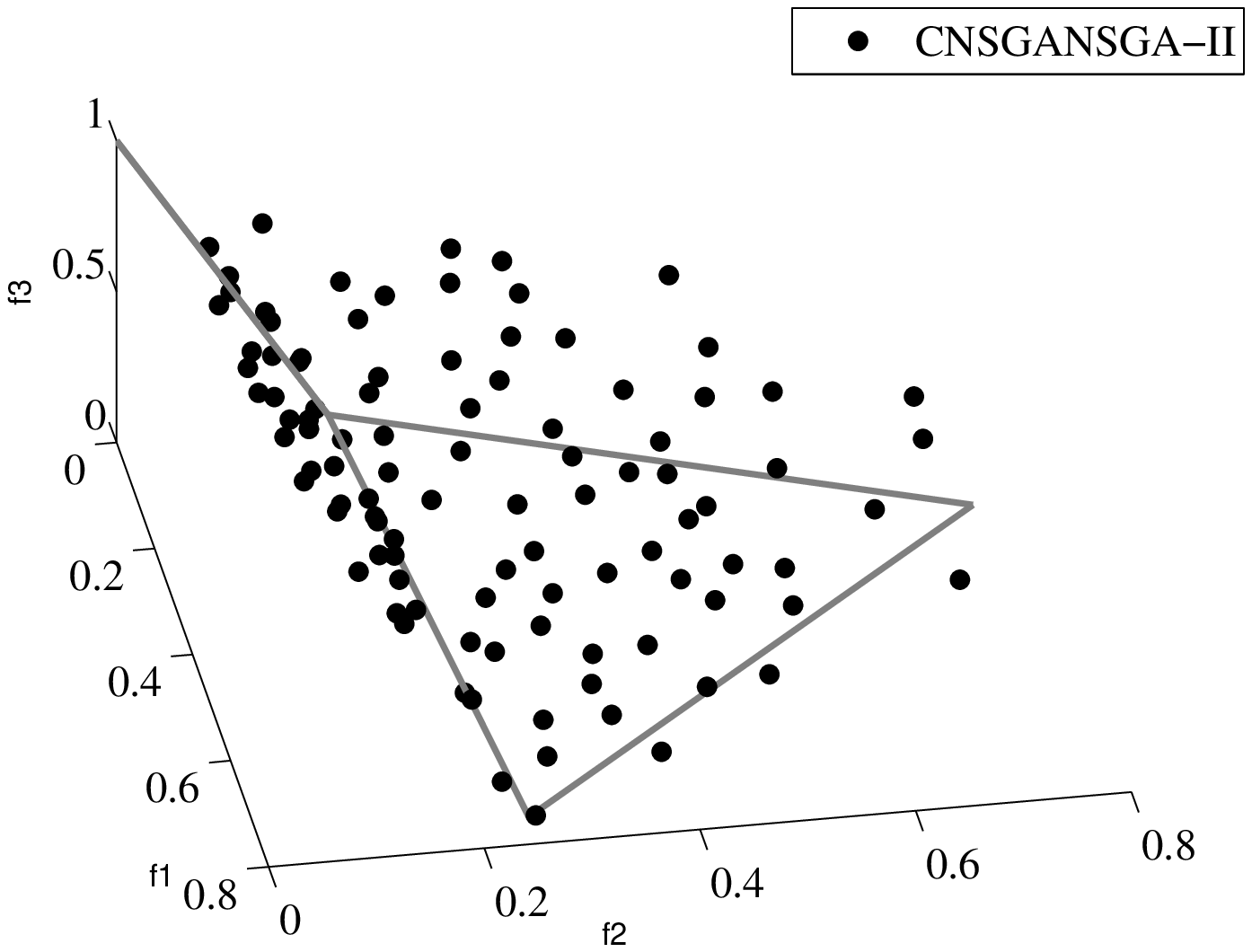}}
\subfigure[\small SPEA2.]{\includegraphics[width=4.4cm]{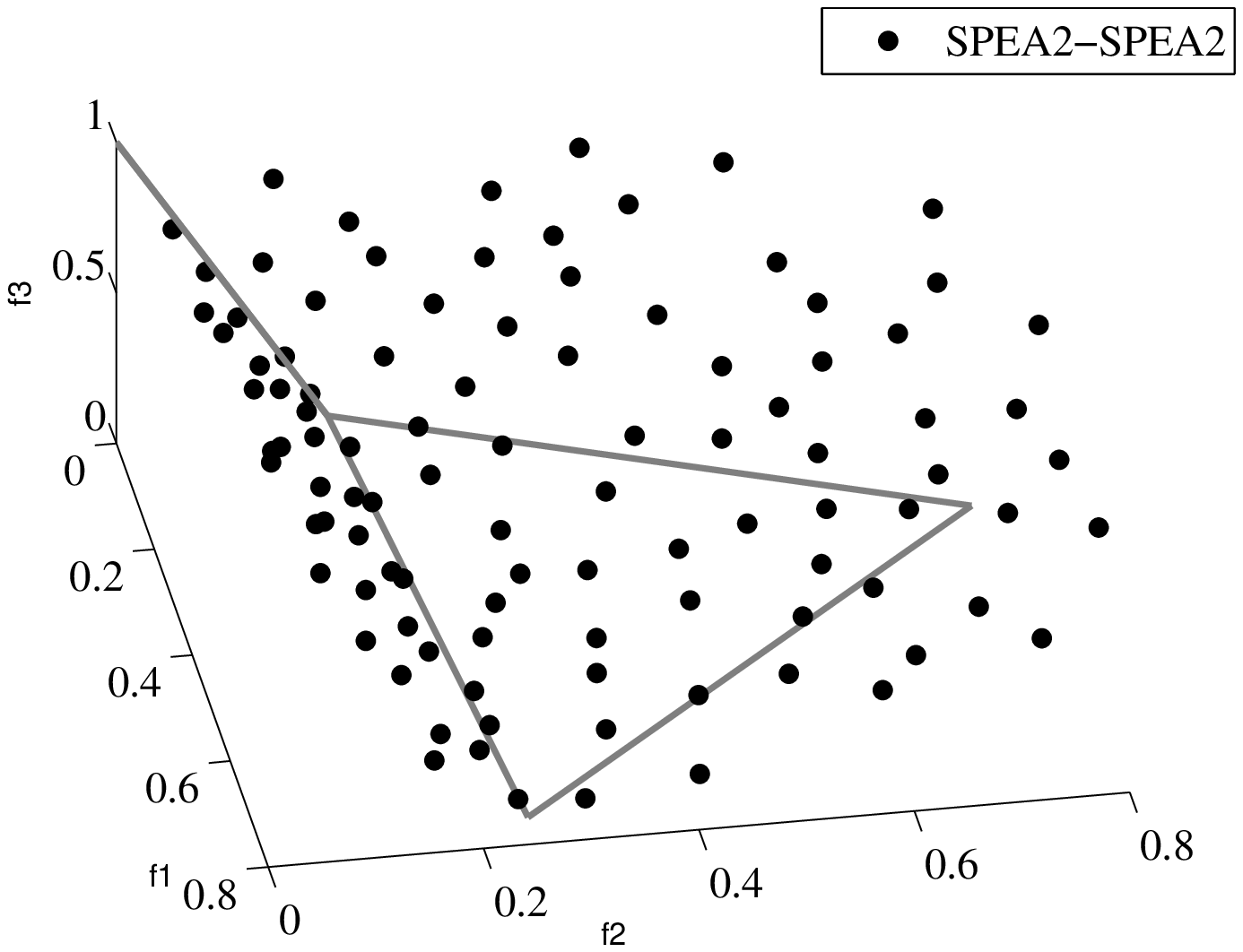}}
\subfigure[\small $\eps$-MOEA.]{\includegraphics[width=4.4cm]{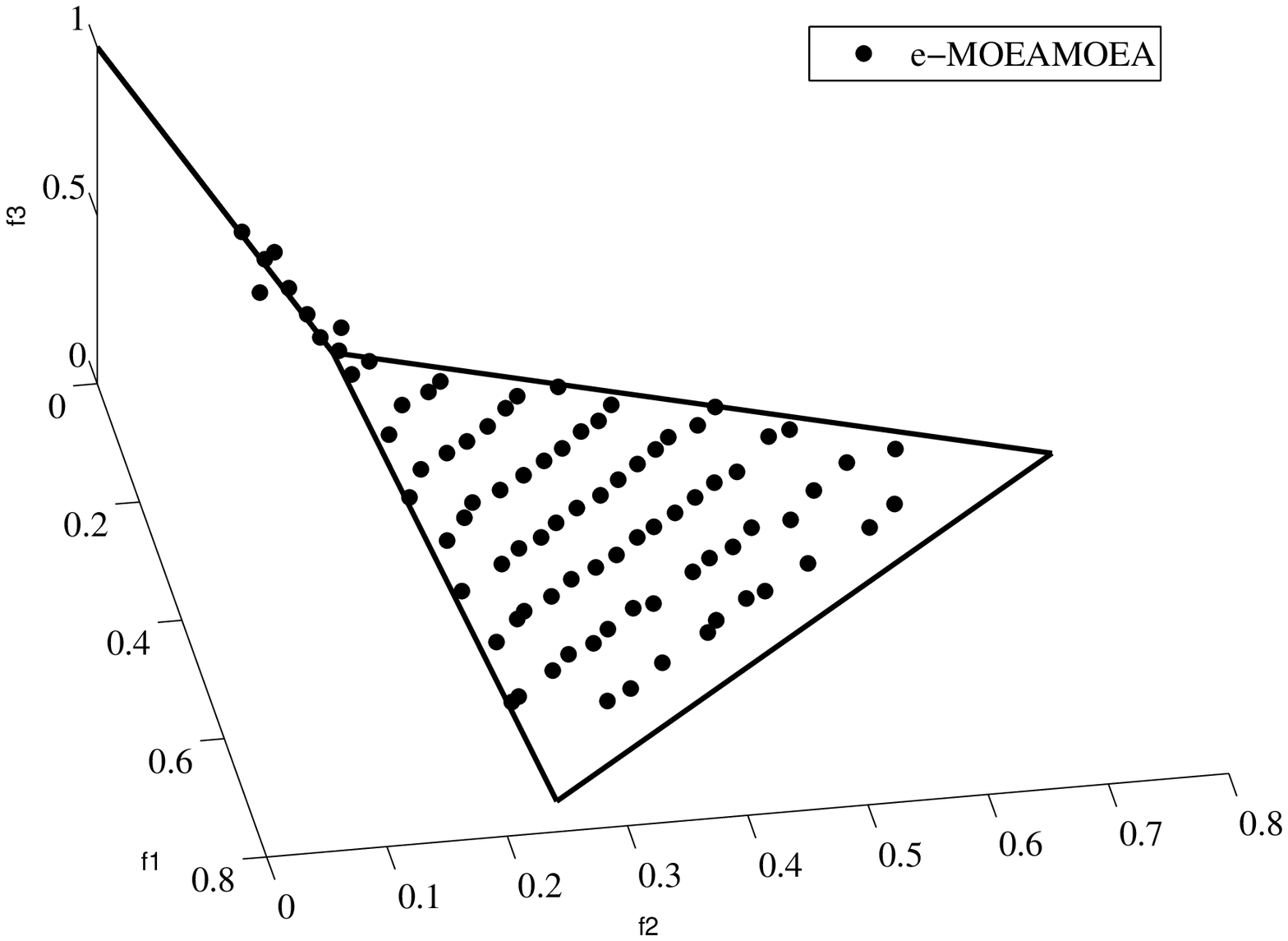}\label{fig:viewDTLZ8b}}
\subfigure[\small cone$\eps$-MOEA.]{\includegraphics[width=4.4cm]{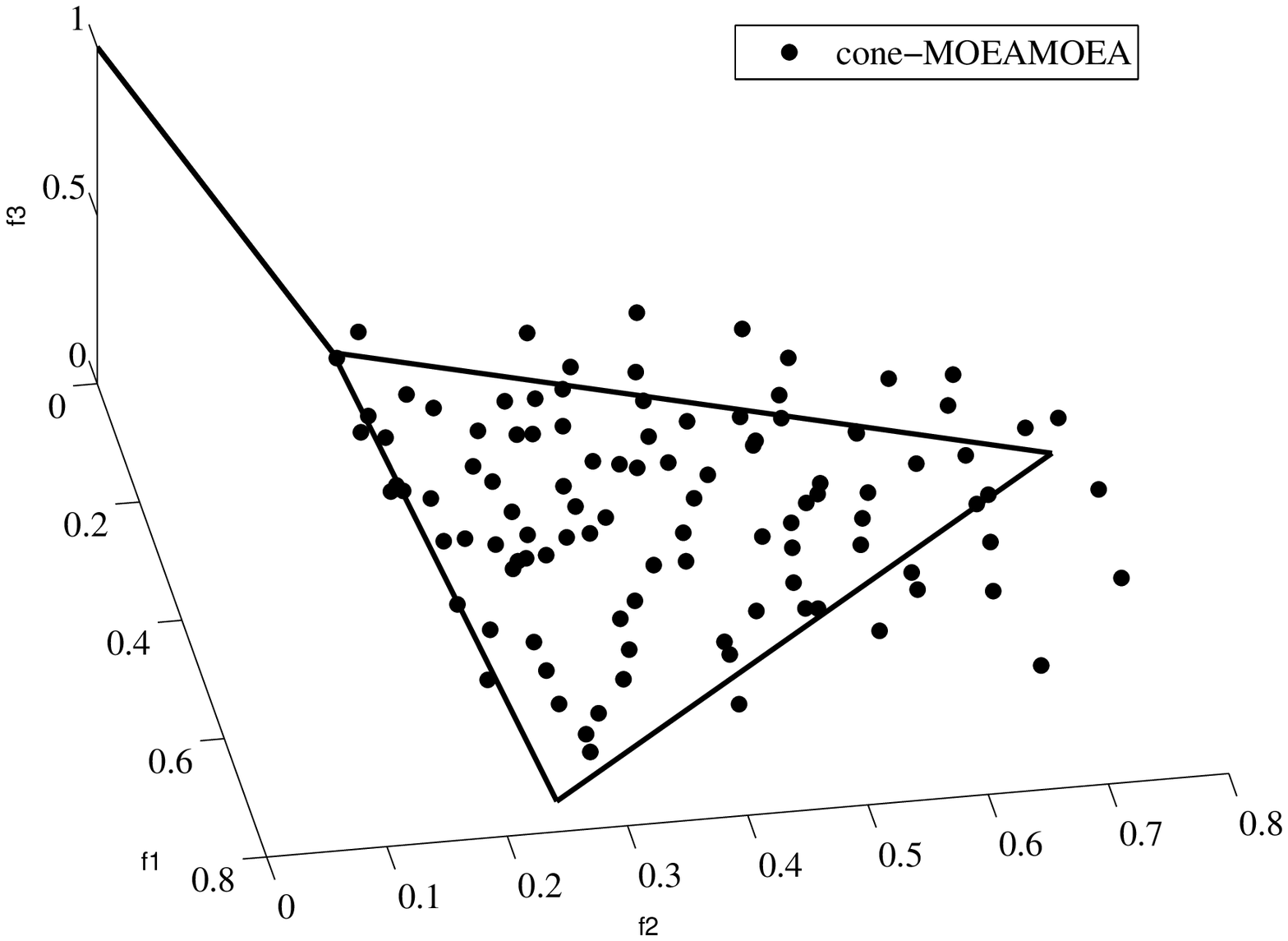}\label{fig:viewDTLZ8c}}
\caption{Efficient solutions generated for problem DTLZ8, considering the estimated $\pmb{\eps}$ values provided in Table \ref{tab:eps_values}. The fronts presented are the outcome of a typical run.}
\label{fig:viewDTLZ8}
\end{figure*}

In the three-objective 30-variable DTLZ8 problem, the overall Pareto-optimal front is a combination of a straight line and a hyperplane. This problem involves $n_{g} = 3$ inequality constraints: the straight line is the intersection of the first $n_{g}-1$ constraints, and the hyperplane is represented by the last one. MOEAs may find difficulty in achieving solutions in both the Pareto regions and also in maintaining a good distribution of solutions on the hyperplane \citep{proc.Deb2005}. In this study, the constraints were handled using a simple penalty method ($f_{i} + 1000 \sum_{j=1}^{n_{g}} \max\left(0,g_{j}\right)$) for all the algorithms. As discussed elsewhere \citep{proc.Deb2003,journal.Deb2005a,proc.Deb2005}, the domination-based MOEAs suffer, in this particular problem, from what is known as the ``redundancy problem''. In general terms, even though some solutions are found on the global Pareto front, there exist many other weakly nondominated solutions in the final archive. In Fig. \ref{fig:viewDTLZ8a}, these ``redundant solutions'' are those that are on the adjoining surfaces of the Pareto-optimal front, and their presence in the final nondominated set is very difficult to eradicate in real-parameter MOEAs \citep{inproc.Ikeda2001}. Due to this feature, the obtained set of solutions may wrongly find a higher-dimensional surface than the Pareto-optimal front, although the global Pareto front may be of smaller dimension. As a direct consequence, the S-metric ($HV$) is not appropriate for this problem, as the redundant solutions may affect the hypervolume indicator, and was not considered in this case. Figure \ref{fig:viewDTLZ8b} shows that many of these redundant solutions get $\eps$-dominated by the Pareto-optimal solutions found. In the case of the cone$\eps$-approach, as the cone of dominance presented a local effect in the objective space when using $\kappa = 0.5$, and only few redundant solutions were removed from the archive (see Fig. \ref{fig:viewDTLZ8c}). Again, it is important to observe that a global effect can be performed by choosing a smaller $\kappa$ value (e.g, a $\kappa\rightarrow 0$ would achieve the same results as the $\eps$-dominance). From Table \ref{tab:problemwise2}, it can be seen that the clustering-based methods achieved the best $\Delta$ values. Although no significant difference was found between the $\eps$-methods in finding well-distributed solutions, they have outperformed the NSGA-II and NSGA-II* in this indicator. The $\eps$-MOEA found the best result for the $\gamma$ metric, followed by the cone$\eps$-algorithm. The overall best $CS$ measure was obtained by the cone$\eps$-MOEA.

\begin{figure*}[!ht]
\centering
\psfrag{f1}[][]{$f_{1}$}
\psfrag{f2}[][]{$f_{2}$}
\psfrag{f3}[][]{$f_{3}$}
\psfrag{Fronteira Pareto Global}[][]{\tiny Pareto front}
\psfrag{cone-MOEAMOEA}[][]{\tiny c$\eps$-MOEA}
\psfrag{e-MOEAMOEA}[][]{\tiny $\eps$-MOEA}
\psfrag{NSGANSGA-II}[][]{\tiny NSGA-II}
\psfrag{NSGANSGA-II-s}[][]{\tiny NSGA-II*}
\psfrag{CNSGANSGA-II}[][]{\tiny C-NSGA-II}
\psfrag{SPEA2-SPEA2}[][]{\tiny SPEA2}
\subfigure[\small NSGA-II.]{\includegraphics[width=4.2cm]{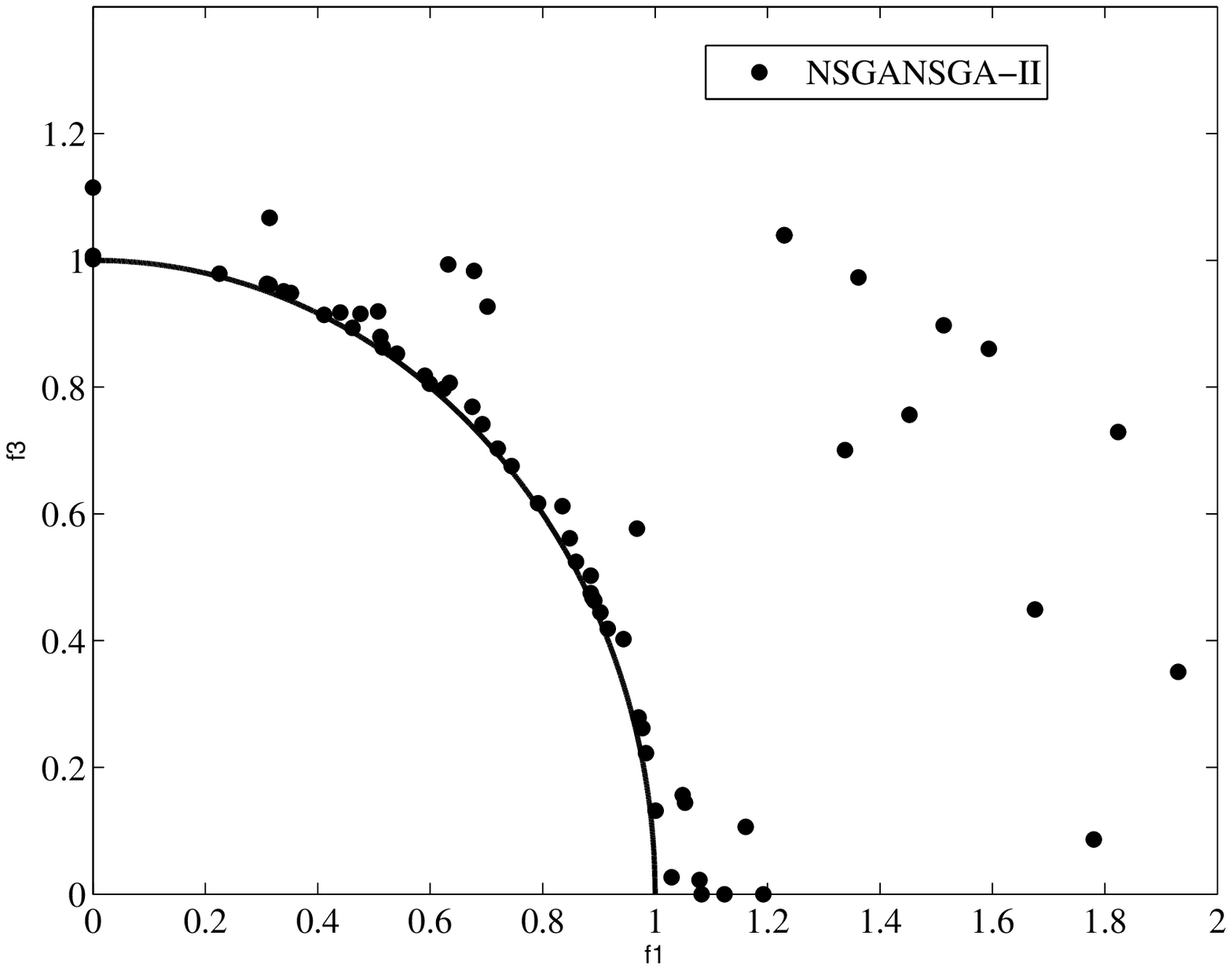}}
\subfigure[\small NSGA-II*.]{\includegraphics[width=4.5cm]{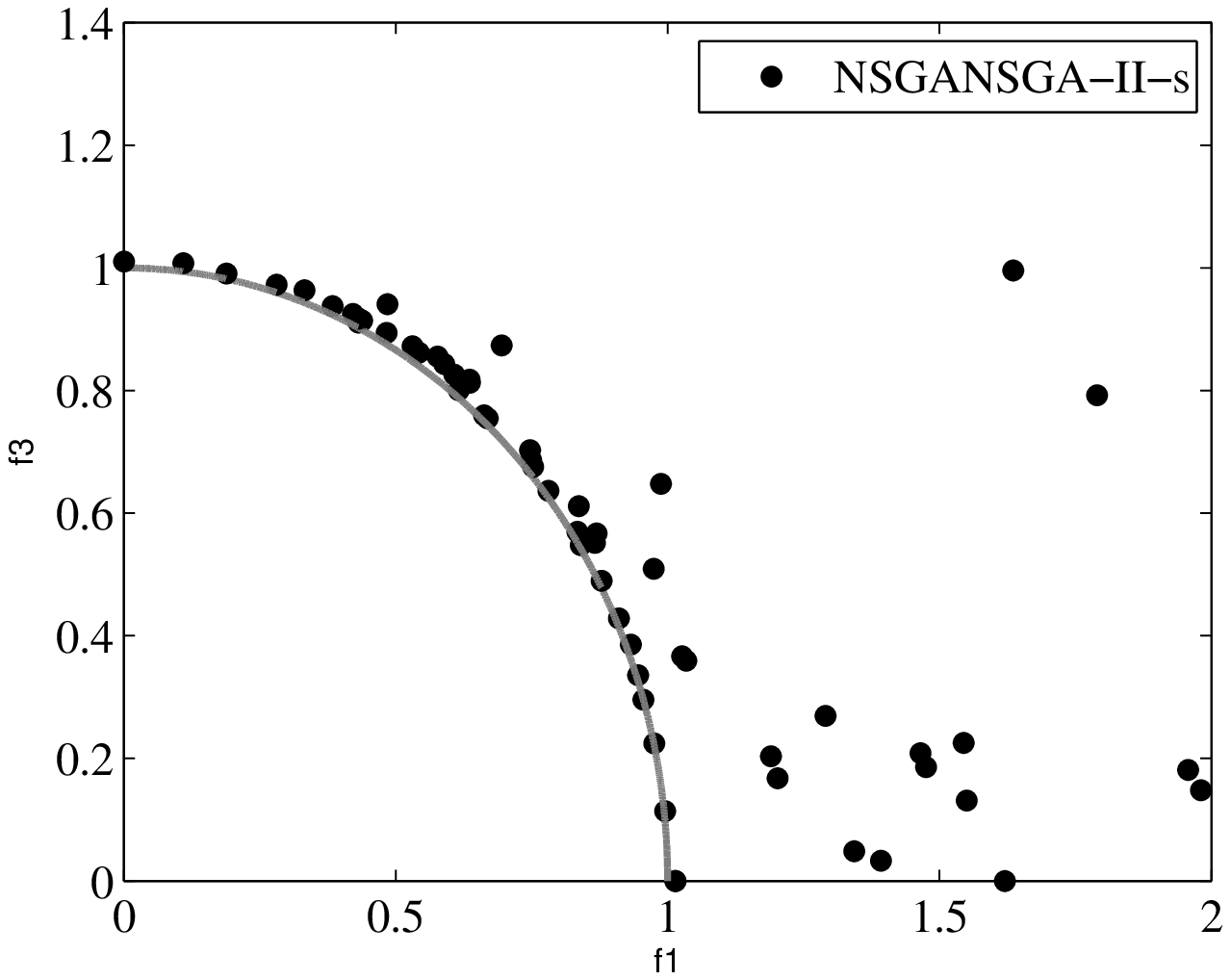}}
\subfigure[\small C-NSGA-II.]{\includegraphics[width=4.5cm]{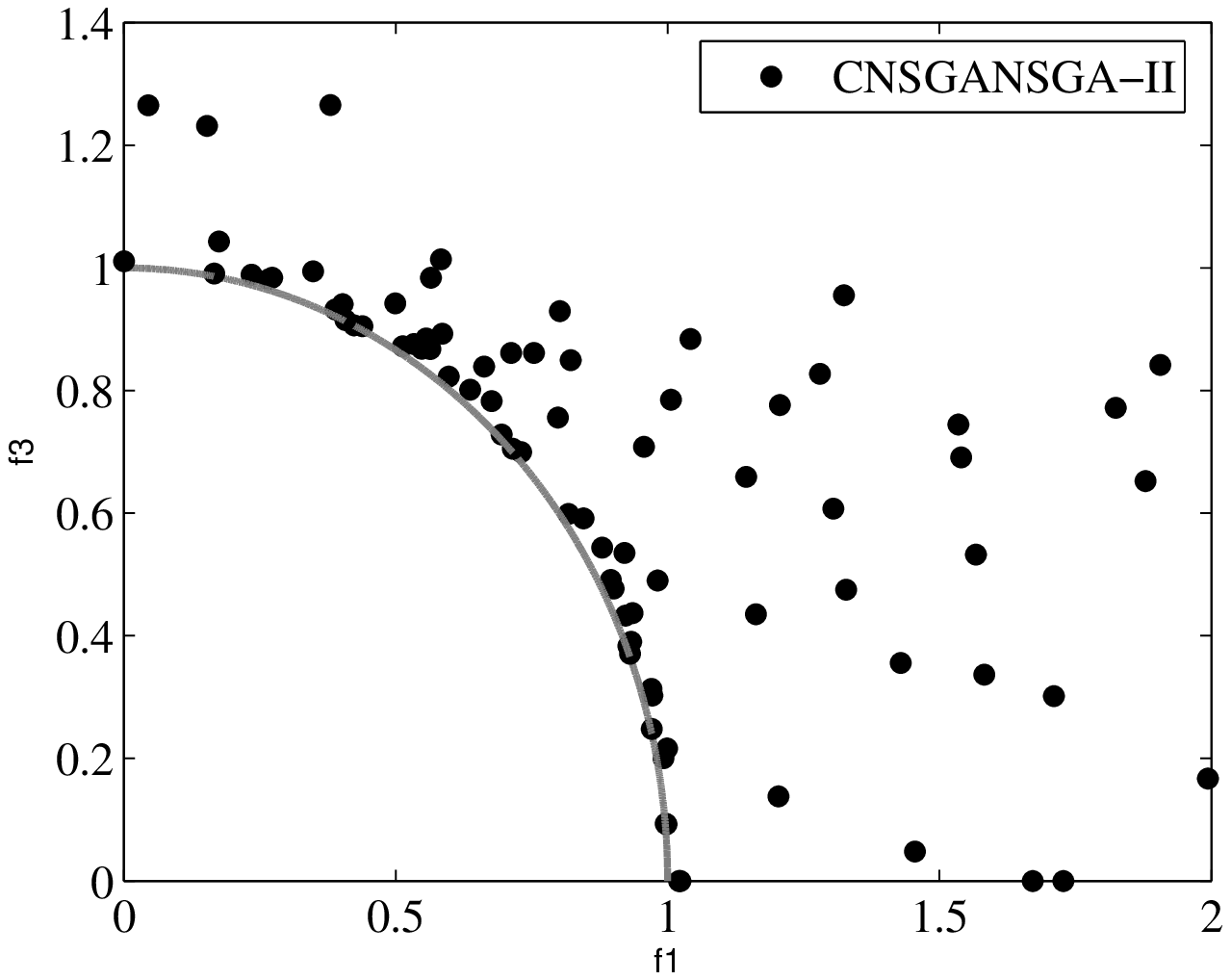}}
\subfigure[\small SPEA2.]{\includegraphics[width=4.6cm]{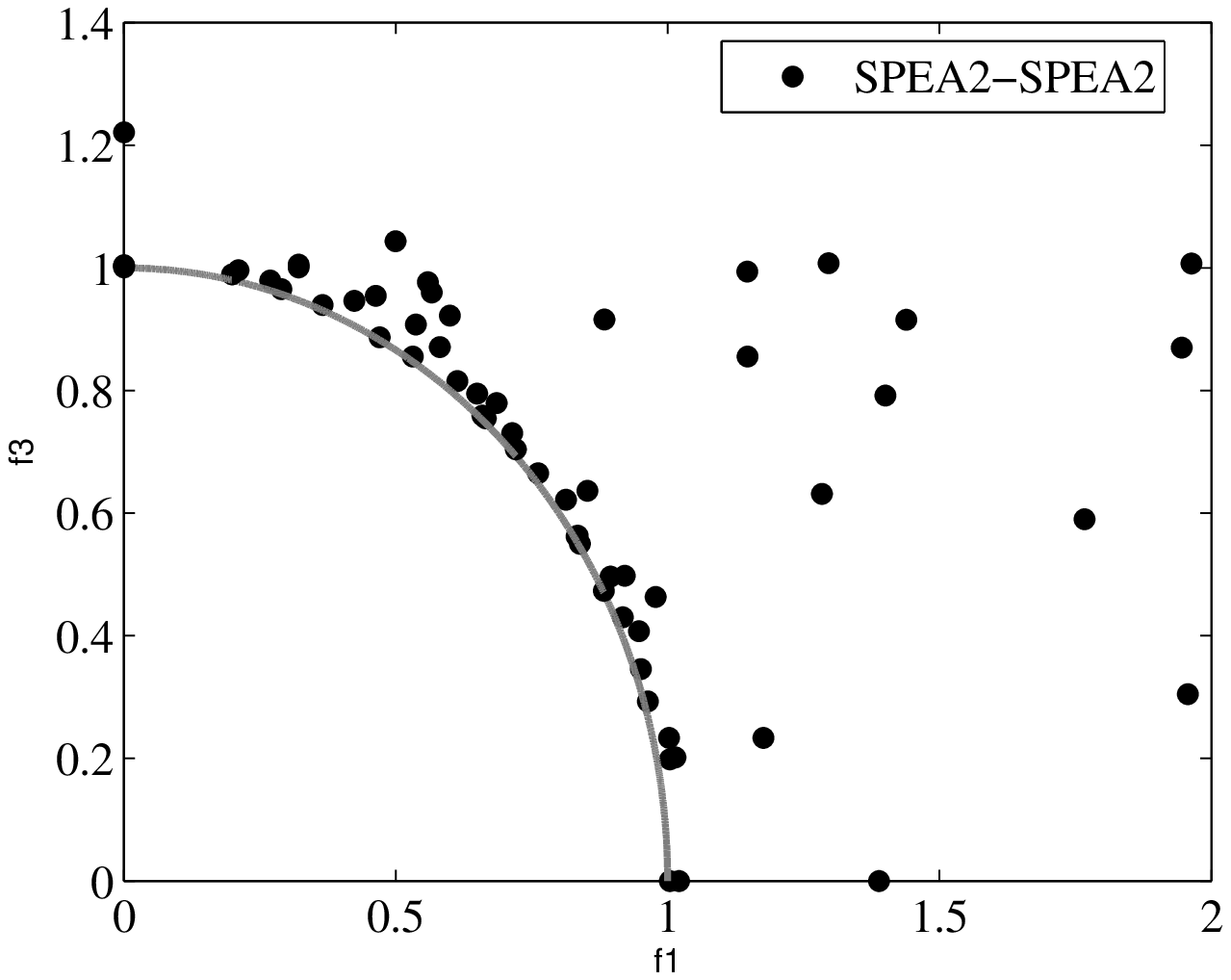}}
\subfigure[\small $\eps$-MOEA.]{\includegraphics[width=4.3cm]{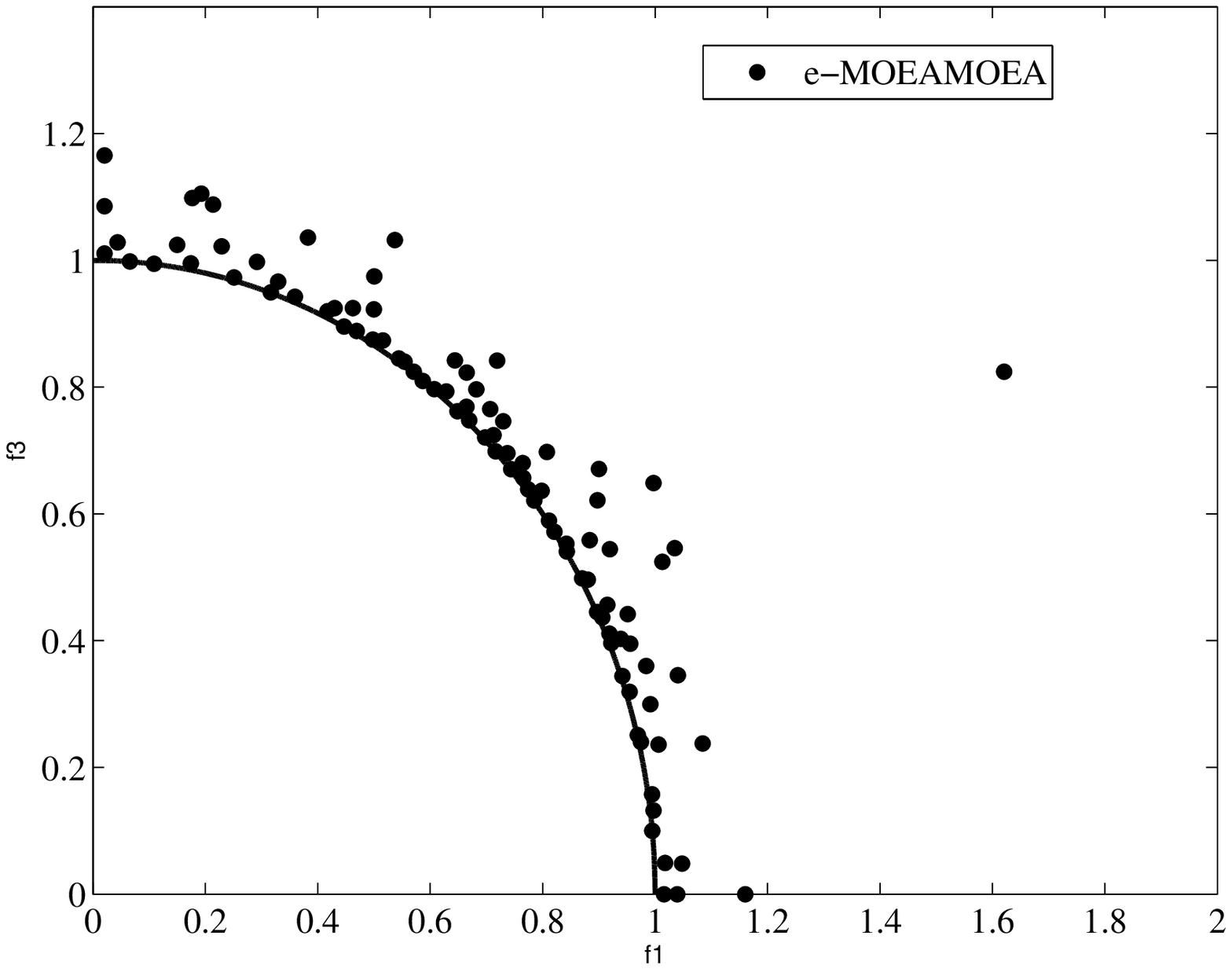}}
\subfigure[\small cone$\eps$-MOEA.]{\includegraphics[width=4.3cm]{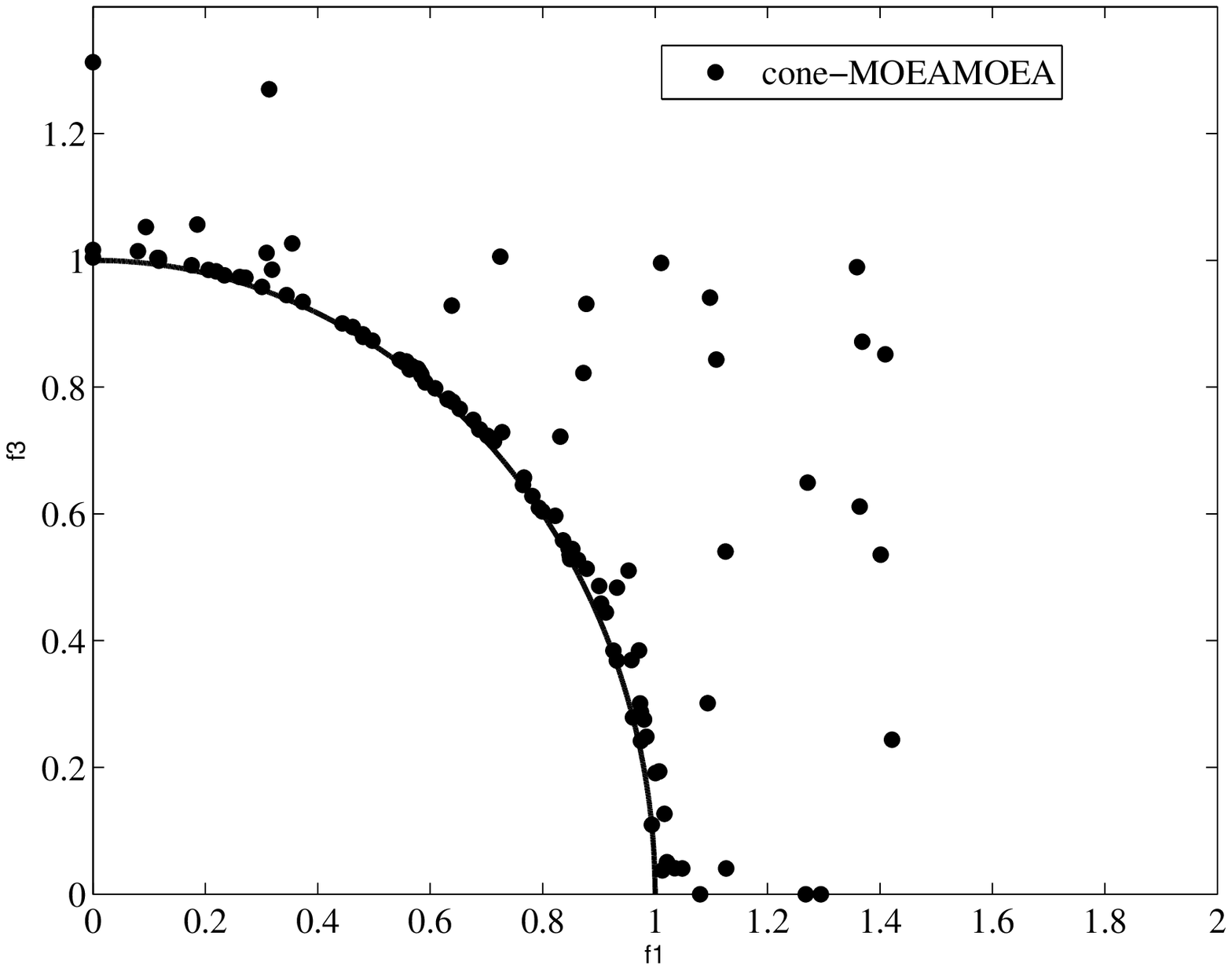}\label{fig:viewDTLZ9c}}
\caption{Efficient solutions generated for problem DTLZ9, considering the estimated $\pmb{\eps}$ values provided in Table \ref{tab:eps_values}. The fronts presented are the outcome of a typical run.}
\label{fig:viewDTLZ9}
\end{figure*}

The 30-variable DTLZ9 problem is also created using the constraint surface approach employed by the DTLZ8 test. The Pareto-optimal front is a curve in a three-dimensional objective space, and the density of solutions gets thinner towards the Pareto-optimal region. This curve lies on the intersection of all the inequality constraints. As in the previous problem, the DTLZ9 test investigates the ability of the MOEA to deal with redundant solutions \citep{proc.Deb2005}. Again, the constraints were handled using a penalty method ($f_{i} + 1000 \sum_{j=1}^{n_{g}} \max\left(0,g_{j}\right)$) for all the algorithms. A two-dimensional plot of the Pareto-optimal curve on the $f_{1}$--$f_{3}$ plane allows an easier visualization of the solutions obtained (see Fig. \ref{fig:viewDTLZ9}). From Table \ref{tab:problemwise2}, it can be seen that the clustering-based methods and the cone$\eps$-MOEA have presented the best measures in terms of solution diversity. However, the $\eps$-MOEA strongly outperformed all other approaches in terms of $\gamma$ and $CS$ values, followed by the cone$\eps$-MOEA in these metrics. Once again, the S-metric ($HV$) was not considered in this problem since redundant solutions may affect the ability of the hypervolume indicator to adequately quantify the quality of the fronts. 

\subsection{Cardinality analysis}
\begin{figure*}[!t]
\centering
\psfrag{cone}[][]{\small cone}
\psfrag{epsi}[][]{\small $\eps$}
\psfrag{0}[][]{\tiny 0}
\psfrag{20}[][]{\tiny 20}
\psfrag{40}[][]{\tiny 40}
\psfrag{60}[][]{\tiny 60}
\psfrag{80}[][]{\tiny 80}
\psfrag{100}[][]{\tiny 100}
\includegraphics[width=\textwidth]{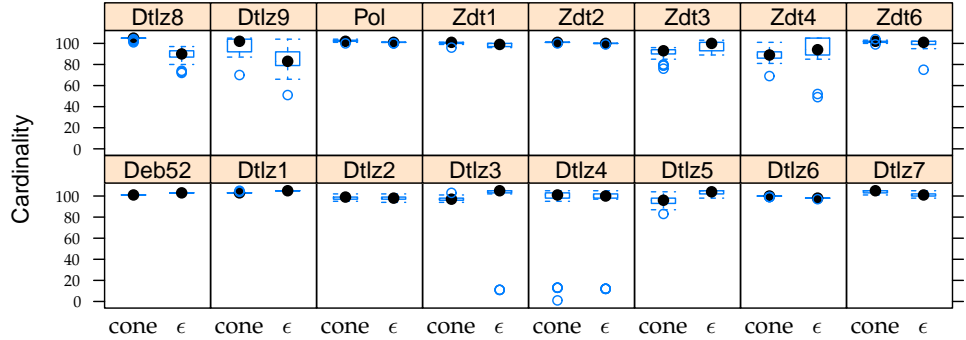}
\caption{Distribution of cardinalities of the final fronts returned by the $\eps$-MOEA and cone$\eps$-MOEA for the test problems considered.}
\label{fig:card_probs}
\end{figure*}

It is interesting to verify that the estimated values of $\eps$ used for the $\eps$-MOEA and cone$\eps$-MOEA were able to provide final archive sizes of approximately 100 solutions. To be more precise, the mean and standard error for the cardinality of the $\eps$-MOEA was $96.6\pm 0.51$, and for the cone$\eps$-MOEA it was $98.2\pm 0.40$. While both approaches had mean values slightly smaller than the nominal 100 (which was achieved in all runs by the NSGA-II, NSGA-II*, C-NSGA-II and SPEA2), the analysis of the results obtained does not indicate that this had any deleterious effect on these two algorithms. Figure \ref{fig:card_probs} illustrates the distribution of the cardinalities of the final fronts obtained by the $\eps$-MOEA and cone$\eps$-MOEA for the 16 test problems considered.

While the results above suggest no evidence that the mean final population sizes of the algorithms are different, some questions remain open to question. In particular the issue of whether the $\eps$- or cone$\eps$-dominance criteria are capable of producing competitive or superior results when exactly the same number of points is considered is a point that deserves some attention.

To address the issue above, a complementary experiment was performed in order to isolate the possible effects of different cardinalities from the effects of the dominance criteria themselves. The experiment takes advantage from the fact that the test problems employed in this work have known, very finely-sampled Pareto-optimal fronts, and was designed as follows.

A number of desired points was chosen, and each dominance criteria was allowed to choose a set of exactly the prescribed size. In the case of the $\eps$-dominance, the values of $\eps$ were calculated using \eqref{eq:eps_value}, and afterwards adjusted using small perturbations until the desired number of solutions was reached. For the cone$\eps$-dominance, eleven values of $\kappa$ were chosen, covering the entire range of this parameter. For each value, the $\eps$ values were calculated using \eqref{eq:coneps}, and then adjusted to match the desired cardinality. For the one-at-a-time crowding distance and the clustering criteria from SPEA1/C-NSGA-II and SPEA2, no parameters other than the desired set size had to be set.

After each of the dominance criteria returned the selected sample from the Pareto set, the quality metrics $\Delta$ and $HV$ were calculated for the sets\footnote{In this case comparisons of the $\gamma$ and $CS$ metrics would be pointless, since all criteria are already working on samples of the Pareto-optimal front.}, and compared using the criteria as the experimental factor levels, and both Problem and Cardinality as blocking factors. The parameters used in the experiment were:
\begin{itemize}
\item \textbf{Cardinality}: twenty values of cardinality, from 181 to 200 points, were used;
\item \textbf{Problems}: six problems from the test set were used: DEB52, POL, DTLZ2, DTLZ7, ZDT1 and ZDT3;
\item \textbf{Criteria}: the criteria tested were:\begin{itemize}
\item Crowding distance (CD), as implemented in the NSGA-II*;
\item Clustering (CL1), as implemented in the C-NSGA-II;
\item Clustering (CL2), as implemented in the SPEA2;
\item $\eps$-dominance ($\eps$);
\item cone$\eps$-dominance (cone);
\end{itemize}
\item $\kappa$ values: the kappa values were chosen so as to cover the entire variation range of this parameter. Nine equally-spaced points were included ($0.1,\ 0.2, \ldots,\ 0.9$) together with two extreme points ($0.01$ and $0.99$);
\end{itemize}

The analysis of this experiment was performed in two parts: first, the effects of the $\kappa$ values on the performance of the cone$\eps$-criterion on the $\Delta$ and $HV$ metrics was investigated. The best-, median-, and worst-case instances of the cone$\eps$-dominance were then included in an overall comparison against the performance of the remaining criteria. In both cases, the randomized complete block design \citep{book.Montgomery2008} was used, with both \textit{Problems} and \textit{Desired Cardinality} as blocking factors.

\begin{figure*}[!htb]
\centering
\psfrag{0.01}[][]{\scriptsize $0.01$}
\psfrag{0.1}[][]{\scriptsize $0.1$}
\psfrag{0.2}[][]{\scriptsize $0.2$}
\psfrag{0.3}[][]{\scriptsize $0.3$}
\psfrag{0.4}[][]{\scriptsize $0.4$}
\psfrag{0.5}[][]{\scriptsize $0.5$}
\psfrag{0.6}[][]{\scriptsize $0.6$}
\psfrag{0.7}[][]{\scriptsize $0.7$}
\psfrag{0.8}[][]{\scriptsize $0.8$}
\psfrag{0.9}[][]{\scriptsize $0.9$}
\psfrag{0.99}[][]{\scriptsize $0.99$}
\psfrag{0.44}[][]{\scriptsize $0.44$}
\psfrag{0.46}[][]{\scriptsize $0.46$}
\psfrag{0.48}[][]{\scriptsize $0.48$}
\psfrag{0.50}[][]{\scriptsize $0.50$}
\psfrag{0.52}[][]{\scriptsize $0.52$}
\psfrag{0.9816}[][]{\scriptsize $0.9816$}
\psfrag{0.9818}[][]{\ }
\psfrag{0.9820}[][]{\scriptsize $0.9820$}
\psfrag{0.9822}[][]{\ }
\psfrag{0.9824}[][]{\scriptsize $0.9824$}
\psfrag{delta}[][]{\footnotesize $\Delta$}
\psfrag{kappa}[][]{\footnotesize $\kappa$}
\psfrag{HV}[][]{\footnotesize $HV$}
\psfrag{kappaDelta}[][]{\ }
\psfrag{kappaHV}[][]{\ }
\subfigure[\small $\kappa$ effects on the $\Delta$ metric]{\includegraphics[width=0.45\textwidth]{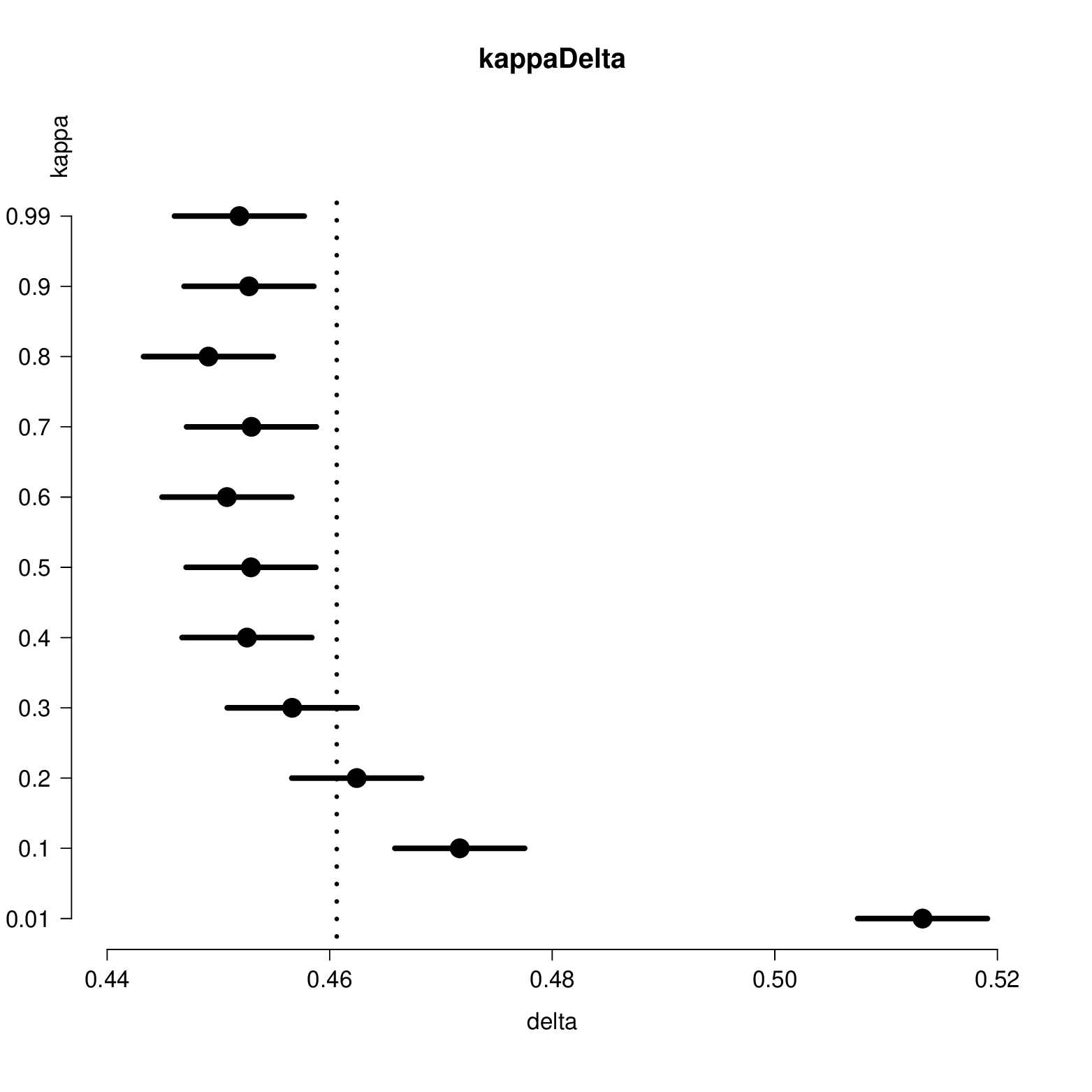}}
\subfigure[\small $\kappa$ effects on the $HV$ metric]{\includegraphics[width=0.45\textwidth]{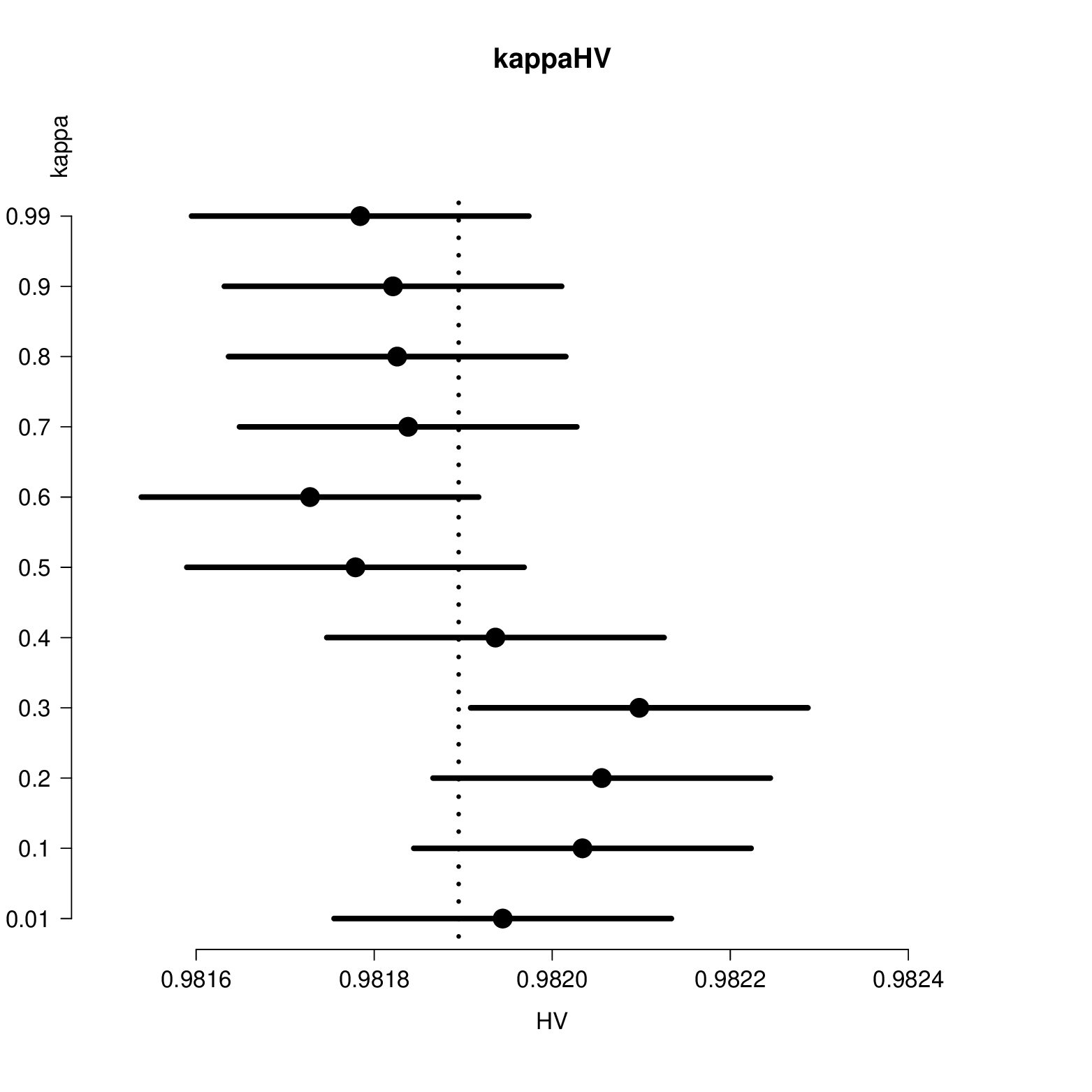}}
\caption{Effect of different $\kappa$ values on the performance of the cone$\eps$-dominance, after removing problem and cardinality effects. The bars represent $99\%$ confidence intervals, and the dotted line shows the grand mean for each metric.}
\label{fig:expkappa}
\end{figure*}

Figure \ref{fig:expkappa} shows the results concerning the effects of $\kappa$ on the performance of the cone$\eps$-dominance approach in this matched-cardinality experiment. These results suggest that, for the test problems employed, the performance of this approach regarding both metrics is reasonably robust to variations in the value of $\kappa$ within a broad range, with no significant differences for  $\kappa\geq 0.3$. As the value of $\kappa$ is reduced below $0.3$, the performance concerning the $\Delta$ metric significantly worsens, with a marked decrease in the effectiveness of the criterion for $\kappa = 0.01$. For the $HV$ metric no statistically significant differences were detected.

\begin{figure*}[!htb]
\centering
\psfrag{0.35}[][]{\scriptsize $0.35$}
\psfrag{0.40}[][]{\scriptsize $0.40$}
\psfrag{0.45}[][]{\scriptsize $0.45$}
\psfrag{0.50}[][]{\scriptsize $0.50$}
\psfrag{0.55}[][]{\scriptsize $0.55$}
\psfrag{0.60}[][]{\scriptsize $0.60$}
\psfrag{0.9800}[][]{\scriptsize $0.980$}
\psfrag{0.9805}[][]{\ }
\psfrag{0.9810}[][]{\scriptsize $0.981$}
\psfrag{0.9815}[][]{\ }
\psfrag{0.9820}[][]{\scriptsize $0.982$}
\psfrag{0.9825}[][]{\ }
\psfrag{0.9830}[][]{\scriptsize $0.983$}
\psfrag{cone(kappa=0.8)}[][]{\scriptsize Cone($0.8$)}
\psfrag{cone(kappa=0.5)}[][]{\scriptsize Cone($0.5$)}
\psfrag{cone(kappa=0.01)}[][]{\scriptsize Cone($0.01$)}
\psfrag{eps}[][]{\footnotesize $\eps$}
\psfrag{CL1}[][]{\scriptsize CL1}
\psfrag{CL2}[][]{\scriptsize CL2}
\psfrag{CD}[][]{\scriptsize CD}
\psfrag{meanDelta}[][]{\ }
\psfrag{meanHV}[][]{\ }
\psfrag{delta}[][]{\footnotesize $\Delta$}
\psfrag{HV}[][]{\footnotesize $HV$}
\subfigure[\small Comparison of mean $\Delta$ performance]{\includegraphics[width=0.45\textwidth]{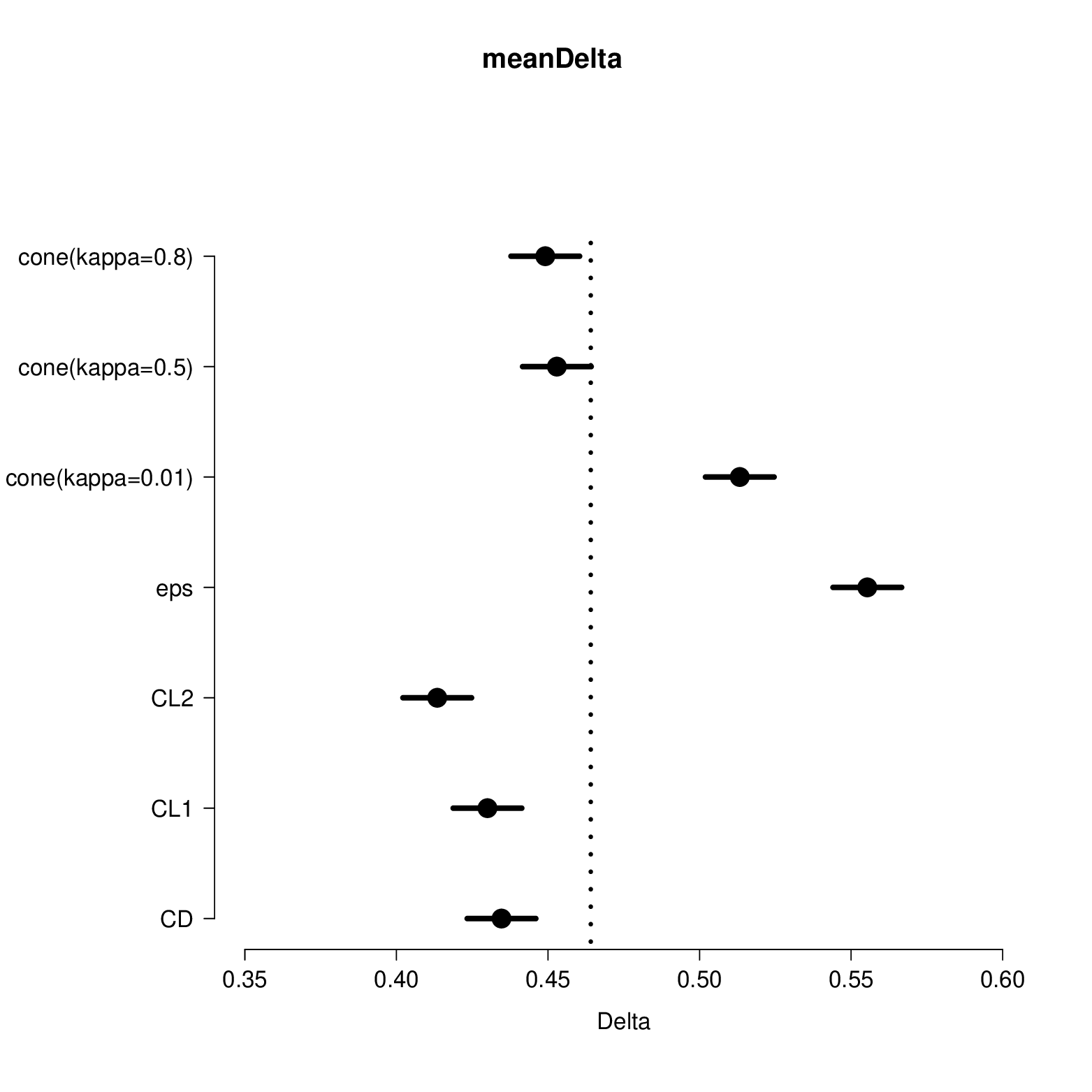}}
\subfigure[\small Results for $HV$ metric]{\includegraphics[width=0.45\textwidth]{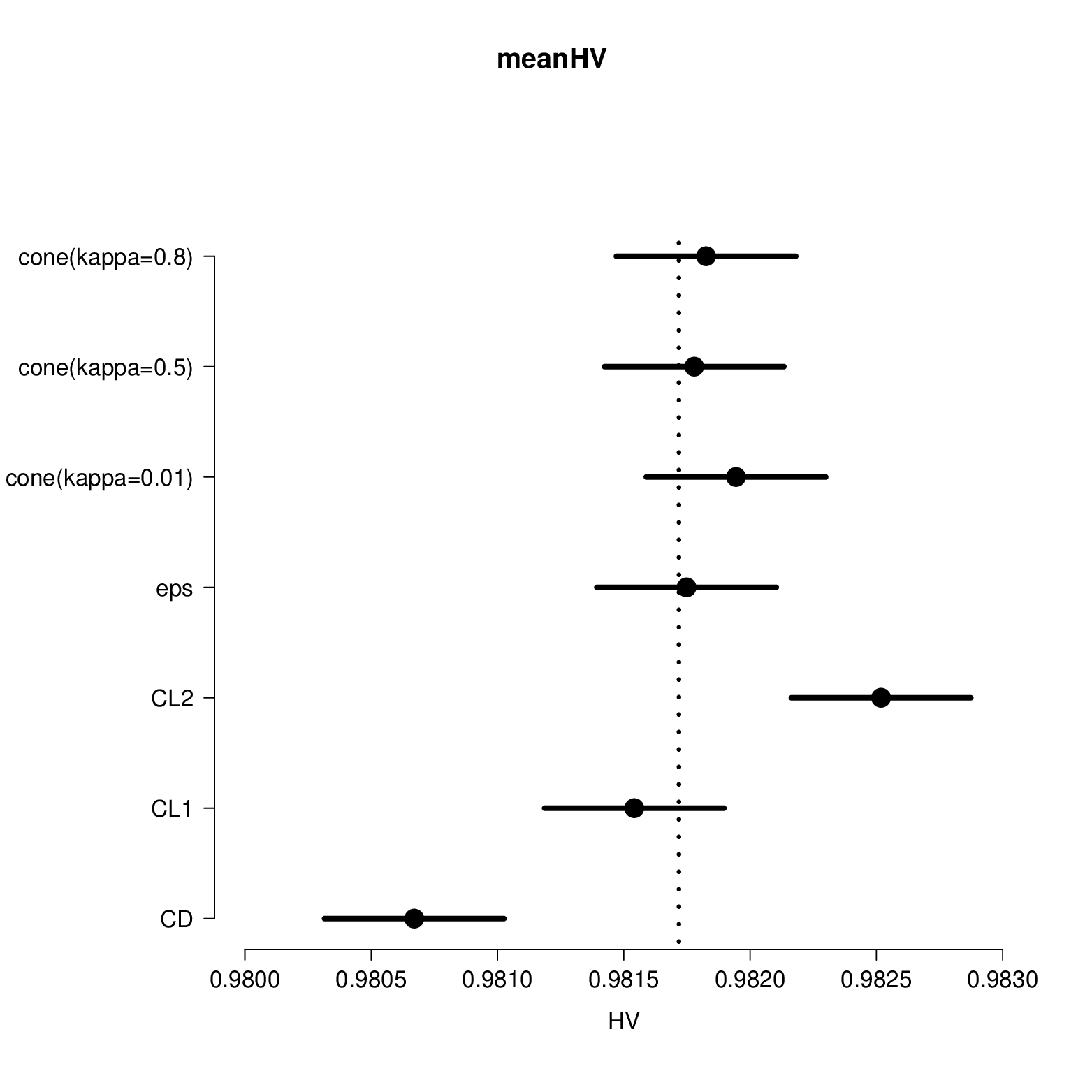}}
\caption{Performance comparison of the dominance criteria used in the MOEAs employed in this work, after removing problem and cardinality effects. The bars represent $99\%$ confidence intervals, and the dotted line shows the grand mean for each metric.}
\label{fig:card_res}
\end{figure*}

From the results of the $\kappa$ analysis, three cases were selected for the overall analysis: worst, median, and best case performance. Since the $HV$ metric presented no significant differences, the point estimator of the $\Delta$ metric was used as the ranking variable. The worst, median, and best cases were, respectively, $\kappa = 0.01$, $\kappa = 0.5$ and $\kappa=0.8$. These three variants of the cone$\eps$ approach were compared with the remaining criteria, yielding the results shown in Figure \ref{fig:card_res}.

These results suggest an interesting view of the ability of the dominance criteria examined to maintain a diverse set of solutions. As expected, the $HV$ metric did not present relevant differences\footnote{While the differences were statistically significant, the practical effect is essentially zero, as the magnitude of the differences for this metric is vanishingly small.}, as the criteria were already working on samples taken from the  Pareto-optimal set. As for the $\Delta$ metric, it can be seen that the statistical ordering of the results, at the $99\%$ confidence level, was:

\begin{center}
$\overline{\mbox{CL2\ \ }\underline{\mbox{\ CL1\ \ \ CD\ }}}\underline{\mbox{\ Cone0.8\ \ \ Cone0.5\  }}\ \ \ \  \mbox{Cone0.01}\ \ \ \ \eps$
\end{center}

\vskip 1em
These results suggest that the only criterion significantly better than the best- and median-case cone$\eps$-approach was the clustering mechanism used in the SPEA2. The $\eps$-dominance had the worst diversity performance in this experiment, followed by the worst-case cone$\eps$ which, with $\kappa=0.01$, was close to the condition where it effectively becomes equal to the $\eps$ criterion. In general, the cone$\eps$-dominance criterion seems to be a competitive alternative to the clustering and crowding distance approaches as long as it employs its usual values for the $\kappa$ parameter, e.g., the median case $\kappa=0.5$ used in the experiments reported throughout this paper. Moreover, the $\eps$-based methods were not designed for the purpose of imposing an ordering on a non-dominated set, while the clustering and crowding distance methods allow for this ordering, making them less sensitive, in terms of diversity metrics, to the removal of points. These characteristics may explain the statistical ordering observed in the experiment.

\subsection{Possible Extensions}
From the results discussed previously, one can notice that while generally effective, the cone$\eps$-method appears to have difficulties in dealing with the redundant solutions (see Fig. \ref{fig:viewDTLZ9c}), which may negatively affect its behavior. In problems where redundant solutions are known to exist, the performance of the cone$\eps$-approach could be improved by using smaller $\kappa$ values. More importantly, some authors argue that the probability of occurrence of this ``redundancy problem'' becomes higher as the number of objectives is increased \citep{proc.Deb2005}. Indeed, an extension of the cone$\eps$-concept should be used to handle these classes of problems. A simple and efficient strategy can be obtained by extending the area cone-dominated by a solution, e.g., as in the $\alpha$-dominance \citep{inproc.Ikeda2001}. Since in this case the cone of dominance ensures the Pareto optimality, only a single linear system should be solved, thus decreasing the computational complexity of the method. Also, the adoption of a hypergrid in the objective domain remains very useful in order to guarantee the convergence and the diversity preservation during the evolution of a MOEA. However, a deeper investigation on the characteristics and performance of the cone$\eps$-dominance on many-objective optimization problems (MaOPs) is left for a future work. Preliminary results on the ability of the $\eps$- and  cone$\eps$-dominance criteria to adequately order sets of points in multi and many-objective optimization problems can be found in  \citep{inproc.Batista2011b}.

\section{Conclusions}
\label{sec:7}

In this paper we presented a relaxed form of dominance, named the cone$\eps$-dominance, for improving the convergence and diversity preservation capabilities of multiobjective evolutionary algorithms. This alternative dominance criterion can be seen as a hybridization of the concepts of $\eps$-dominance and of proper efficiency with respect to cones. After presenting the mathematical definition of the cone$\eps$-dominance, we incorporated this concept into a steady-state MOEA, and showed that the computational complexity of this MOEA is $\mathcal{O}\left((mN)^2\right)$ in the worst case. 

A comprehensive evaluation of the cone$\eps$-dominance strategy was also presented. We employed six MOEAs to evaluate the relative effectiveness of the proposed methodology: NSGA-II, SPEA2, C-NSGA-II, $\eps$-MOEA, NSGA-II*, and cone$\eps$-MOEA. The performance of these algorithms was quantified by means of four different metrics, selected to evaluate the convergence and diversity of the final fronts obtained by each algorithm: the convergence metric ($\gamma$), the diversity metric ($\Delta$), the hypervolume or s-metric ($HV$), and a proposed generalized form of the coverage of two sets metric, named the coverage of many sets ($CS$). For the experimental design, sixteen two- and three-objective test problems were selected, including the classic DTLZ and ZDT benchmark families.

The results obtained for the four metrics were analyzed both in terms of statistical significance and magnitude of the differences. From the analysis of the effect sizes, calculated for each algorithm on each metric after removing the problem effects, we could conclude that the cone$\eps$-MOEA was capable of providing an efficient balance between convergence and diversity. These results strongly support the conclusion that the proposed algorithm is an interesting and competitive approach for the solutions of multiobjective optimization problems.

\section*{Acknowledgment}
The authors would like to acknowledge the insightful comments provided by the anonymous reviewers, who greatly contributed for the quality of this paper. Support for this work was provided by the National Council for Research and Development (CNPq), grants 306910/2006-3, 141819/2009-0, 305506/2010-2 and 472446/2010-0; and by the Research Foundation of the State of Minas Gerais (FAPEMIG, Brazil), grants Pronex: TEC 01075/09 and Pronem: CEX APQ-04611-10. 

\small

\bibliographystyle{apalike}
\bibliography{TechnicalReport2013-01}

\end{document}